\documentclass[letterpaper, 10 pt, conference]{ieeeconf}

\IEEEoverridecommandlockouts

\usepackage{url}
\usepackage{multirow}
\usepackage{float}
\usepackage{booktabs}
\usepackage{amsmath}
\usepackage{amssymb}
\usepackage{graphicx}
\usepackage{multirow}
\usepackage{url}
\usepackage{colortbl}
\usepackage{hyperref}
\usepackage{pifont} 
\usepackage{subcaption}
\usepackage{xcolor} 
\usepackage{mathabx}
\usepackage{hhline}
\usepackage{makecell}

\newcommand{\argmin}{\mathop{\mathrm{argmin}}}

\definecolor{RowColor}{rgb}{0.97, 0.97, 1}

\newcommand{\cmark}{\textcolor[rgb]{0.004, 0.663, 0}{\ding{51}}} 
\newcommand{\xmark}{\textcolor{red}{\ding{55}}}
\title{\LARGE \bf
Multi-Cali Anything: Dense Feature Multi-Frame Structure-from-Motion for Large-Scale Camera Array Calibration
}

\author{
Jinjiang~You$^{*}$, Hewei~Wang$^{*}$, Yijie~Li, Mingxiao~Huo, Long~Vân~Tran~Ha, Mingyuan Ma, \\Jinfeng~Xu, Jiayi~Zhang, Puzhen~Wu, Shubham~Garg, Wei~Pu 
\thanks{* Equal contribution}
\thanks{Jinjiang You, Hewei Wang, Yijie Li, and Long~Vân~Tran~Ha are with School of Computer Science, Carnegie Mellon University, Pittsburgh, PA 15213, USA (e-mail: jinjiany@alumni.cmu.edu, heweiw@alumni.cmu.edu, yijieli@andrew.cmu.edu, ltranha@alumni.cmu.edu)}
\thanks{Mingxiao Huo is with the Electrical and Computer Engineering, Carnegie Mellon University, Pittsburgh, PA 15213, USA (e-mail: mingxiah@andrew.cmu.edu)}
\thanks{Mingyuan Ma is with Harvard John A. Paulson School of Engineering And Applied Sciences, Harvard University, Boston, MA 02134, USA (e-mail: mingyuan\_ma@g.harvard.edu)}
\thanks{Jinfeng Xu is with the Department of Electrical and Electronic Engineering, The University of Hong Kong, Hong Kong 999077, China (e-mail: jinfeng@connect.hku.hk)}
\thanks{Jiayi Zhang is with the School of Mathematical Science, University of Nottingham, Nottingham, NG7 2RD, United Kingdom (e-mail: smyjz19@nottingham.edu.cn)}
\thanks{Puzhen Wu is with Weill Cornell Medicine, Cornell University, New York, NY 10065, USA (e-mail: puw4002@med.cornell.edu)}
\thanks{Shubham~Garg and Wei~Pu are with Meta Reality Labs, Pittsburgh, PA 15222, USA (e-mail: ssgarg@meta.com, wpu@meta.com)}
\thanks{Send correspondence to Shubham Garg (e-mail: ssgarg@meta.com)}
}

\begin{document}

\maketitle
\thispagestyle{empty}
\pagestyle{empty}

\begin{abstract}
Calibrating large-scale camera arrays, such as those in dome-based setups, is time-intensive and typically requires dedicated captures of known patterns. While extrinsics in such arrays are fixed due to the physical setup, intrinsics often vary across sessions due to factors like lens adjustments or temperature changes. In this paper, we propose a dense-feature-driven multi-frame calibration method that refines intrinsics directly from scene data, eliminating the necessity for additional calibration captures. Our approach enhances traditional Structure-from-Motion (SfM) pipelines by introducing an extrinsics regularization term to progressively align estimated extrinsics with ground-truth values, a dense feature reprojection term to reduce keypoint errors by minimizing reprojection loss in the feature space, and an intrinsics variance term for joint optimization across multiple frames. Experiments on the Multiface dataset show that our method achieves nearly the same precision as dedicated calibration processes, and significantly enhances intrinsics and 3D reconstruction accuracy. Fully compatible with existing SfM pipelines, our method provides an efficient and practical plug-and-play solution for large-scale camera setups. Our code is publicly available at: \url{https://github.com/YJJfish/Multi-Cali-Anything}
\end{abstract}

\section{INTRODUCTION}
\label{sec:introduction}

Camera calibration is a crucial task in computer vision, particularly for applications like multi-view 3D reconstruction, virtual reality (VR), and augmented reality (AR). It involves estimating intrinsic parameters (e.g., focal lengths, principal points, and distortion coefficients) and extrinsic parameters (rotation and translation) that define the geometric relationship between the image and the world. Traditional methods use images of well-defined patterns, such as checkerboards, to achieve accurate parameter estimation.

While effective, these methods face limitations in large-scale systems, such as VR headsets or camera domes with dozens or hundreds of cameras. The primary challenge lies in the necessity for an additional calibration capture session before each scene capture. Although the extrinsics in these systems remain stable due to their physical setup, intrinsics often change due to factors like lens adjustments or temperature changes. Consequently, every time the intrinsics change, a new time-consuming calibration process is required before capturing actual scene data. The inefficiency of this workflow motivates the need for more efficient solutions.

Recent advances in Structure-from-Motion (SfM) methods, such as Pixel-Perfect SfM~\cite{pixelsfm} and VGGSfM~\cite{wang2023visual}, have provided efficient tools for joint estimation of camera parameters and scene structure. These methods can perform camera calibration directly from scene data without separate calibration sessions. However, SfM pipelines are typically designed to estimate extrinsics and intrinsics simultaneously, and do not provide an option to utilize known ground-truth extrinsics and refine only intrinsics, leading to suboptimal intrinsics calibration. In large-scale camera array setups where the extrinsics are already well-known, this lack of flexibility results in unnecessary estimation errors and reduced accuracy. Besides, the single frame processing of these pipelines causes inconsistencies in multi-frame datasets, reducing robustness.

To overcome these challenges, we propose a dense-feature-driven multi-frame calibration method optimized for large-scale setups. By leveraging outputs from SfM pipelines and assuming stable extrinsics from a one-time calibration, our method refines intrinsic parameters and improves sparse 3D reconstructions without requiring dedicated calibration captures. This approach is ideal for scenarios where extrinsics are stable but intrinsics frequently change, achieving state-of-the-art results on the Multiface~\cite{multiface} dataset.

The primary contributions of our method are threefold:
\begin{itemize}
    \item We propose an extrinsics regularization method to iteratively refine estimated extrinsics toward ground-truth, ensuring robust convergence without local minima.
    \item We introduce a dense feature reprojection term that minimizes reprojection errors in the feature space, reducing the influence of keypoint noise.
    \item We design a multi-frame optimization method which uses an intrinsics variance term to get more accurate intrinsics across frames for multi-frame datasets.
\end{itemize}

\begin{figure*}[h]
        \vspace{1.5mm}
	\centering
	\includegraphics[width=7in]{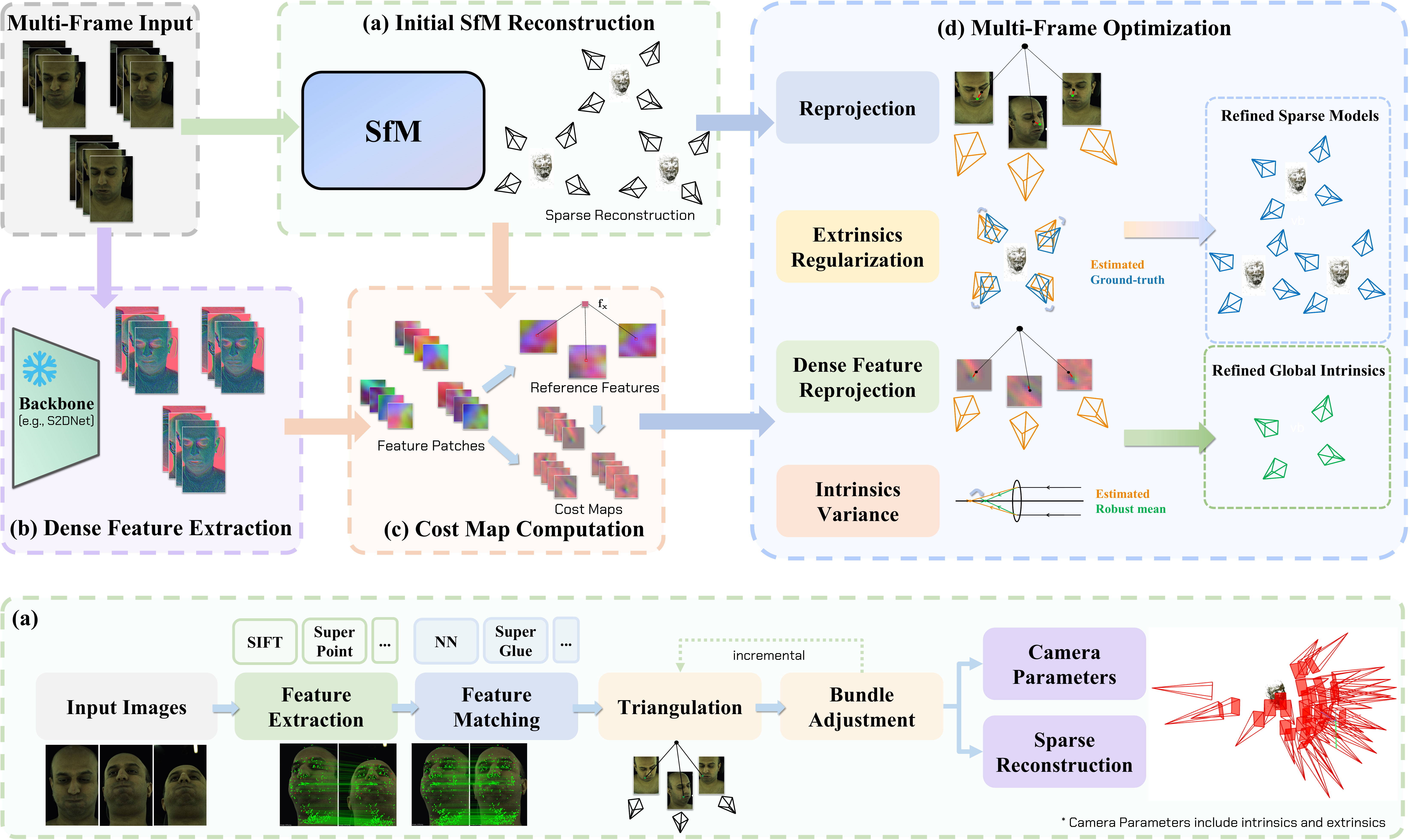}
	\caption{The overall pipeline of our proposed method. Inputs include multi-frame, multi-view images, and camera extrinsics, with outputs being camera intrinsics and SfM sparse reconstructions. (a) to (d) illustrate several key components.}
        \label{architecture}
\end{figure*}

This paper is organized as follows. Section~\ref{sec:related-work} reviews prior work on camera calibration and SfM. Section~\ref{sec:approach} presents our proposed method in detail. Section~\ref{sec:experiments} conducts experiments on the Multiface dataset, including quantitative comparisons, qualitative results, and ablation study. Section~\ref{sec:conclusion} summarizes the contributions and outlines future work.

\section{RELATED WORK}
\label{sec:related-work}
Camera calibration and SfM have been extensively studied for reconstructing 3D scenes and estimating camera parameters. Here, we review prior work in three key areas: traditional calibration methods, general SfM techniques, and recent advances leveraging dense features, highlighting their limitations in the context of large-scale camera arrays with fixed extrinsics and varying intrinsics.

Traditional camera calibration typically relies on controlled setups with known patterns or objects. For instance, Loaiza \textit{et al.}~\cite{loaiza2011multi} proposed a one-dimensional invariant pattern with four collinear markers, leveraging its simplicity for multi-camera systems. Usamentiaga \textit{et al.}~\cite{usamentiaga2019multi} introduced a calibration plate with laser-illuminated protruding cylinders for precise 3D measurements, while Huang \textit{et al.}~\cite{huang2019research} utilized a cubic chessboard to enforce multi-view constraints. These methods offer high accuracy but require dedicated calibration sessions, limiting scalability for large-scale arrays due to spatial constraints and setup overhead. To address scalability, Moreira \textit{et al.}~\cite{vican2024} proposed VICAN, an efficient algorithm for large camera networks that estimates extrinsics using dynamic objects and bipartite graph optimization. While effective for extrinsic calibration, VICAN does not refine intrinsics independently, a critical need in fixed setups where intrinsics vary across sessions due to lens adjustments or environmental factors.

SfM-based calibration methods provide a scalable alternative by jointly estimating camera parameters and 3D structures from scene data. COLMAP~\cite{sfm-revisited, pixel-view-select-mvs} is a widely adopted pipeline, balancing robustness and efficiency through incremental optimization. OpenMVG~\cite{moulon2017openmvg} and HSfM~\cite{cui2017hsfm} enhance scalability with modular designs and hierarchical processing, respectively. Pixel-Perfect SfM~\cite{pixelsfm} improves precision by refining keypoints with featuremetric optimization, reducing reprojection errors. Recent global approaches, such as GLOMAP~\cite{pan2025global}, revisit SfM with deep features for consistent multi-view registration. However, these methods typically estimate intrinsics and extrinsics simultaneously, lacking flexibility to refine intrinsics alone when extrinsics are fixed—a common scenario in dome-based camera arrays.

Deep learning has further advanced SfM. VGGSfM~\cite{wang2023visual} introduces an end-to-end differentiable pipeline with visual geometry grounding, improving feature robustness, though it assumes centered principal points, reducing adaptability. DUSt3R~\cite{wang2024dust3r} predicts dense point maps unsupervised, enhancing 3D reconstruction but sacrificing geometric precision in camera parameters. Feature extraction methods like SIFT~\cite{sift}, SuperPoint~\cite{superpoint}, R2D2~\cite{r2d2}, and DISK~\cite{disk}, paired with matching techniques such as SuperGlue~\cite{superglue} and S2DNet~\cite{s2dnet}, provide reliable correspondences, indirectly supporting SfM. Despite these advances, general SfM methods lack mechanisms for multi-frame consistency, crucial for refining intrinsics across sessions in fixed arrays.

Recent SfM research has shifted toward dense feature matching to improve robustness and reconstruction quality. Seibt \textit{et al.}~\cite{dfm4sfm2023} introduced DFM4SFM, which leverages dense feature matching with homography decomposition to enhance correspondence accuracy in SfM, particularly for challenging scenes. Similarly, Lee and Yoo~\cite{lee2025densesfmstructuremotiondense} proposed Dense-SfM, integrating dense consistent matching with Gaussian splatting to achieve dense, accurate 3D reconstructions, excelling in texture-sparse regions. Detector-Free SfM~\cite{he2024detector} eliminates traditional keypoint detectors, relying entirely on dense matching for robustness, though at a high computational cost. These methods demonstrate the power of dense features but are designed for general SfM tasks, optimizing both intrinsics and extrinsics without exploiting known extrinsics or ensuring multi-frame consistency.

Our approach is motivated to target large-scale camera arrays with fixed extrinsics. We build on existing SfM pipelines by introducing additional refinements with extrinsics regularization, dense-feature reprojection, and multi-frame optimization to refine intrinsics directly from scene data. Unlike traditional calibration methods, our method eliminates the necessity for dedicated calibration captures while still achieving high precision. Different from other SfM pipelines~\cite{sfm-revisited, pixelsfm, wang2023visual} which do not provide intrinsic-only refinement, our method utilizes known ground-truth extrinsics to improve the accuracy of intrinsics. Achieving nearly the same precision as dedicated calibration methods on the Multiface~\cite{multiface} dataset, our plug-and-play solution advances scalable calibration for real-world applications.

\section{APPROACH}
\label{sec:approach}
Fig. \ref{architecture} illustrates the architecture of our proposed method: (a) shows the initial SfM process, which acts as the foundation of the pipeline. (b) highlights the dense feature extraction module, where a backbone network, such as S2DNet, is used to compute dense feature maps from the input images, capturing rich feature details. (c) focuses on cost map computation, a crucial step for optimizing memory and computational efficiency. Computing reprojection errors based on raw dense feature maps can be time-consuming and memory-intensive. Instead, these dense feature maps can be proprocessed into compact cost maps to significantly reduce resource usage and improve efficiency. Finally, (d) presents the optimization stage, which performs a multi-frame dense-feature-driven bundle adjustment (BA). This module refines both sparse reconstructions and camera parameters, outputting a globally consistent set of camera intrinsics and enhanced sparse reconstructions.

\subsection{Background}
\label{sec:approach-background}

In this work, a frame refers to a specific time point at which multiple cameras simultaneously capture the object. Each camera captures one image per frame, resulting in a unique image for every combination of a frame and a camera. Formally, given a dataset with $N_F$ frames and $N_C$ cameras, the total number of captured images is $N_F \cdot N_C$. This assumption is valid under large-scale camera array in dome-based setups. Besides, we assume all cameras follow the pinhole camera model, with four intrinsic parameters: $f_x$, $f_y$, $c_x$, $c_y$. However, it is easy to be extended to more complex models.

Traditional BA minimizes the total reprojection loss:

\begin{equation}
    \label{eq:reprojection}
    \text{\textit{Reproj}} = 
	\sum_{\mathbf{x} \in \mathcal{X}_i}{
		\sum_{(j, \mathbf{p}) \in \mathcal{T}_\mathbf{x}} {
			\rho(\left\|
				\Pi(\mathbf{r}_{i,j}, \mathbf{t}_{i,j}, K_{i,j}, \mathbf{x}) - 
				\mathbf{p}
			\right\|)
		}
	}
\end{equation}
where $i$ is the frame index. The subscript $i$ is needed because traditional BA only works with single frame data. $\mathcal{X}_i$ is the set of triangulated 3D points in the $i$-th frame. $\mathcal{T}_\mathbf{x}$ is a set of (camera index, 2D keypoint) pairs in the track of the 3D point $\mathbf{x}$. For example, $(j, \mathbf{p}) \in \mathcal{T}_\mathbf{x}$ means the 3D point $\mathbf{x}$ is observed in the $j$-th camera view, and its corresponding 2D keypoint in the image space is $\mathbf{p}$. $\mathbf{r}_{i,j}$, $\mathbf{t}_{i,j}$, $K_{i, j}$ are estimated extrinsics and intrinsics of the $j$-th camera, based on the $i$-th frame data. $\Pi(\cdot)$ transforms and projects a point in the 3D world space to 2D image space. $\rho(\cdot)$ is a loss function (e.g. Cauchy loss). In this work, we use:

\begin{equation}
\label{eq:normalized-reprojection}
\mathcal{L}_0=\frac{\lambda_0}{\sum\limits_{\mathbf{x} \in \mathcal{X}_i}\left|\mathcal{T}_{\mathbf{x}}\right|} 
\sum\limits_{\mathbf{x} \in \mathcal{X}_i} \sum\limits_{(j, \mathbf{p}) \in \mathcal{T}_{\mathbf{x}}} 
\rho\left(\left\|\Pi(\mathbf{r}_{j}, \mathbf{t}_{j}, K_{j}, \mathbf{x})-\mathbf{p}\right\|\right)
\end{equation}
where $\mathcal{L}_0$ represents the normalized reprojection loss, which is normalized by the total number of observations $\sum_{\mathbf{x} \in \mathcal{X}i} \left|\mathcal{T}_{\mathbf{x}}\right|$ and incorporates a weighting factor $\lambda_0$ to balance with other loss terms, ensuring robustness across frames with varying numbers of observations.

\begin{table*}[htbp]
\renewcommand{\arraystretch}{1.2}
	\centering
        \vspace{1.5mm}
        \caption{Quantitative comparison on the Multiface Dataset with state-of-the-art methods and our method on multiple metrics. We highlight the best-performing and second-best methods in \textbf{bold} and \underline{underline}, respectively.}
	\scalebox{0.605}{
    	\begin{tabular}{lcccccccccccc}
    		\Xhline{1pt}
    		\multicolumn{1}{l|}{\textbf{Method}} & $\mathrm{focal}_\mathrm{abs,mean}$ & $\mathrm{focal}_\mathrm{abs,max}$ & $\mathrm{focal}_\mathrm{abs,min}$ & $\mathrm{focal}_\mathrm{rel,mean}$\textperthousand & $\mathrm{focal}_\mathrm{rel,max}$\textperthousand & \multicolumn{1}{c|}{$\mathrm{focal}_\mathrm{rel,min}$\textperthousand} & $\mathrm{pp}_\mathrm{abs,mean}$ & $\mathrm{pp}_\mathrm{abs,max}$ & $\mathrm{pp}_\mathrm{abs,min}$ & $\mathrm{pp}_\mathrm{rel,mean}$\textperthousand & $\mathrm{pp}_\mathrm{rel,max}$\textperthousand & $\mathrm{pp}_\mathrm{rel,min}$\textperthousand \\ 
    		\Xhline{1pt}
                \multicolumn{1}{l|}{COLMAP~\cite{sfm-revisited, pixel-view-select-mvs} (SIFT~\cite{sift}+NN)} &
                240.107 & 370.769 & 197.697 & 31.610 & 48.867 & \multicolumn{1}{c|}{26.095} & 66.786 & 112.163 & 47.742 & 42.415 & 72.848 & 29.834 \\
                
                \rowcolor{RowColor} \multicolumn{1}{l|}{$\hookrightarrow$ Fix Extrinsic} & 
                117.781 & 122.004 & 112.627 & 15.574 & 16.136 & \multicolumn{1}{c|}{14.897} & \textbf{1.594} & \textbf{1.682} & \underline{1.455} & \textbf{0.955} & \textbf{1.010} & \textbf{0.873} \\

                \multicolumn{1}{l|}{Pixel-Perfect~\cite{pixelsfm} (Disk~\cite{disk}+NN)} &
                313.362 & 359.700 & 268.225 & 41.203 & 47.331 & \multicolumn{1}{c|}{35.277} & 66.886 & 76.557 & 58.248 & 42.040 & 49.425 & 36.684 \\
                
                \rowcolor{RowColor} \multicolumn{1}{l|}{$\hookrightarrow$ Fix Extrinsic} & 
                179.227 & 183.157 & 176.467 & 23.683 & 24.205 & \multicolumn{1}{c|}{23.320} & 5.264 & 5.447 & 5.078 & 2.909 & 3.001 & 2.798 \\

                \multicolumn{1}{l|}{Pixel-Perfect (R2D2~\cite{r2d2}+NN)} &
                126.477 & 211.908 & 64.953 & 16.540 & 27.800 & \multicolumn{1}{c|}{8.501} & 61.212 & 96.716 & 42.744 & 39.093 & 63.352 & 26.318 \\
                
                \rowcolor{RowColor} \multicolumn{1}{l|}{$\hookrightarrow$ Fix Extrinsic} & 
                91.132 & 103.760 & 83.591 & 12.052 & 13.727 & \multicolumn{1}{c|}{11.054} & 2.950 & 3.358 & 2.603 & 1.860 & 2.043 & 1.647 \\
                
                \multicolumn{1}{l|}{Pixel-Perfect (SuperPoint~\cite{superpoint}+NN)} &
                406.536 & 728.230 & 138.871 & 53.937 & 97.041 & \multicolumn{1}{c|}{18.185} & 220.455 & 469.903 & 107.160 & 137.272 & 279.020 & 66.045 \\

                \rowcolor{RowColor} \multicolumn{1}{l|}{$\hookrightarrow$ Fix Extrinsic} & 
                10.456 & 24.965 & 4.414 & 1.389 & 3.313 & \multicolumn{1}{c|}{0.582} & 2.686 & 4.290 & 1.883 & 1.726 & 2.521 & 1.257 \\
                
                \multicolumn{1}{l|}{Pixel-Perfect (SuperPoint+SuperGlue \cite{superglue})} &
                490.563 & 634.495 & 227.274 & 65.008 & 84.246 & \multicolumn{1}{c|}{30.119} & 207.518 & 300.653 & 142.050 & 130.619 & 190.479 & 86.182 \\

                \rowcolor{RowColor} \multicolumn{1}{l|}{$\hookrightarrow$ Fix Extrinsic} & 
                7.307 & \underline{14.157} & \underline{3.808} & 0.965 & \underline{1.873} & \multicolumn{1}{c|}{\underline{0.502}} & 2.391 & 3.534 & \textbf{1.384} & 1.563 & 2.146 & \underline{0.986} \\

                \multicolumn{1}{l|}{DUSt3R \cite{wang2024dust3r} (Linear)} &
                5854.966 & 5982.099 & 5570.567 & 774.200 & 791.681 & \multicolumn{1}{c|}{739.994} & 163.994 & 164.348 & 163.778 & 104.706 & 104.928 & 104.567 \\

                \rowcolor{RowColor} \multicolumn{1}{l|}{$\hookrightarrow$ Fix Extrinsic} & 
                5885.447 & 6087.489 & 5602.540 & 776.848 & 805.607 & \multicolumn{1}{c|}{742.758} & 156.384 & 157.863 & 154.685 & 93.320 & 94.269 & 92.218 \\

                \multicolumn{1}{l|}{DUSt3R (DPT)} &
                9959.293 & 10114.131 & 9760.731 & 1309.893 & 1329.840 & \multicolumn{1}{c|}{1283.910} & 164.321 & 164.702 & 163.716 & 104.895 & 105.115 & 104.549 \\

                \rowcolor{RowColor} \multicolumn{1}{l|}{$\hookrightarrow$ Fix Extrinsic} & 
                9878.869 & 10023.006 & 9727.524 & 1300.423 & 1318.947 & \multicolumn{1}{c|}{1280.676} & 170.561 & 171.531 & 169.178 & 104.892 & 105.471 & 104.069 \\

                \multicolumn{1}{l|}{VGGSfM \cite{wang2023visual}} &
                560.150 & 1056.412 & 413.168 & 73.191 & 137.347 & \multicolumn{1}{c|}{53.895} & 164.713 & 164.713 & 164.713 & 105.052 & 105.052 & 105.052 \\

                \rowcolor{RowColor} \multicolumn{1}{l|}{$\hookrightarrow$ Fix Extrinsic} & 
                120.997 & 131.599 & 111.526 & 15.918 & 17.316 & \multicolumn{1}{c|}{14.686} & 6.645 & 7.894 & 5.964 & 4.067 & 4.869 & 3.666 \\

                \rowcolor{gray!30} \multicolumn{1}{l|}{Ours (Single frame)} &
                \underline{6.598} & \textbf{8.916} & \textbf{2.950} & \underline{0.870} & \textbf{1.178} & \multicolumn{1}{c|}{\textbf{0.396}} & 2.227 & \underline{2.845} & 1.546 & 1.483 & \underline{1.789} & 1.106 \\

                \rowcolor{gray!45} \multicolumn{1}{l|}{Ours (Multiple frames)} &
                \textbf{5.405} & / & / & \textbf{0.712} & / & \multicolumn{1}{c|}{/} & \underline{1.994} & / & / & \underline{1.335} & / & / \\
    	    \Xhline{1pt}
    	\end{tabular}
	}
	\vspace{0.05in}
	\label{quantitative_result}
\end{table*}

\subsection{Extrinsics Regularization}
\label{sec:approach-extrinsics}
Traditional SfM methods jointly estimate extrinsics and intrinsics, but they lack mechanisms to refine intrinsics effectively when ground-truth extrinsics are known. We address this problem by introducing an extrinsics regularization term:

\begin{equation}
\label{eq:extrinsics-regularization}
\mathcal{L}_1 = \frac{\lambda_1}{N_C} \sum_{j=0}^{N_C-1} \rho\left(\left\|\mathbf{r}_{i,j} - \hat{\mathbf{r}}_j\right\|\right) + \frac{\lambda_2}{N_C} \sum_{j=0}^{N_C-1} \rho\left(\left\|\mathbf{t}_{i,j} - \hat{\mathbf{t}}_j\right\|\right)
\end{equation}
where $\hat{\mathbf{r}}_j$ and $\hat{\mathbf{t}}_j$ represent the ground-truth extrinsics of the $j$-th camera. Rather than directly substituting the estimated extrinsics with the ground-truth values, we use an iterative method to guide the estimated values toward the ground-truth values. We initialize $\lambda_1$ and $\lambda_2$ with small values in the first optimization iteration. These coefficients are gradually increased in subsequent iterations, progressively constraining extrinsics. Once $\lambda_1$ and $\lambda_2$ are sufficiently large, the estimated extrinsics converge to ground-truth values, ensuring more accurate intrinsics estimation.

This method effectively mitigates the risk of converging to local minima in BA, which is inherently a non-convex problem. Although there may be dozens of iterations depending on the termination threshold, each iteration converges rapidly in just a few steps, ensuring overall efficiency of the process.

\subsection{Dense Feature Reprojection}
\label{sec:approach-featuremetric}

Even with known ground-truth extrinsics, it is challenging to achieve intrinsics as accurate as those obtained through a dedicated calibration process. This limitation arises because the aforementioned BA process relies heavily on 2D keypoints, which are inherently noisy. The noise in keypoint detection propagates through the optimization process, leading to estimated intrinsics deviating from ground-truth values.

To address this issue, we are inspired by Pixel-Perfect SfM~\cite{pixelsfm}, which mitigates keypoint noise by performing optimization in feature space rather than relying solely on raw reprojection error. Specifically, for the $i$-th frame and $j$-th camera, we use a convolutional neural network (CNN) to compute a dense feature map $\mathbf{F}_{i,j}$ of size $H \times W \times 128$, where $H$ and $W$ are the height and width of the input image, and the feature map has $128$ channels. The CNN is trained to enforce viewpoint consistency: for a given interest point in the scene, the CNN generates identical feature vectors at its projected locations in images, regardless of the observing cameras. This property allows the reprojection error to be measured not only in image space but also in feature space, effectively reducing the influence of noisy keypoints. After extracting the dense feature map $\mathbf{F}_{i,j}$, we compute a reference feature vector $\mathbf{f}_\mathbf{x}$ for each 3D point $\mathbf{x}$. The reference feature is defined as the keypoint feature vector closest to the mean of all keypoint feature vectors within the track:

\begin{align}
    \label{eq:track-feature}
    \mathcal{F}_{\mathcal{T}_\mathbf{x}} & = \left\{
	\mathbf{F}_{i,j}( \mathbf{p} ) \mid (j,\mathbf{p}) \in \mathcal{T}_\mathbf{x}
    \right\}
    \\
    \label{eq:track-mean-feature}
    \bar{\mathbf{f}}_{\mathcal{T}_\mathbf{x}} & =
	\argmin_{
		\mathbf{f} \in \mathbb{R}^D
	}{
		\sum_{ \mathbf{f}' \in \mathcal{F}_{\mathcal{T}_\mathbf{x}} } {
			\rho(\left\| \mathbf{f} - \mathbf{f}' \right\|)
		}
	}
    \\
    \label{eq:reference-feature}
    \mathbf{f}_\mathbf{x} & =
	\argmin_{
		\mathbf{f}' \in \mathcal{F}_{\mathcal{T}_\mathbf{x}}
	} {
		\left\| \mathbf{f}' - \bar{\mathbf{f}}_{\mathcal{T}_\mathbf{x}} \right\|
	}
\end{align}
where $\mathbf{F}_{i,j}(\cdot)$ performs bicubic interpolation on the dense feature map. $\mathcal{F}_{\mathcal{T}_\mathbf{x}}$ denotes the set of feature vectors interpolated at the keypoint positions in the track. $\bar{\mathbf{f}}_{\mathcal{T}_\mathbf{x}}$ represents the robust mean of these feature vectors, computed under the loss function $\rho(\cdot)$. Finally, $\mathbf{f}_\mathbf{x}$ is defined as the reference feature, chosen as the keypoint feature vector closest to $\bar{\mathbf{f}}_{\mathcal{T}_\mathbf{x}}$.

We then incorporate a dense feature reprojection term into the loss function, which aligns dense feature representations across frames. This term penalizes discrepancies between the feature vectors at projection locations and the reference features, reducing the influence of noisy observations on intrinsics estimation:

\begin{align}
\label{eq:featuremetric-reprojection-rep}
e_{i, \mathbf{x},j} & = \mathbf{F}_{i,j}\left(\Pi(\mathbf{r}_{i,j}, \mathbf{t}_{i,j}, K_{i,j}, \mathbf{x})\right) - \mathbf{f}_\mathbf{x}
\\
\label{eq:featuremetric-reprojection}
\mathcal{L}_2 & = \frac{\lambda_3}{\sum\limits_{\mathbf{x} \in \mathcal{X}_i} | \mathcal{T}_\mathbf{x} |} 
\sum\limits_{\mathbf{x} \in \mathcal{X}_i} \sum\limits_{(j, \mathbf{p}) \in \mathcal{T}_\mathbf{x}} 
\rho\left(\left\|    e_{i, \mathbf{x},j}    \right\|\right)
\end{align}
where $\lambda_3$ is a weighting factor that balances the contribution of this term relative to others in the total loss.

In practice, it is impossible to load all dense feature maps in memory at the same time, and is very inefficient to perform bicubic interpolation $\mathbf{F}_{i,j}(\cdot)$ and compute losses $\rho(\cdot)$ in the $128$-dimensional feature space. To address this issue, inspired by Pixel-Perfect SfM~\cite{pixelsfm}, we only keep $16\times16$ patches around keypoints, and preprocess these feature patches into $3$-dimensional cost maps:

\begin{equation}
\label{eq:cost-map}
    \mathbf{G}_{i,j,\mathbf{x}}(u, v) = 
    \begin{bmatrix}
        \left\| \mathbf{F}_{i,j}(u, v) - \mathbf{f}_\mathbf{x} \right\| \\ 
        \frac {\partial \left\| \mathbf{F}_{i,j}(u, v) - \mathbf{f}_\mathbf{x} \right\|}{\partial u} \\ 
        \frac {\partial \left\| \mathbf{F}_{i,j}(u, v) - \mathbf{f}_\mathbf{x} \right\|}{\partial v} 
    \end{bmatrix}
\end{equation}

Then, we use the following equation to evaluate the new dense feature reprojection loss:

\begin{align}
\label{eq:featuremetric-reprojection-with-cost-map-rep}
e'_{i, \mathbf{x},j} & = \mathbf{G}_{i,j,\mathbf{x}}\left(\Pi(\mathbf{r}_{i,j}, \mathbf{t}_{i,j}, K_{i,j}, \mathbf{x})\right)
\\
\label{eq:featuremetric-reprojection-with-cost-map}
\mathcal{L}_2 & = \frac{\lambda_3}{\sum\limits_{\mathbf{x} \in \mathcal{X}_i} | \mathcal{T}_\mathbf{x} |} 
\sum\limits_{\mathbf{x} \in \mathcal{X}_i} \sum\limits_{(j, \mathbf{p}) \in \mathcal{T}_\mathbf{x}} 
\rho\left(\left\|    e'_{i, \mathbf{x},j}    \right\|\right)
\end{align}

This helps us improve memory and computational usage.

\begin{table*}[htbp]
\renewcommand{\arraystretch}{1.2}
	\centering
        \vspace{1.5mm}
        \caption{Ablation study evaluating the impact of different module compositions on multiple metrics.}
	\scalebox{0.68}{
    	\begin{tabular}{cccccccccccccc}
    		\Xhline{1pt}
    		\multicolumn{1}{c|}{\multirow{2}{*}{\textbf{No.}}} & \multicolumn{1}{c|}{\multirow{2}{*}{Reprojection}} & \multicolumn{1}{c|}{Extrinsics} & \multicolumn{1}{c|}{Progressive} & \multicolumn{1}{c|}{Dense Feature} & \multicolumn{1}{c|}{Intrinsics} & \multicolumn{8}{c}{\textbf{Multiface}} \\
    		\multicolumn{1}{c|}{} & \multicolumn{1}{c|}{} & \multicolumn{1}{c|}{Regularization} & \multicolumn{1}{c|}{Coefficient} & \multicolumn{1}{c|}{Reprojection} & \multicolumn{1}{c|}{Variance} & $\mathrm{focal}_\mathrm{abs,mean}$ & $\mathrm{focal}_\mathrm{abs,max}$ & $\mathrm{focal}_\mathrm{abs,min}$ & \multicolumn{1}{c|}{$\mathrm{focal}_\mathrm{rel,mean}$\textperthousand} & $\mathrm{pp}_\mathrm{abs, mean}$ & $\mathrm{pp}_\mathrm{abs,max}$ & $\mathrm{pp}_\mathrm{abs,min}$ & $\mathrm{pp}_\mathrm{rel,mean}$\textperthousand \\ 
    		\Xhline{1pt}
            
    		\rowcolor{RowColor} \multicolumn{1}{c|}{(1)} & \cmark & \xmark & \xmark & \xmark & \multicolumn{1}{c|}{\xmark}  & 7.307 & 14.157 & 3.808 & \multicolumn{1}{c|}{0.965} & 2.391 & 3.534 & 1.384 & 1.563 \\
            
                \multicolumn{1}{c|}{(2)} & \cmark & \cmark & \xmark & \xmark & \multicolumn{1}{c|}{\xmark}  & 9.318 & 18.778 & 4.815 & \multicolumn{1}{c|}{1.229} & \underline{2.200} & 3.175 & \underline{1.176} & \underline{1.437} \\
                
                \rowcolor{RowColor} \multicolumn{1}{c|}{(3)} & \cmark & \cmark & \cmark & \xmark & \multicolumn{1}{c|}{\xmark}  & 6.616 & \underline{8.924} & \underline{2.991} & \multicolumn{1}{c|}{0.873} & 2.241 & \underline{2.859} & 1.548 & 1.492 \\
                
                \multicolumn{1}{c|}{(4)} & \cmark & \xmark & \xmark & \cmark & \multicolumn{1}{c|}{\xmark}  & 7.371 & 13.904 & 3.645 & \multicolumn{1}{c|}{0.975} & 2.407 & 3.431 & 1.455 & 1.574 \\
                
                \rowcolor{RowColor} \multicolumn{1}{c|}{(5)} & \cmark & \cmark & \xmark & \cmark & \multicolumn{1}{c|}{\xmark}  & 9.247 & 18.856 & 4.647 & \multicolumn{1}{c|}{1.220} & 2.203 & 3.176 & \textbf{1.175} & 1.440 \\
                
                \multicolumn{1}{c|}{(6)} & \cmark & \cmark & \cmark & \cmark & \multicolumn{1}{c|}{\xmark}  & \underline{6.598} & \textbf{8.916} & \textbf{2.950} & \multicolumn{1}{c|}{\underline{0.870}} & 2.227 & \textbf{2.845} & 1.546 & 1.483 \\

                \rowcolor{RowColor} \multicolumn{1}{c|}{(7)} & \cmark & \cmark & \cmark & \cmark & \multicolumn{1}{c|}{\cmark}  & \textbf{5.405} & / & / & \multicolumn{1}{c|}{\textbf{0.712}} & \textbf{1.994} & / & / & \textbf{1.335} \\
                
    	    \Xhline{1pt}
    	\end{tabular}
	}
	\vspace{0.05in}
	\label{ablation_result}
\end{table*}

\begin{figure*}[ht]
    \centering
    \rotatebox[origin=l]{90}{\makebox[0.05\textwidth][c]{\scriptsize COLMAP}}
    \hfill
    \begin{subfigure}[b]{0.135\textwidth}
        \centering
        \includegraphics[width=\textwidth]{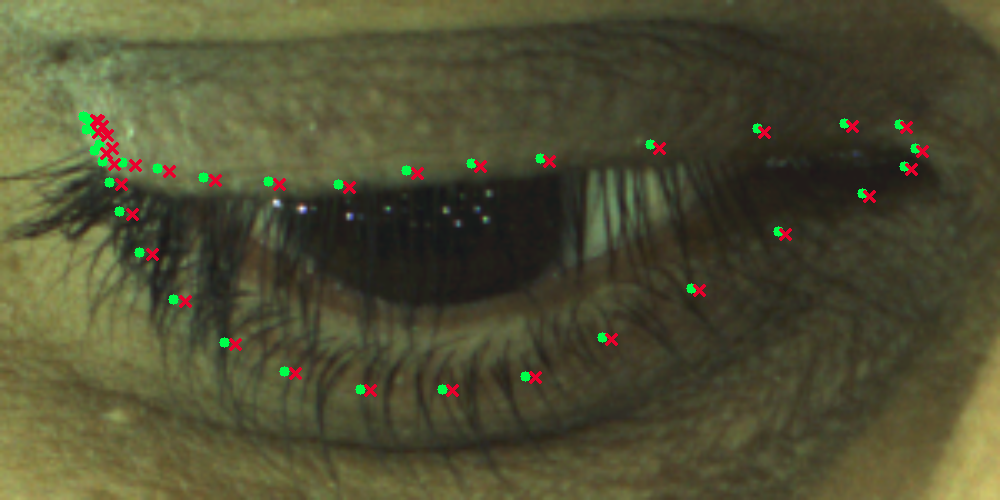}
    \end{subfigure}
    \begin{subfigure}[b]{0.135\textwidth}
        \centering
        \includegraphics[width=\textwidth]{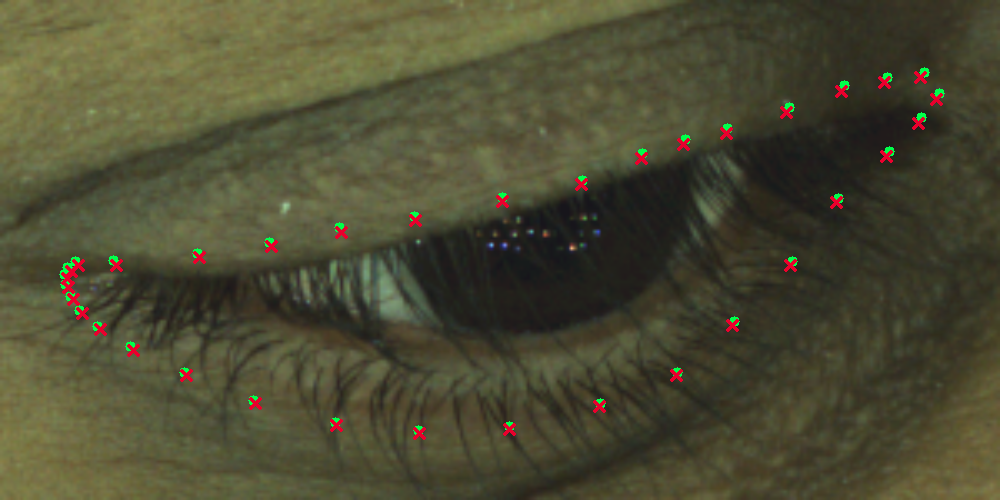}
    \end{subfigure}
    \begin{subfigure}[b]{0.135\textwidth}
        \centering
        \includegraphics[width=\textwidth]{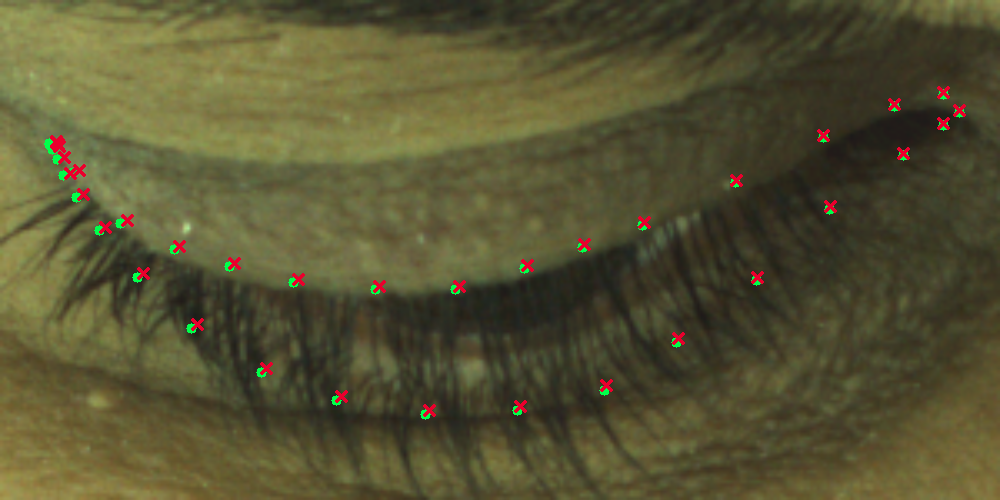}
    \end{subfigure}
    \begin{subfigure}[b]{0.135\textwidth}
        \centering
        \includegraphics[width=\textwidth]{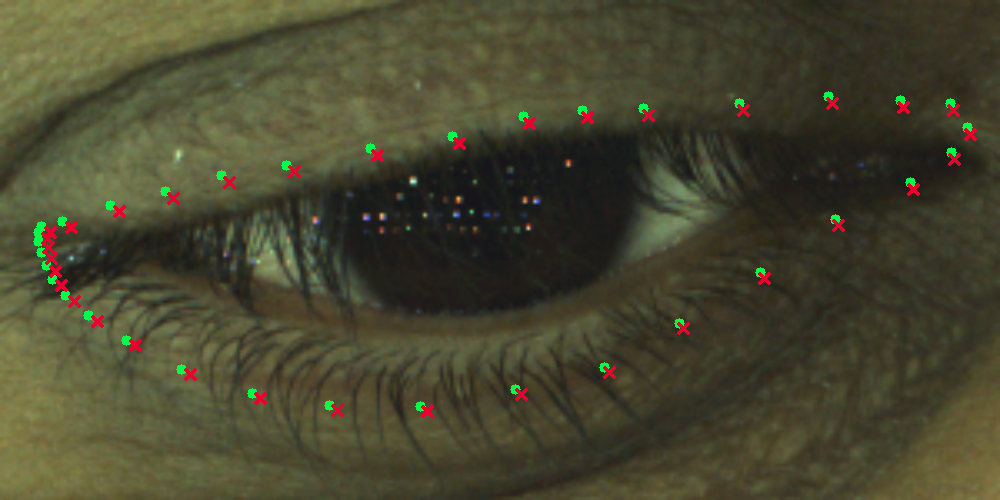}
    \end{subfigure}
    \begin{subfigure}[b]{0.135\textwidth}
        \centering
        \includegraphics[width=\textwidth]{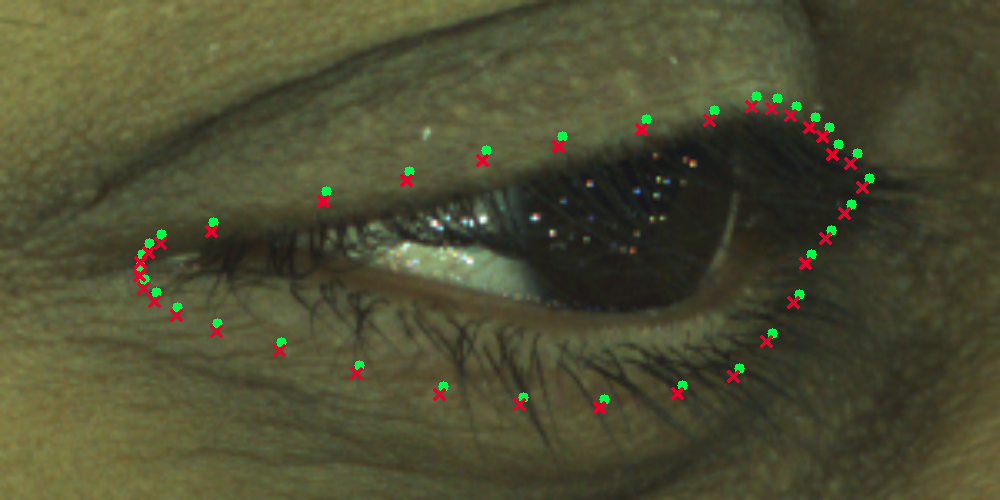}
    \end{subfigure}
    \begin{subfigure}[b]{0.135\textwidth}
        \centering
        \includegraphics[width=\textwidth]{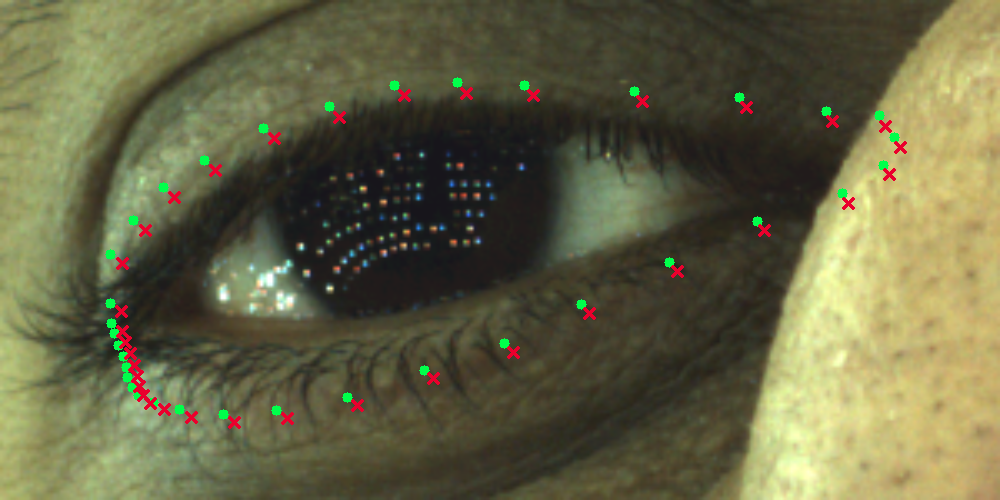}
    \end{subfigure}
    \begin{subfigure}[b]{0.135\textwidth}
        \centering
        \includegraphics[width=\textwidth]{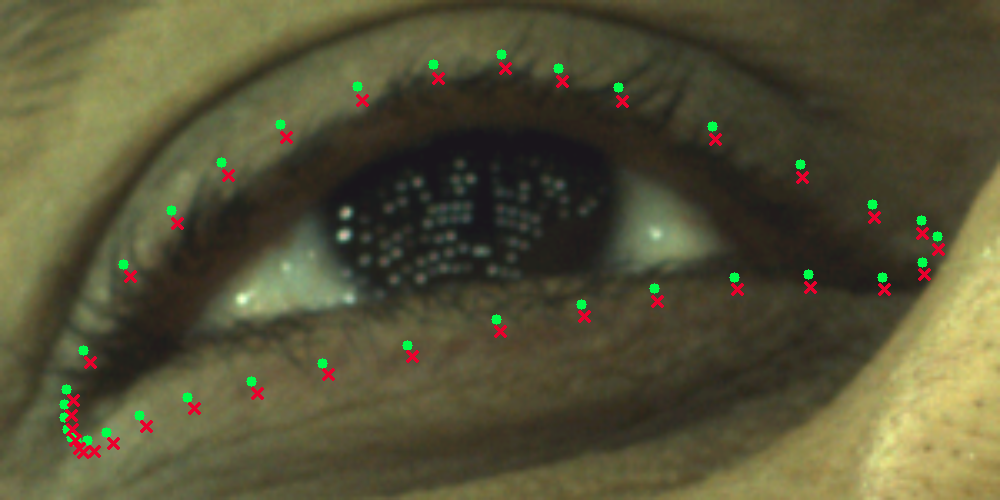}
    \end{subfigure}


    \rotatebox[origin=l]{90}{\makebox[0.05\textwidth][c]{\scriptsize PixelSfM}}
    \hfill
    \begin{subfigure}[b]{0.135\textwidth}
        \centering
        \includegraphics[width=\textwidth]{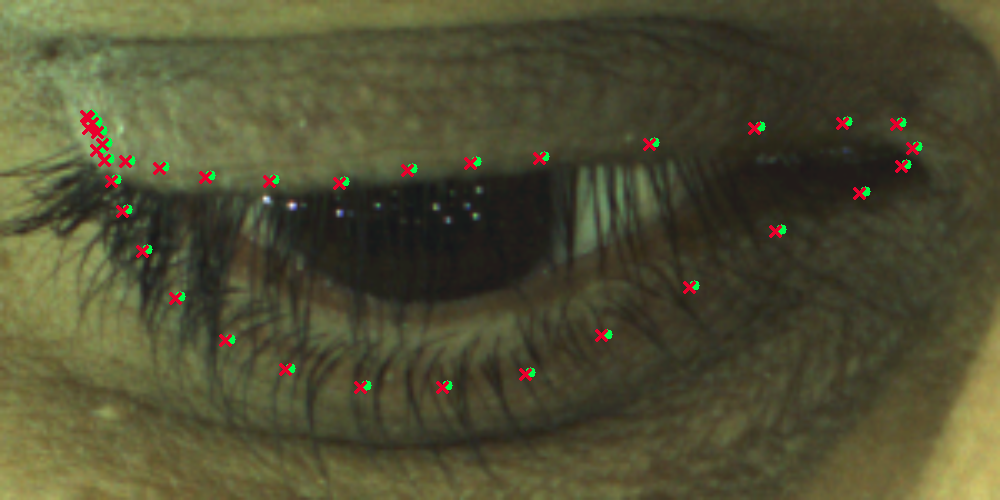}
    \end{subfigure}
    \begin{subfigure}[b]{0.135\textwidth}
        \centering
        \includegraphics[width=\textwidth]{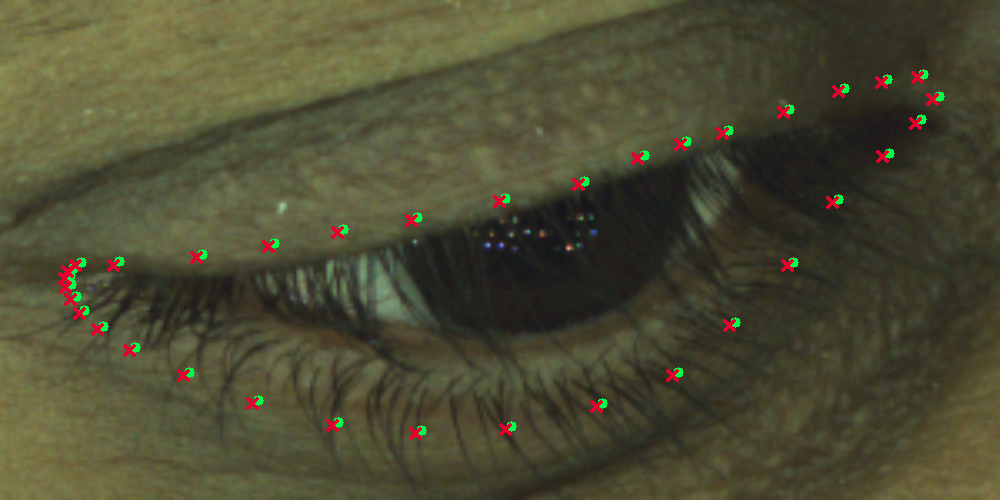}
    \end{subfigure}
    \begin{subfigure}[b]{0.135\textwidth}
        \centering
        \includegraphics[width=\textwidth]{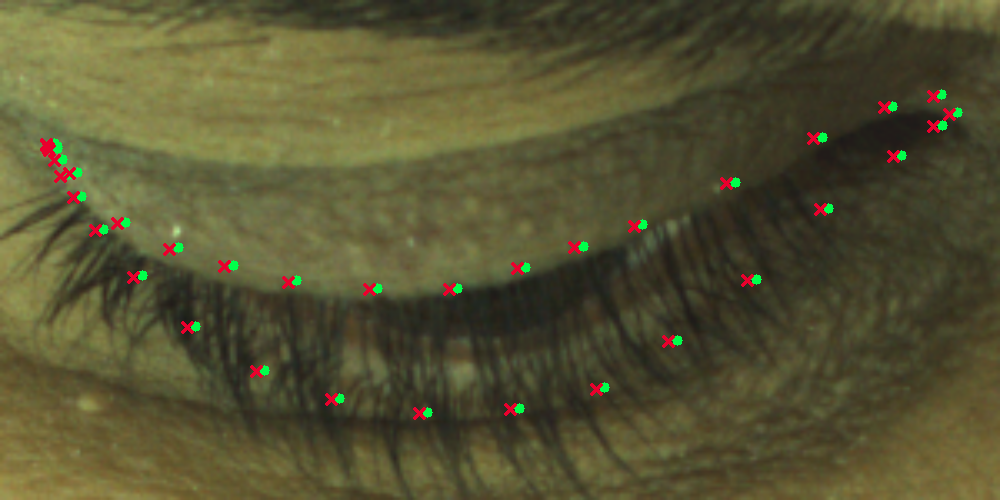}
    \end{subfigure}
    \begin{subfigure}[b]{0.135\textwidth}
        \centering
        \includegraphics[width=\textwidth]{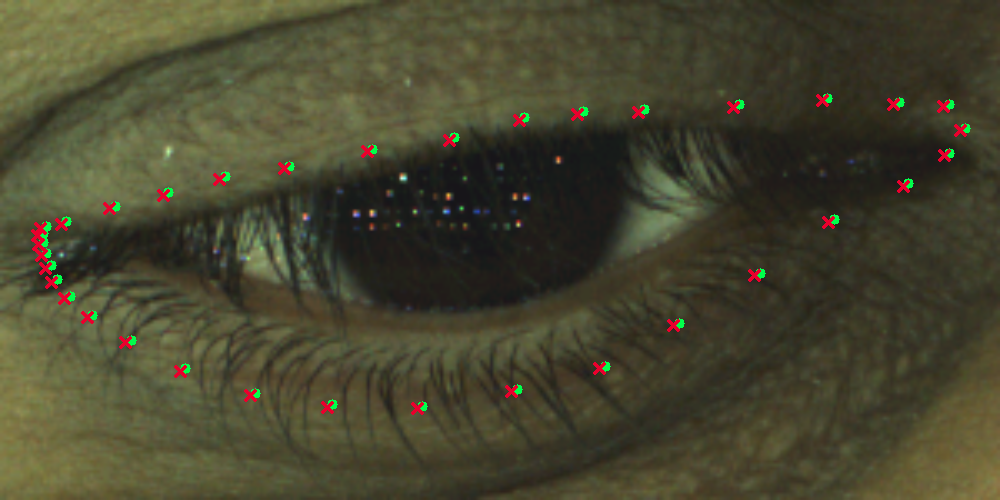}
    \end{subfigure}
    \begin{subfigure}[b]{0.135\textwidth}
        \centering
        \includegraphics[width=\textwidth]{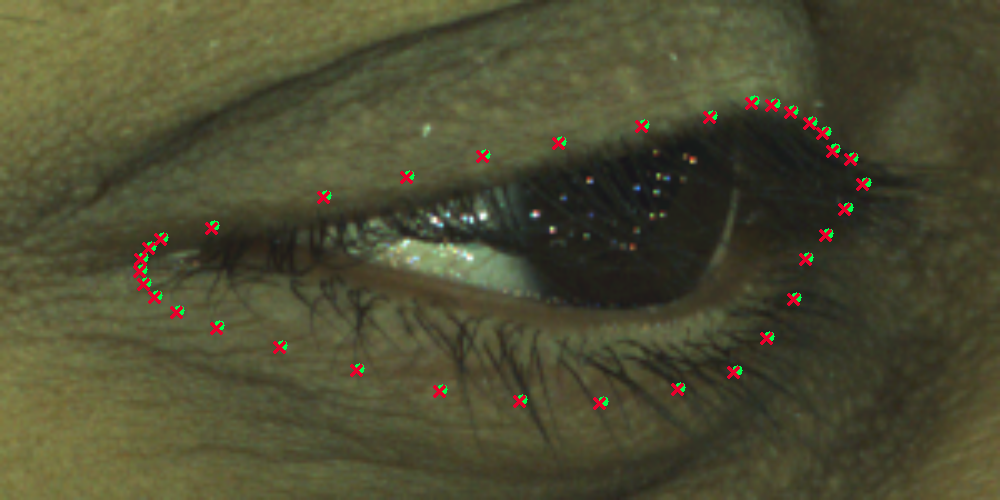}
    \end{subfigure}
    \begin{subfigure}[b]{0.135\textwidth}
        \centering
        \includegraphics[width=\textwidth]{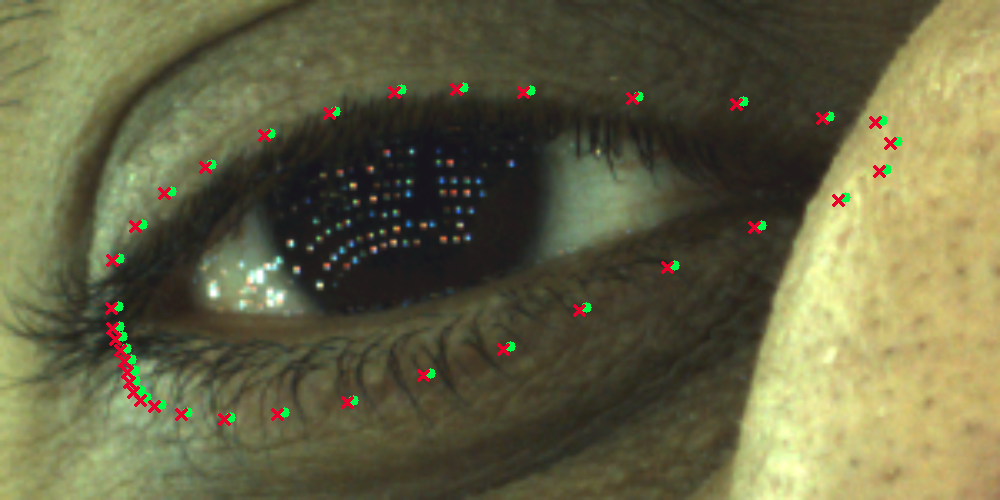}
    \end{subfigure}
    \begin{subfigure}[b]{0.135\textwidth}
        \centering
        \includegraphics[width=\textwidth]{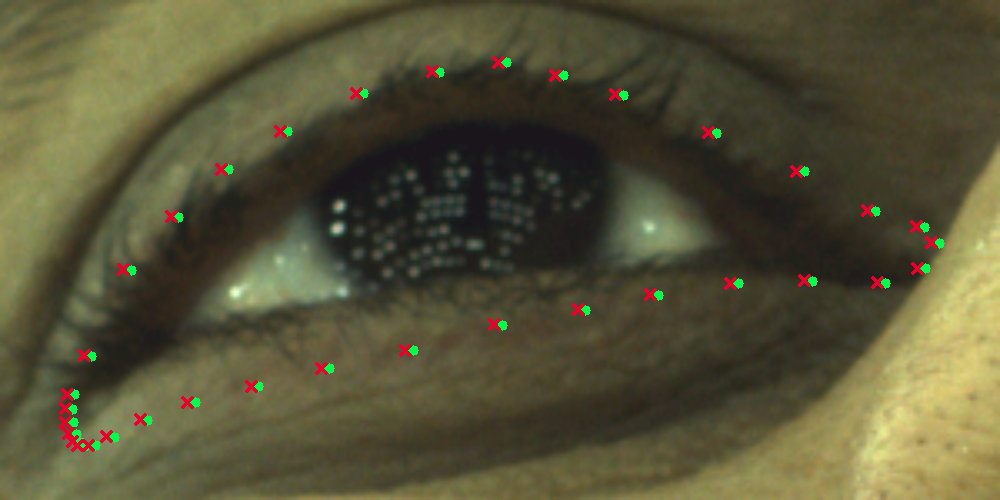}
    \end{subfigure}


    \rotatebox[origin=l]{90}{\makebox[0.05\textwidth][c]{\scriptsize VGGSfM}}
    \hfill
    \begin{subfigure}[b]{0.135\textwidth}
        \centering
        \includegraphics[width=\textwidth]{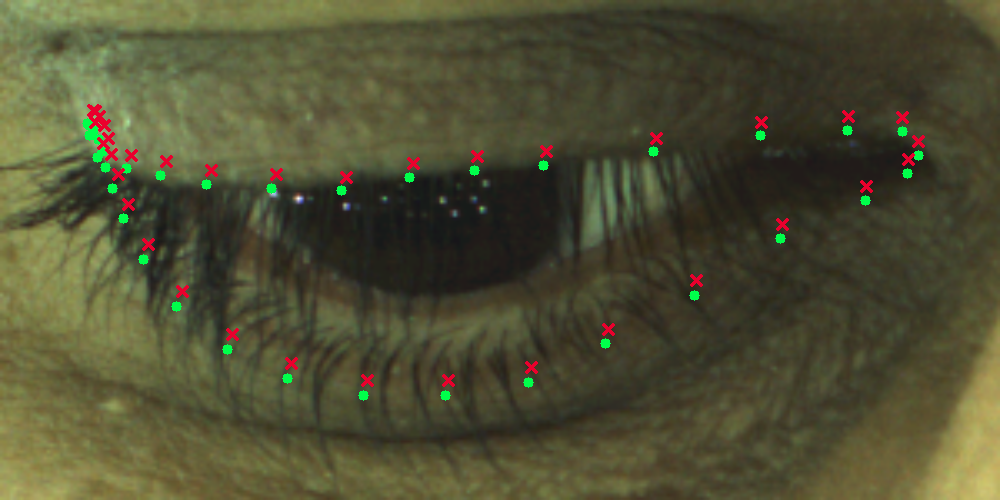}
    \end{subfigure}
    \begin{subfigure}[b]{0.135\textwidth}
        \centering
        \includegraphics[width=\textwidth]{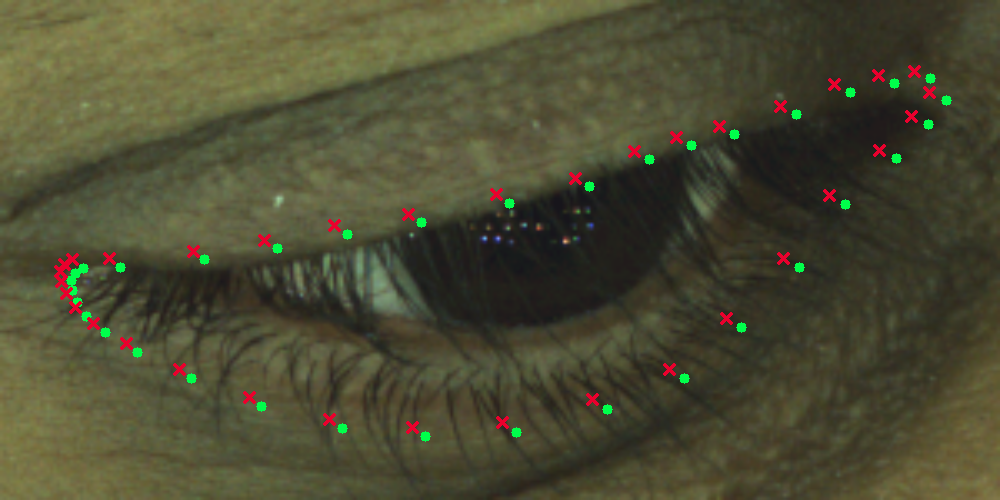}
    \end{subfigure}
    \begin{subfigure}[b]{0.135\textwidth}
        \centering
        \includegraphics[width=\textwidth]{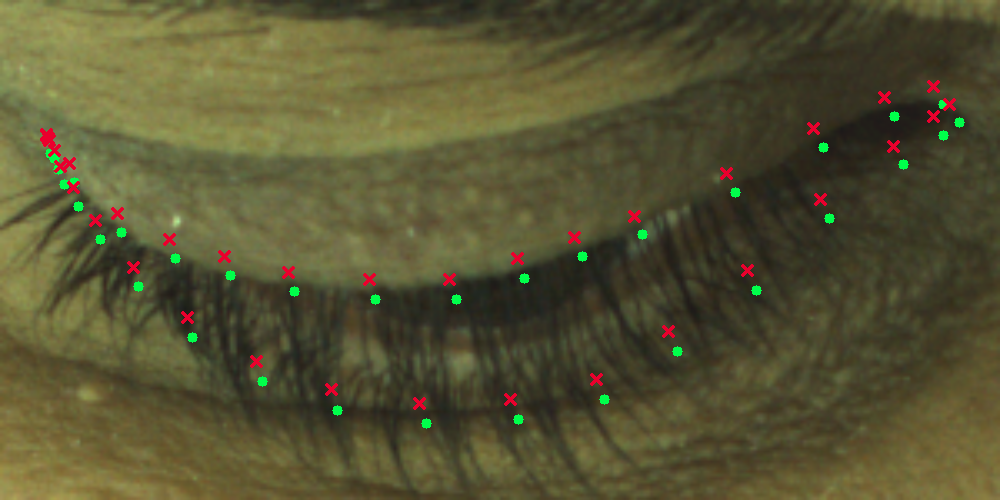}
    \end{subfigure}
    \begin{subfigure}[b]{0.135\textwidth}
        \centering
        \includegraphics[width=\textwidth]{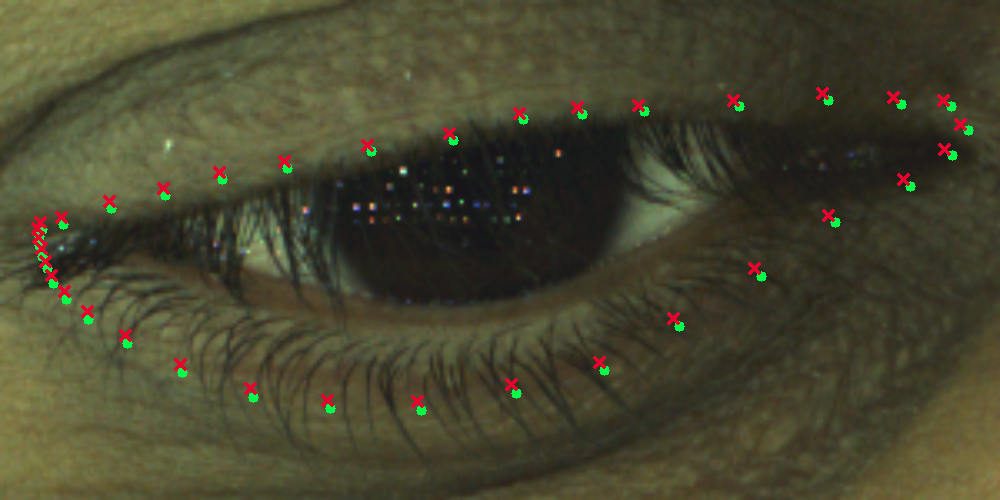}
    \end{subfigure}
    \begin{subfigure}[b]{0.135\textwidth}
        \centering
        \includegraphics[width=\textwidth]{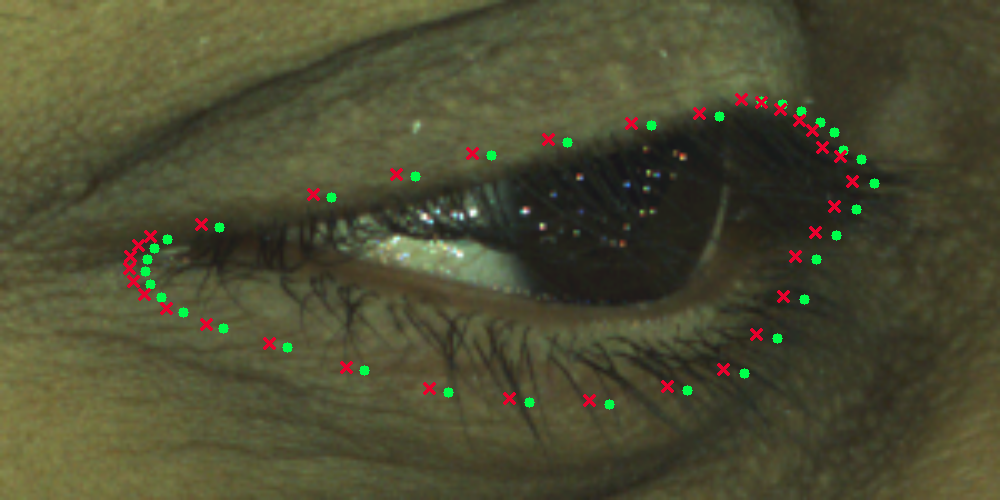}
    \end{subfigure}
    \begin{subfigure}[b]{0.135\textwidth}
        \centering
        \includegraphics[width=\textwidth]{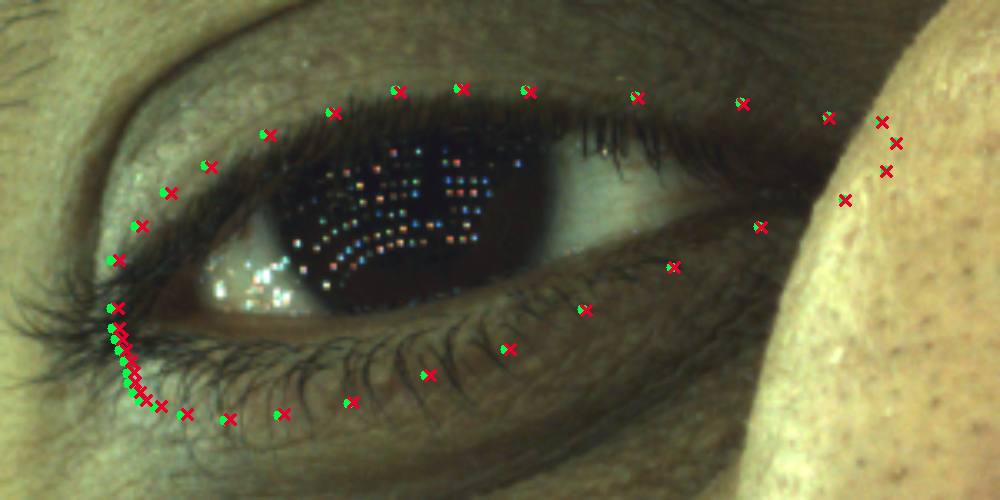}
    \end{subfigure}
    \begin{subfigure}[b]{0.135\textwidth}
        \centering
        \includegraphics[width=\textwidth]{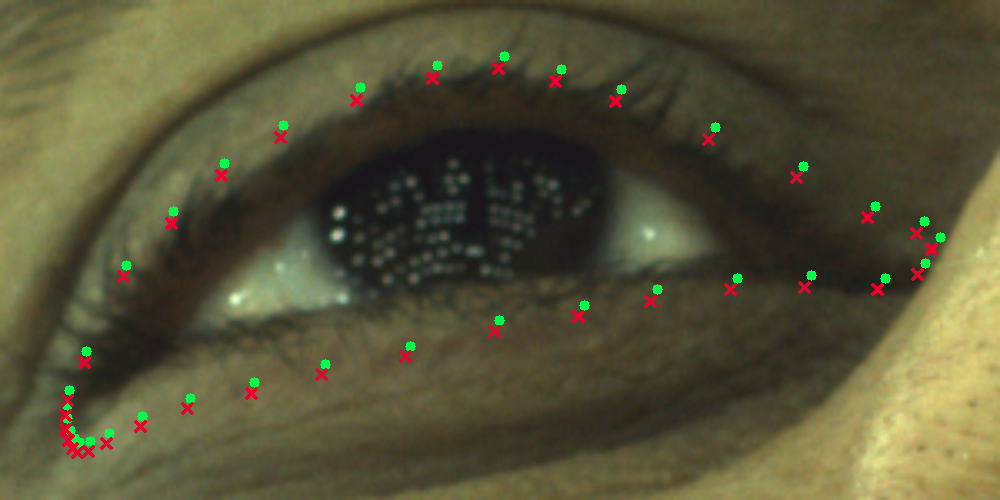}
    \end{subfigure}


    \rotatebox[origin=l]{90}{\makebox[0.05\textwidth][c]{\scriptsize Ours}}
    \hfill
    \begin{subfigure}[b]{0.135\textwidth}
        \centering
        \includegraphics[width=\textwidth]{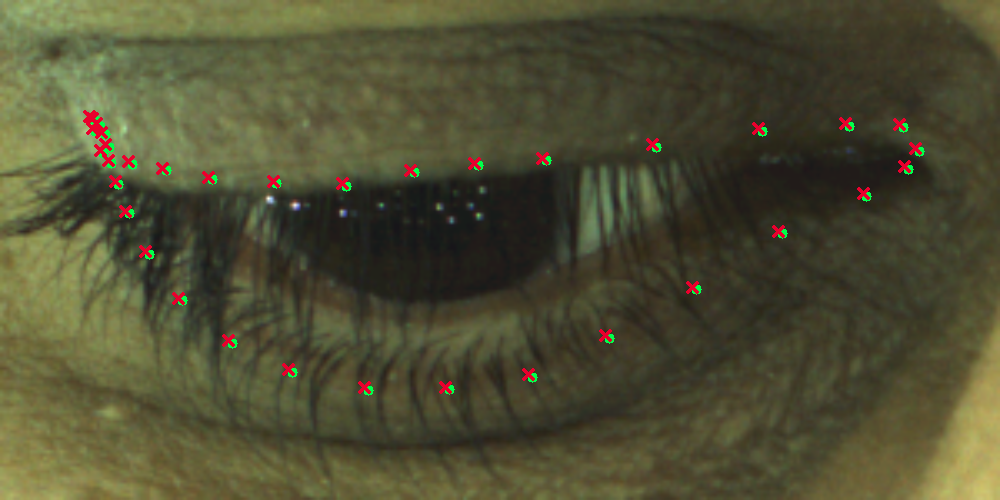}
    \end{subfigure}
    \begin{subfigure}[b]{0.135\textwidth}
        \centering
        \includegraphics[width=\textwidth]{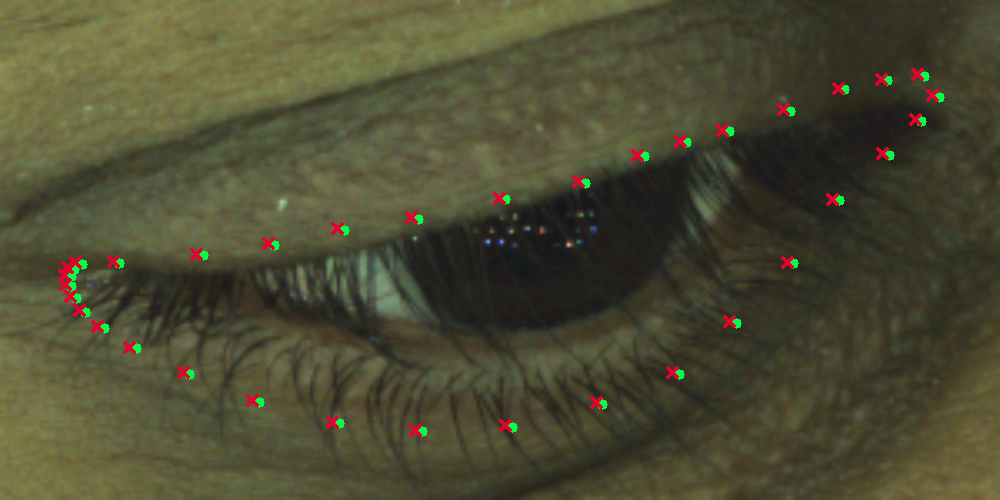}
    \end{subfigure}
    \begin{subfigure}[b]{0.135\textwidth}
        \centering
        \includegraphics[width=\textwidth]{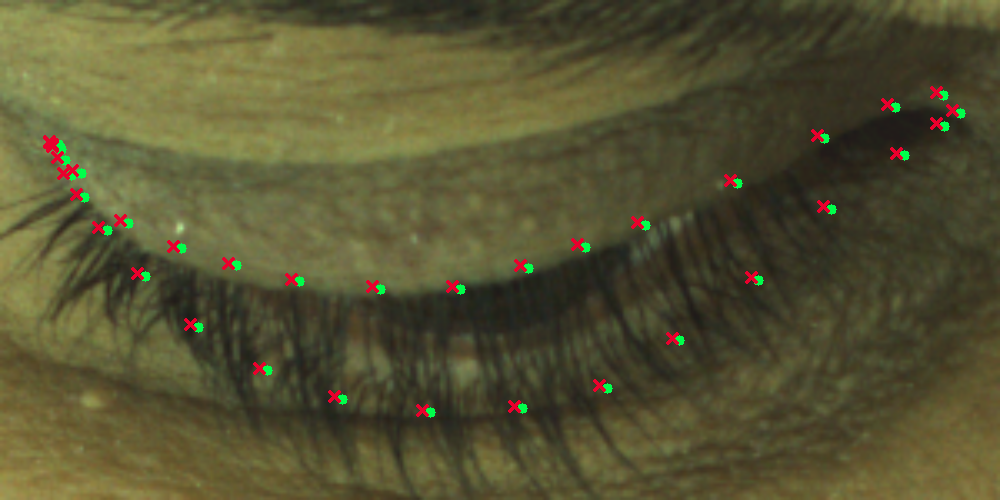}
    \end{subfigure}
    \begin{subfigure}[b]{0.135\textwidth}
        \centering
        \includegraphics[width=\textwidth]{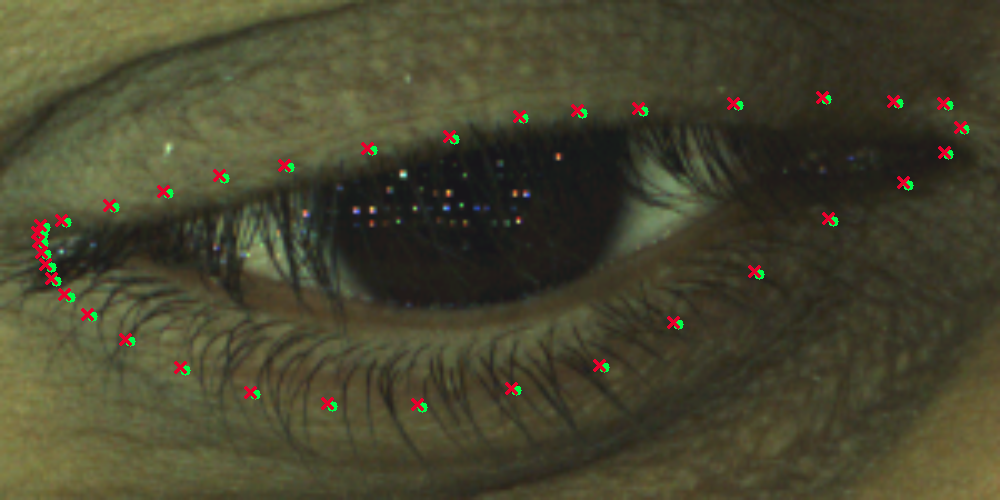}
    \end{subfigure}
    \begin{subfigure}[b]{0.135\textwidth}
        \centering
        \includegraphics[width=\textwidth]{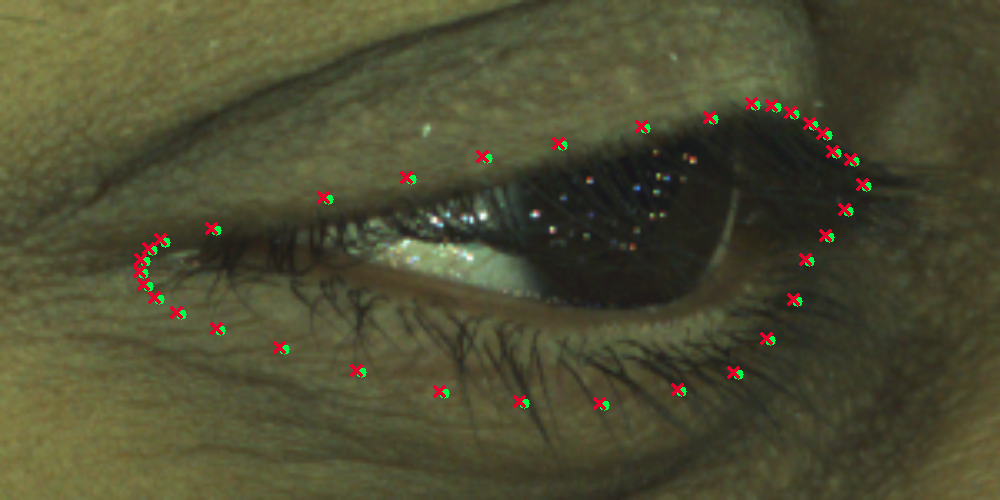}
    \end{subfigure}
    \begin{subfigure}[b]{0.135\textwidth}
        \centering
        \includegraphics[width=\textwidth]{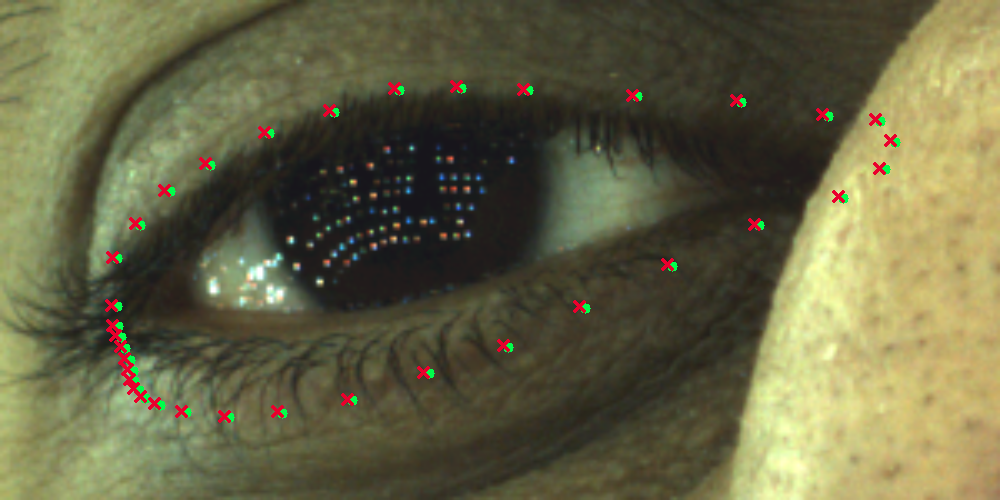}
    \end{subfigure}
    \begin{subfigure}[b]{0.135\textwidth}
        \centering
        \includegraphics[width=\textwidth]{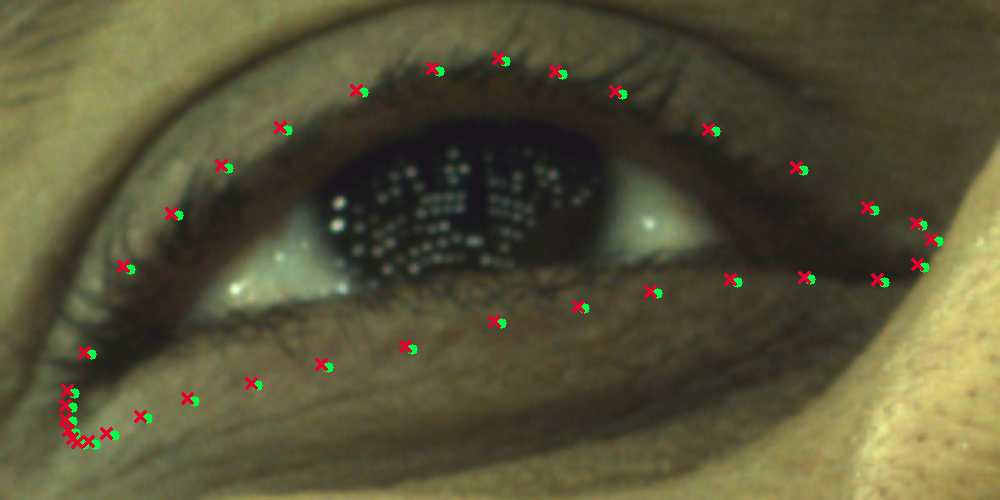}
    \end{subfigure}

    \caption{Visualization of reprojection error between COLMAP (1\textsuperscript{st} row), Pixel-Perfect SfM (2\textsuperscript{nd} row) and VGGSfM (3\textsuperscript{rd} row), and our method (4\textsuperscript{th} row). We sample a few 3D points on the ground-truth mesh and project them to the images using ground-truth (red crossings) and estimated (green dots) intrinsics. Our method yields the lowest reprojection error.}

    \label{fig:reprojection-vis}
\end{figure*}

\subsection{Intrinsics Variance}
\label{sec:approach-intrinsics-variance}

Traditional SfM pipelines are designed to process single-frame data independently. When applied to a dataset with multiple frames, these methods are executed separately for each frame, resulting in multiple sets of camera parameters, which causes inconsistencies, as the same physical camera should ideally have the same intrinsics across all frames.

To address this issue, we propose using global intrinsics, denoted as $\bar{K}_j$, for each camera. These parameters are shared across all frames and remain independent of the frame index. To enforce consistency, we define an intrinsics variance term that penalizes deviations between frame-specific intrinsics $K_{i, j}$ estimated by SfM and global intrinsics $\bar{K}_j$:

\begin{equation}
    \label{eq:intrinsics-variance}
    \begin{aligned}
    \mathcal{L}_3 & = 
	\frac{\lambda_4}{N_C N_F}
	\sum_{j = 0}^{N_C-1} {
		\sum_{i = 0}^{N_F-1} {
			\rho\left(\left\|
				\begin{pmatrix}f_{x,i,j} \\ f_{y,i,j}\end{pmatrix} - \begin{pmatrix}\bar{f}_{x,j} \\ \bar{f}_{y,j}\end{pmatrix}
			\right\|\right)
		}
	}
        \\
	& +
	\frac{\lambda_5}{N_C N_F}
	\sum_{j = 0}^{N_C-1} {
		\sum_{i = 0}^{N_F-1} {
			\rho\left(\left\|
				\begin{pmatrix}c_{x,i,j} \\ c_{y,i,j}\end{pmatrix} - \begin{pmatrix}\bar{c}_{x,j} \\ \bar{c}_{y,j}\end{pmatrix}
			\right\|\right)
		}
	}
    \end{aligned}
\end{equation}
where $f_{x,i,j}, f_{y,i,j}, c_{x,i,j}, c_{y,i,j}$ are the estimated focal length and principal point parameters of the $j$-th camera based on the $i$-th frame data, and $\bar{f}_{x,j}, \bar{f}_{y,j}, \bar{c}_{x,j}, \bar{c}_{y,j}$ are the global intrinsic parameters in $\bar{K}_j$. The coefficients $\lambda_4$ and $\lambda_5$ control the regularization strength, accounting for the different magnitudes of focal lengths and principal points. Similar to the extrinsics regularization term, $\lambda_4$ and $\lambda_5$ are progressively increased during optimization. This iterative strategy ensures that the solution remains flexible during the initial stages to avoid convergence to local minima. In the later stages, the frame-specific intrinsics gradually align with the frame-independent global intrinsics, resulting in consistent intrinsics estimation across all frames.

Finally, the overall objective function is defined as:

\begin{equation}
    \label{eq:objective-function}
    \begin{aligned}
    \mathcal{L}_{\mathrm{final}} = \mathcal{L}_3 + \frac{1}{N_F} \sum_{i=0}^{N_F-1}{ \left( \mathcal{L}_0 + \mathcal{L}_1 + \mathcal{L}_2 \right) }
    \end{aligned}
\end{equation}
\vspace{-0.2in}

\subsection{Implementation}
\label{sec:approach-implementation}

Our implementation is designed as a plug-and-play add-on to existing SfM pipelines, enhancing them with intrinsic refinement. It processes sparse models in the COLMAP~\cite{sfm-revisited, pixel-view-select-mvs} format and performs refinement using the proposed approach. Dense feature maps are extracted using S2DNet~\cite{s2dnet}. For reference feature computation, we use the iteratively reweighted least squares (IRLS)~\cite{irls} method to calculate the robust mean of feature vectors. This process is parallelized using CUDA kernels, enabling efficient computation on large-scale datasets. Similarly, the cost map computation is fully implemented in CUDA to ensure high efficiency and scalability for large inputs. To robustly handle outliers during optimization, we employ a Cauchy loss function $\rho(\cdot)$ with a scale factor of 0.25.

In our implementation, we use $\lambda_0 = 1.0,\lambda_1=\lambda_2=0.01,\lambda_3=0.01,\lambda_4=\lambda_5=0.02$ in the initialization stage. During optimization, $\lambda_1,\lambda_2, \lambda_4, \lambda_5$ are multiplied with a scale factor of $2.0$ after each iteration. The whole process terminates when $\lambda_1$ exceeds the threshold $\theta=1 \times 10^6$. The overall optimization is performed using the Ceres Solver~\cite{ceres}, which offers robust and efficient performance for large-scale non-linear optimization tasks.

\section{EXPERIMENTS}

\label{sec:experiments}

\subsection{Metrics}

\label{sec:experiment-metrics}

We evaluate the accuracy of a camera's intrinsics using absolute and relative L1 errors of focal lengths and principal points. For a single-frame method, it will be executed once for each frame, resulting one set of camera intrinsic parameters for each frame. Assume $\hat{f}_{x,j}, \hat{f}_{y,j}, \hat{c}_{x,j}, \hat{c}_{y,j}$ are the ground-truth intrinsic parameters of the $j$-th camera. The errors of this frame are computed as the mean of all cameras' errors in this frame:

\begin{align}
   \label{eq:abs-focal}
   \text{focal}_\text{abs}^i & = \frac{1}{N_C}\sum_{j=0}^{N_C-1}{ | f_{x,i,j} - \hat{f}_{x,j}| + | f_{y,i,j} - \hat{f}_{y,j}| }
   \\
   \label{eq:rel-focal}
   \text{focal}_\text{rel}^i & = \frac{1}{N_C}\sum_{j=0}^{N_C-1}{ \frac{| f_{x,i,j} - \hat{f}_{x,j}|}{\hat{f}_{x,j}} + \frac{| f_{y,i,j} - \hat{f}_{y,j}|}{\hat{f}_{y,j}} }
   \\
   \label{eq:abs-pp}
   \text{pp}_\text{abs}^i & = \frac{1}{N_C}\sum_{j=0}^{N_C-1}{ | c_{x,i,j} - \hat{c}_{x,j}| + | c_{y,i,j} - \hat{c}_{y,j}| }
   \\
   \label{eq:rel-pp}
   \text{pp}_\text{rel}^i & = \frac{1}{N_C}\sum_{j=0}^{N_C-1}{ \frac{| c_{x,i,j} - \hat{c}_{x,j}|}{w} + \frac{| c_{y,i,j} - \hat{c}_{y,j}|}{h} }
\end{align}

Then we compute the mean, maximum, and minimum errors over all frames to capture result variability.

For our multi-frame method, there is only one global set of camera intrinsics $\bar{f}_{x,j}, \bar{f}_{y,j}, \bar{c}_{x,j}, \bar{c}_{y,j}$ which are independent of the frame index $i$. We then compute the errors as follows:

\begin{align}
   \label{eq:abs-focal-multiframe}
   \text{focal}_\text{abs,mean} & = \frac{1}{N_C}\sum_{j=0}^{N_C-1}{ | \bar{f}_{x,j} - \hat{f}_{x,j}| + | \bar{f}_{y,j} - \hat{f}_{y,j}| }
   \\
   \label{eq:rel-focal-multiframe}
   \text{focal}_\text{rel,mean} & = \frac{1}{N_C}\sum_{j=0}^{N_C-1}{ \frac{| \bar{f}_{x,j} - \hat{f}_{x,j}|}{\hat{f}_{x,j}} + \frac{| \bar{f}_{y,j} - \hat{f}_{y,j}|}{\hat{f}_{y,j}} }
   \\
   \label{eq:abs-pp-multiframe}
   \text{pp}_\text{abs,mean} & = \frac{1}{N_C}\sum_{j=0}^{N_C-1}{ | \bar{c}_{x,j} - \hat{c}_{x,j}| + | \bar{c}_{y,j} - \hat{c}_{y,j}| }
   \\
   \label{eq:rel-pp-multiframe}
   \text{pp}_\text{rel,mean} & = \frac{1}{N_C}\sum_{j=0}^{N_C-1}{ \frac{| \bar{c}_{x,j} - \hat{c}_{x,j}|}{w} + \frac{| \bar{c}_{y,j} - \hat{c}_{y,j}|}{h} }
\end{align}

There are no maximum or minimum errors as the results of our multi-frame method are independent of frame indices. We place slash symbols `` / " in the corresponding table fields.

\begin{figure*}[ht]
    \vspace{1.5mm}
    \centering
    \rotatebox[origin=l]{90}{\makebox[0.09\textwidth][c]{\footnotesize COLMAP}}
    \hfill
    \begin{subfigure}[b]{0.09\textwidth}
        \centering
        \includegraphics[width=\textwidth]{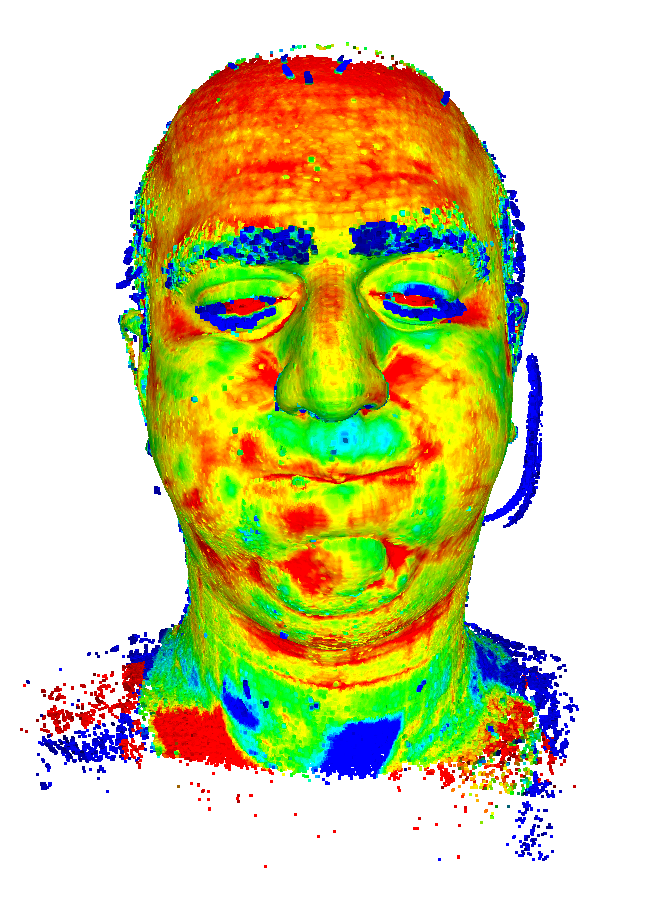}
    \end{subfigure}
    \hfill
    \begin{subfigure}[b]{0.09\textwidth}
        \centering
        \includegraphics[width=\textwidth]{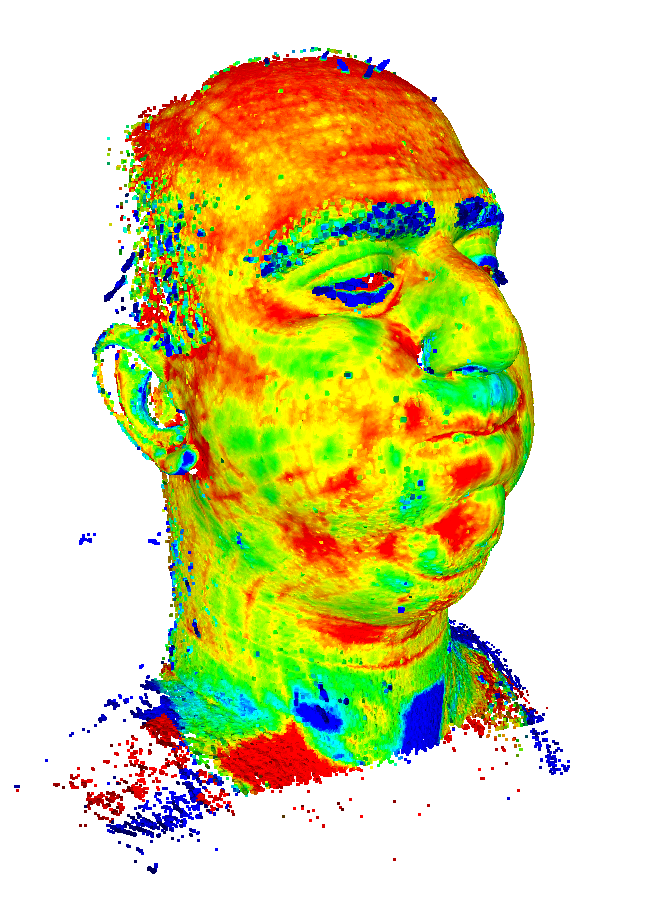}
    \end{subfigure}
    \hfill
    \begin{subfigure}[b]{0.09\textwidth}
        \centering
        \includegraphics[width=\textwidth]{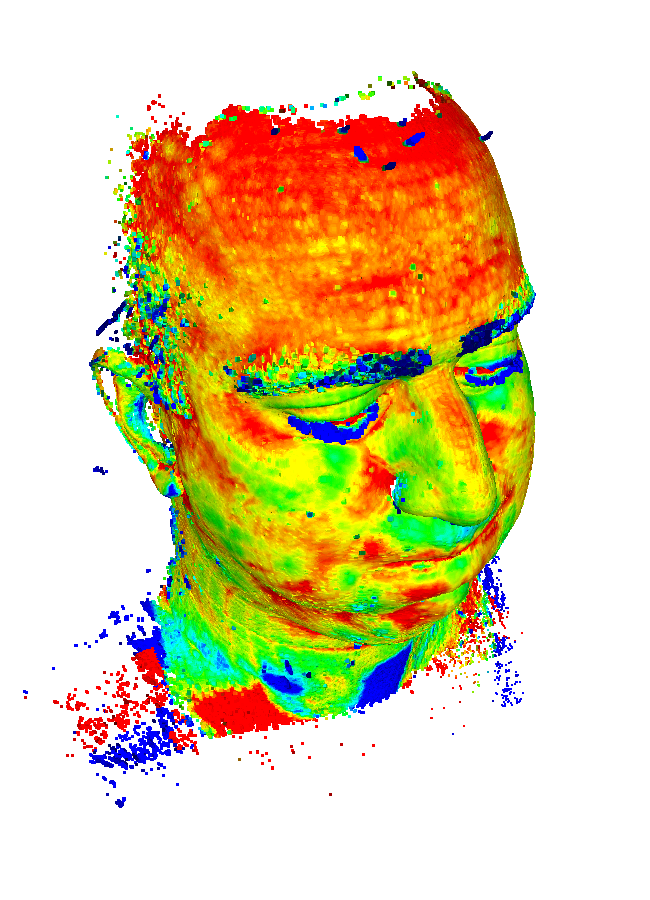}
    \end{subfigure}
    \hfill
    \begin{subfigure}[b]{0.09\textwidth}
        \centering
        \includegraphics[width=\textwidth]{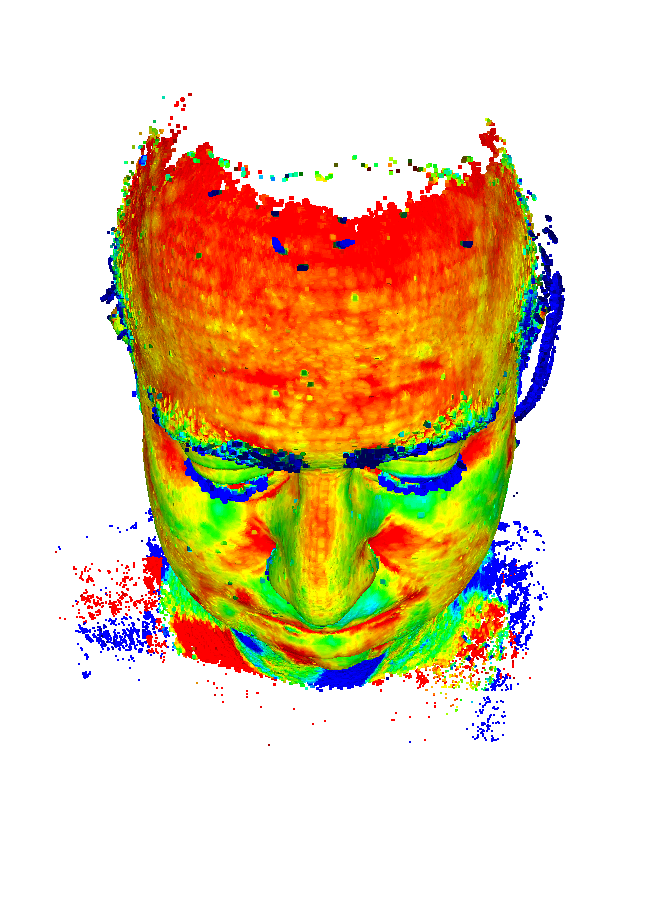}
    \end{subfigure}
    \hfill
    \begin{subfigure}[b]{0.09\textwidth}
        \centering
        \includegraphics[width=\textwidth]{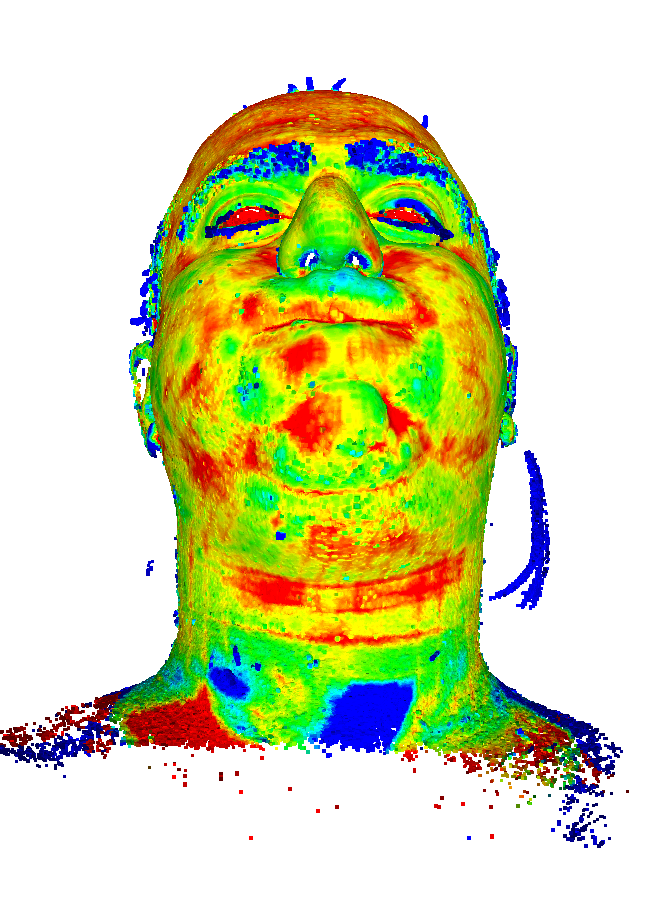}
    \end{subfigure}
    \hfill
    \begin{subfigure}[b]{0.09\textwidth}
        \centering
        \includegraphics[width=\textwidth]{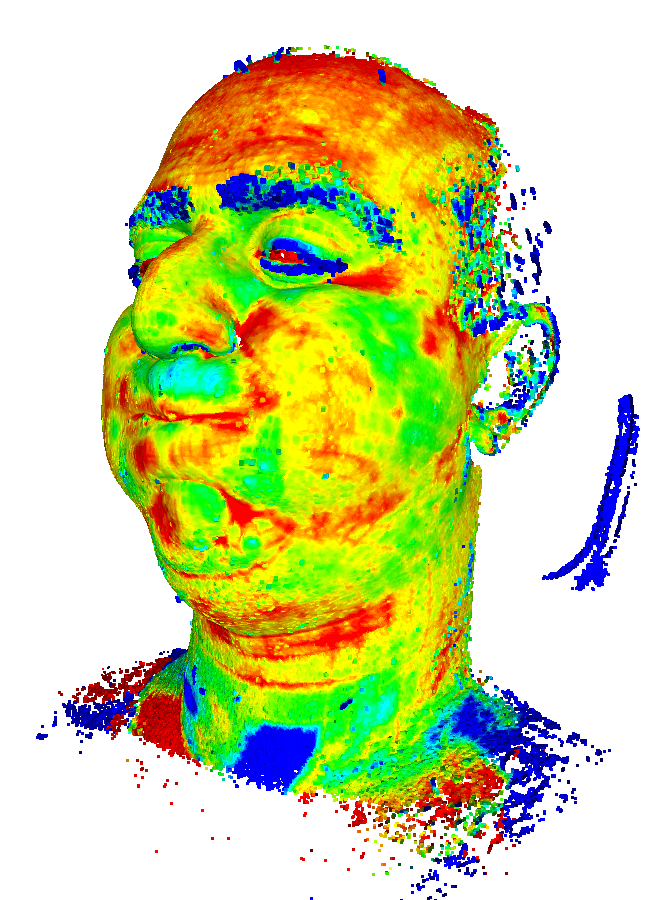}
    \end{subfigure}
    \hfill
    \begin{subfigure}[b]{0.09\textwidth}
        \centering
        \includegraphics[width=\textwidth]{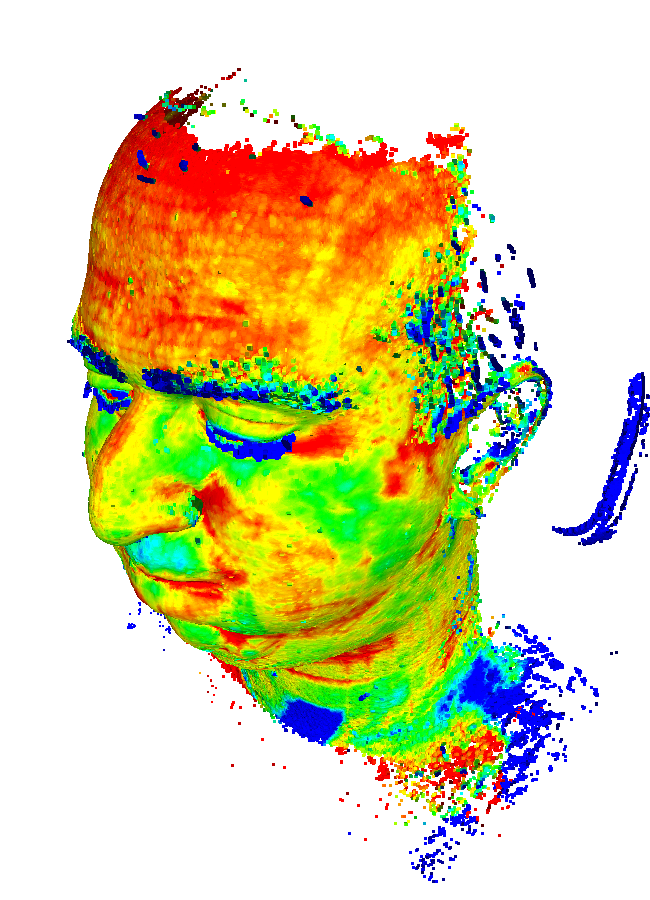}
    \end{subfigure}
    \hfill
    \begin{subfigure}[b]{0.09\textwidth}
        \centering
        \includegraphics[width=\textwidth]{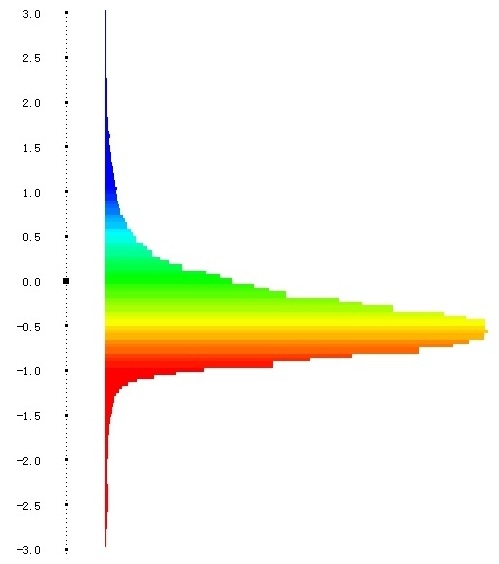}
    \end{subfigure}

    \rotatebox[origin=l]{90}{\makebox[0.09\textwidth][c]{\footnotesize PixelSfM}}
    \hfill
    \begin{subfigure}[b]{0.09\textwidth}
        \centering
        \includegraphics[width=\textwidth]{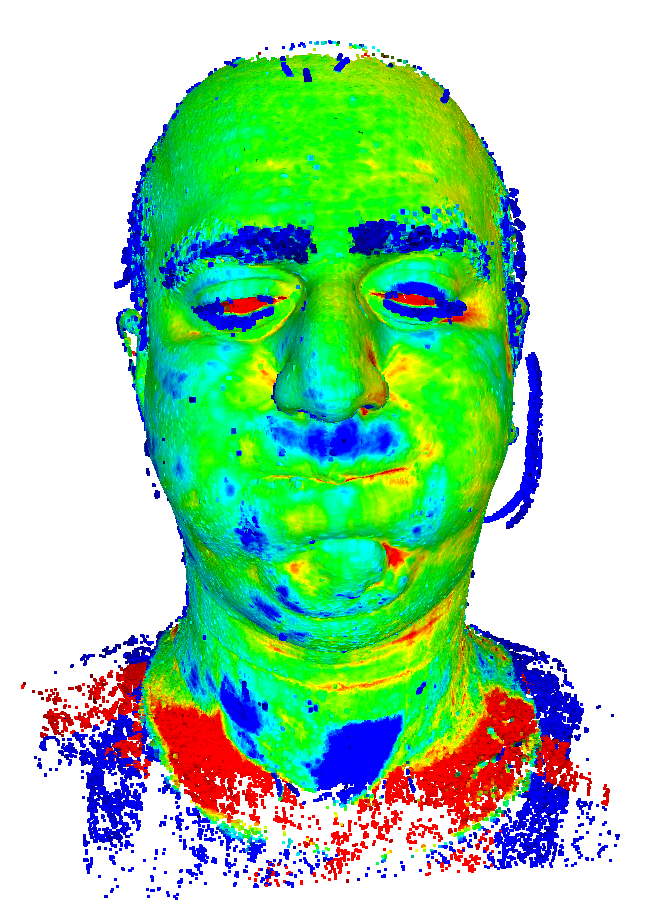}
    \end{subfigure}
    \hfill
    \begin{subfigure}[b]{0.09\textwidth}
        \centering
        \includegraphics[width=\textwidth]{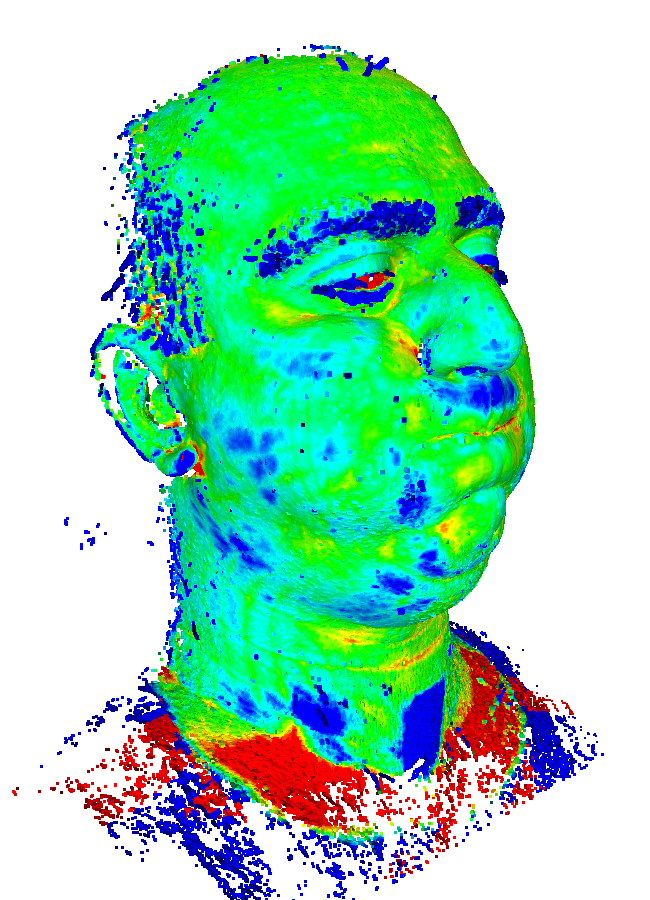}
    \end{subfigure}
    \hfill
    \begin{subfigure}[b]{0.09\textwidth}
        \centering
        \includegraphics[width=\textwidth]{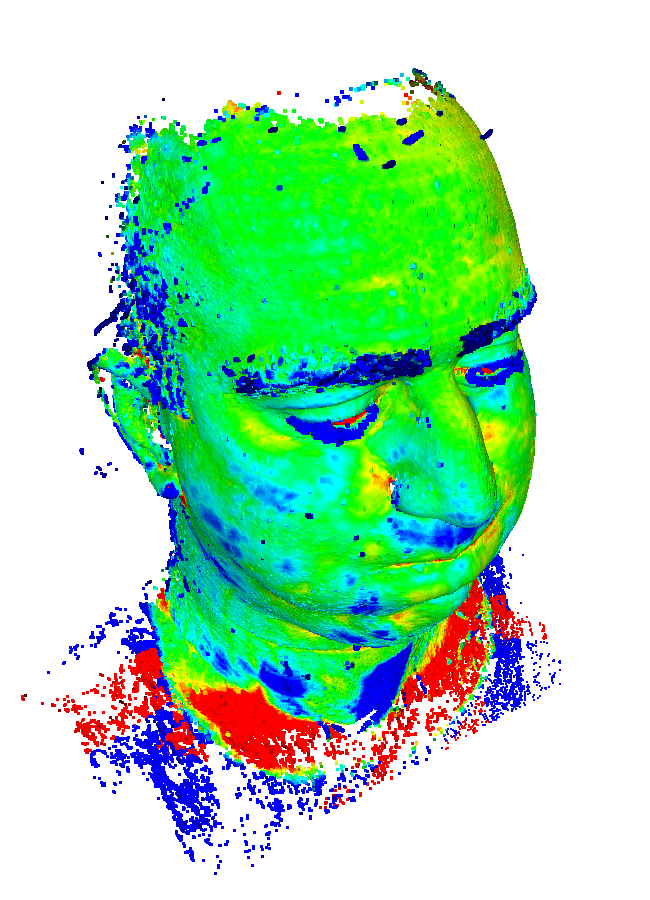}
    \end{subfigure}
    \hfill
    \begin{subfigure}[b]{0.09\textwidth}
        \centering
        \includegraphics[width=\textwidth]{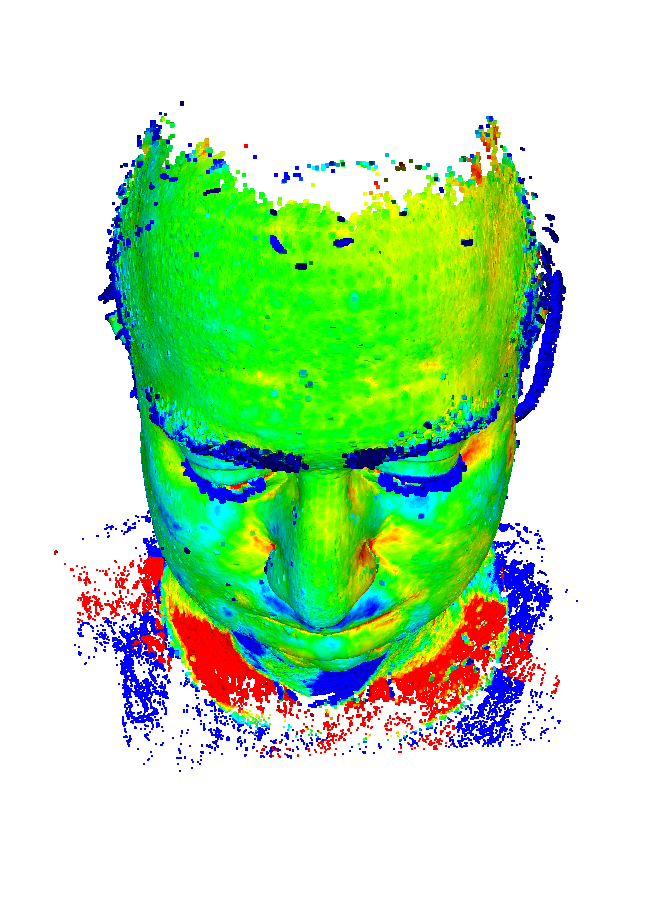}
    \end{subfigure}
    \hfill
    \begin{subfigure}[b]{0.09\textwidth}
        \centering
        \includegraphics[width=\textwidth]{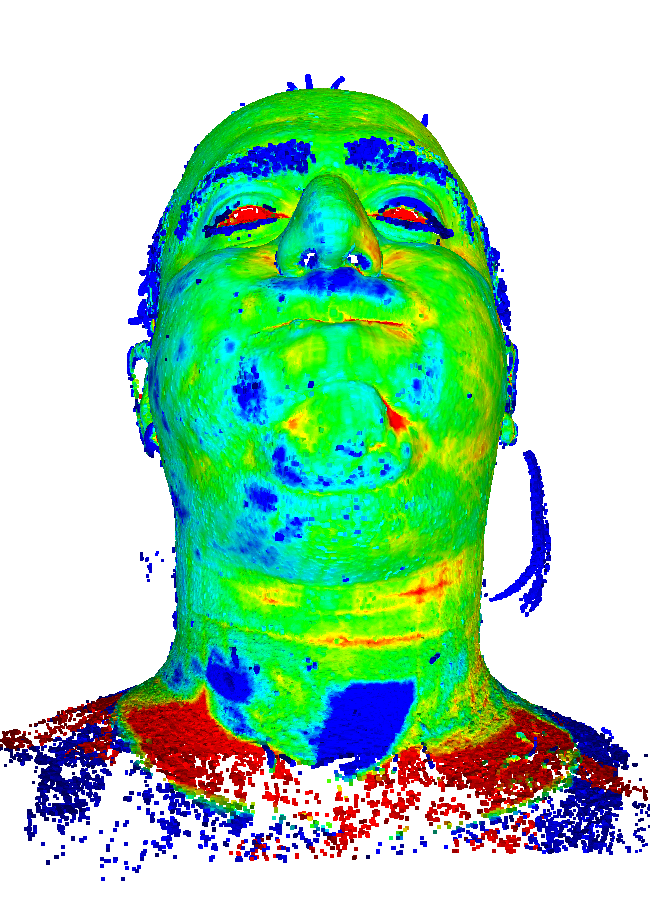}
    \end{subfigure}
    \hfill
    \begin{subfigure}[b]{0.09\textwidth}
        \centering
        \includegraphics[width=\textwidth]{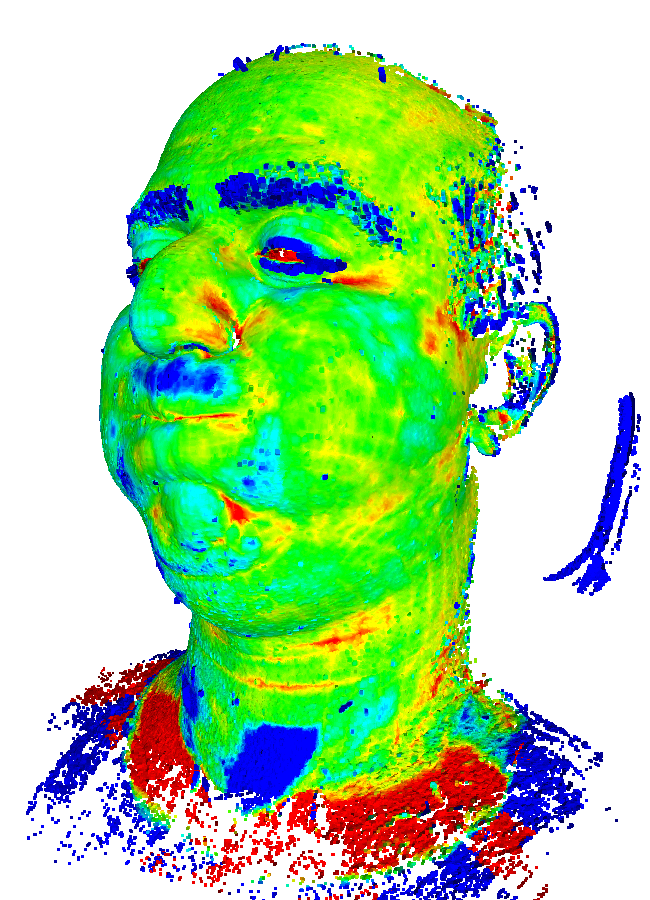}
    \end{subfigure}
    \hfill
    \begin{subfigure}[b]{0.09\textwidth}
        \centering
        \includegraphics[width=\textwidth]{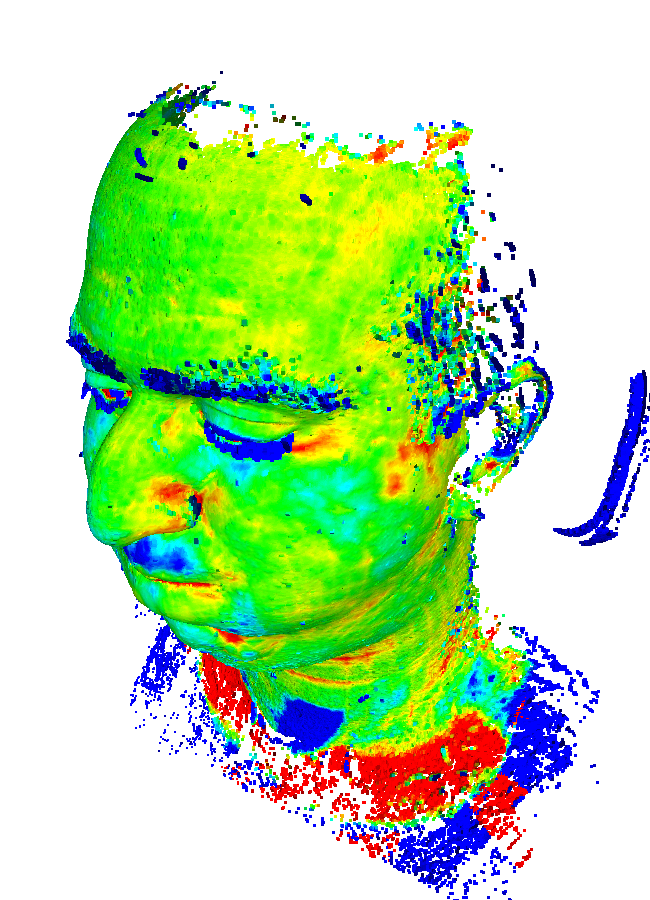}
    \end{subfigure}
    \hfill
    \begin{subfigure}[b]{0.09\textwidth}
        \centering
        \includegraphics[width=\textwidth]{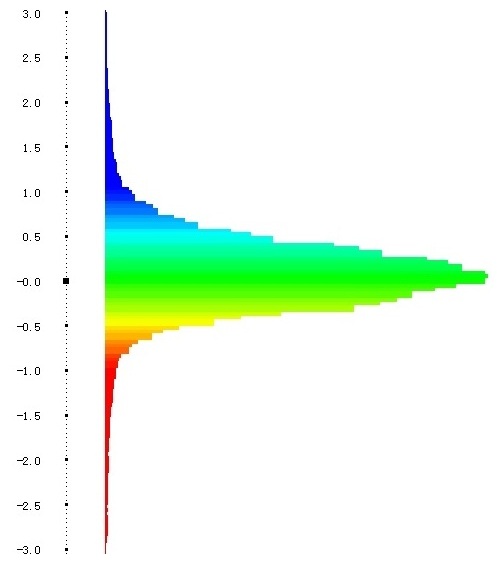}
    \end{subfigure}

    \rotatebox[origin=l]{90}{\makebox[0.09\textwidth][c]{\footnotesize VGGSfM}}
    \hfill
    \begin{subfigure}[b]{0.09\textwidth}
        \centering
        \includegraphics[width=\textwidth]{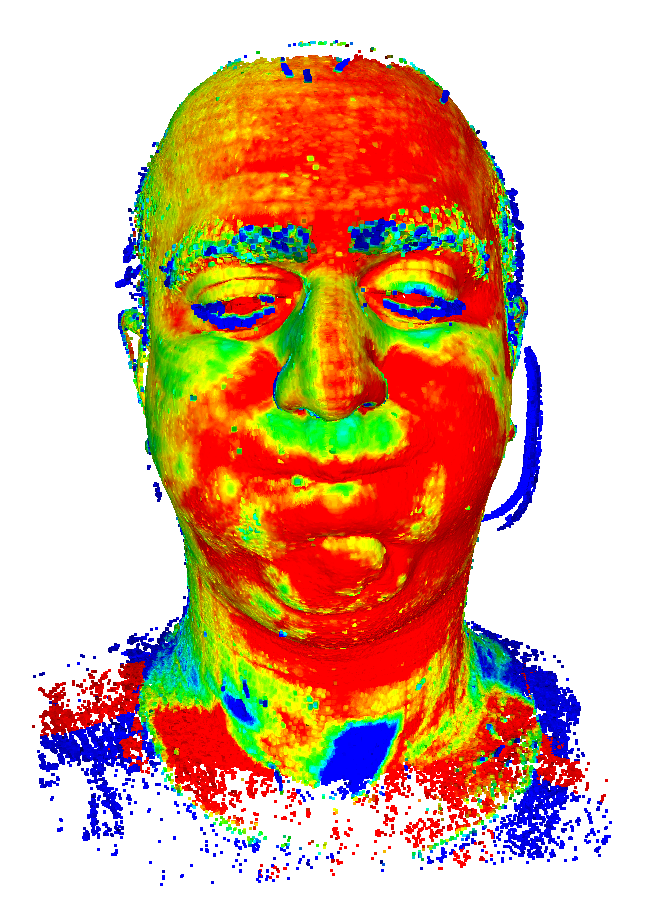}
    \end{subfigure}
    \hfill
    \begin{subfigure}[b]{0.09\textwidth}
        \centering
        \includegraphics[width=\textwidth]{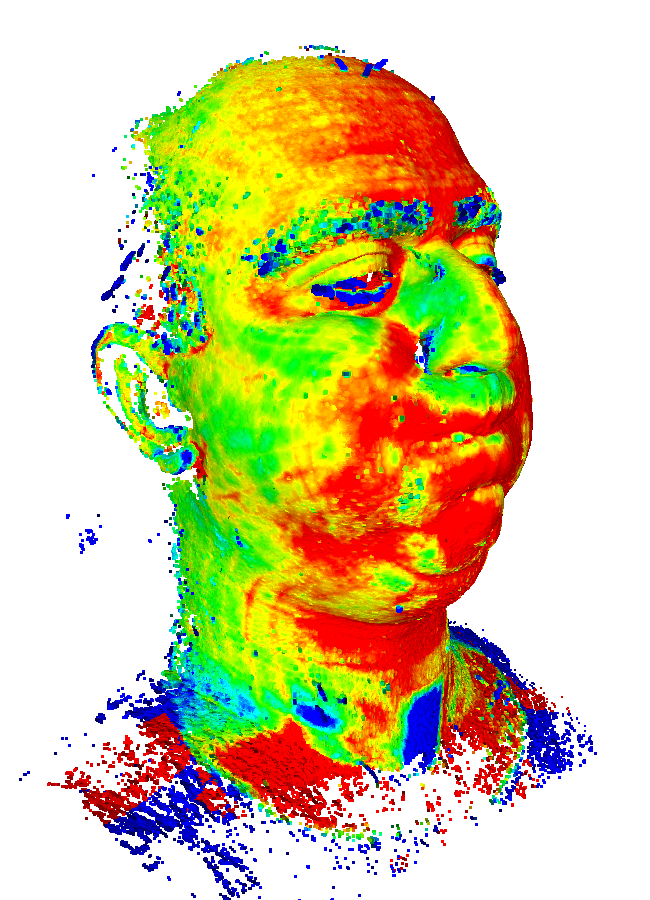}
    \end{subfigure}
    \hfill
    \begin{subfigure}[b]{0.09\textwidth}
        \centering
        \includegraphics[width=\textwidth]{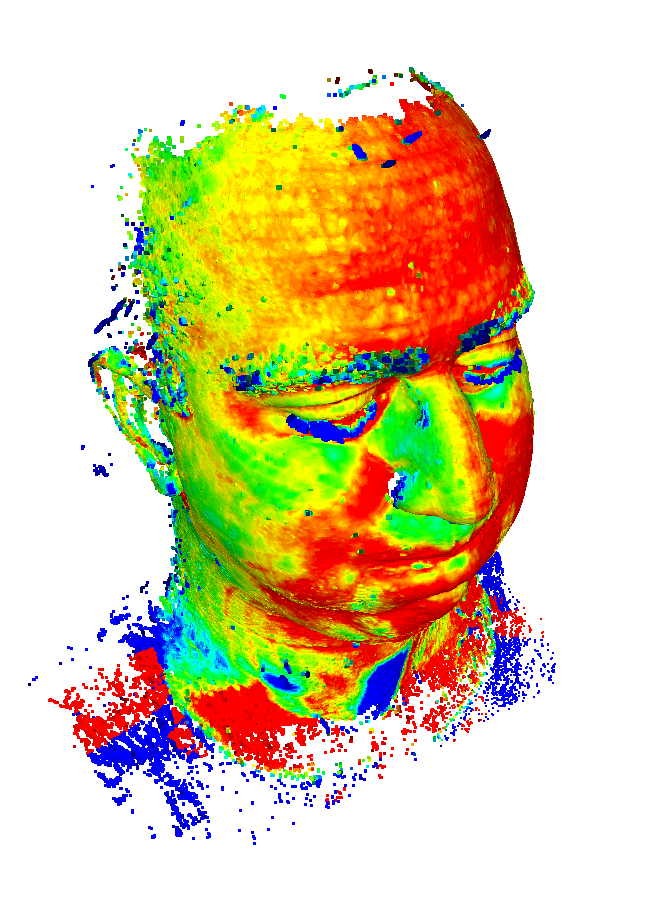}
    \end{subfigure}
    \hfill
    \begin{subfigure}[b]{0.09\textwidth}
        \centering
        \includegraphics[width=\textwidth]{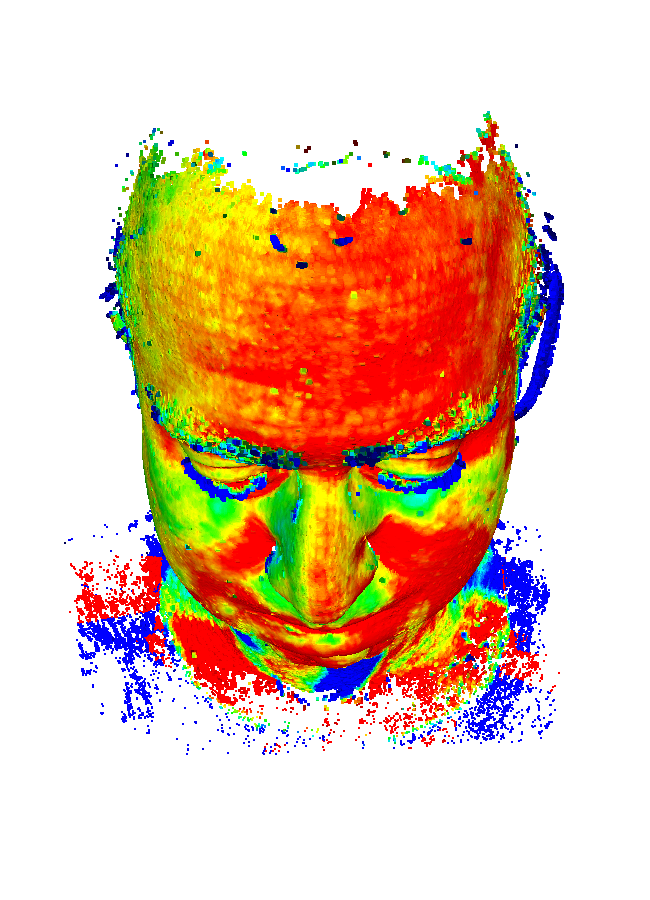}
    \end{subfigure}
    \hfill
    \begin{subfigure}[b]{0.09\textwidth}
        \centering
        \includegraphics[width=\textwidth]{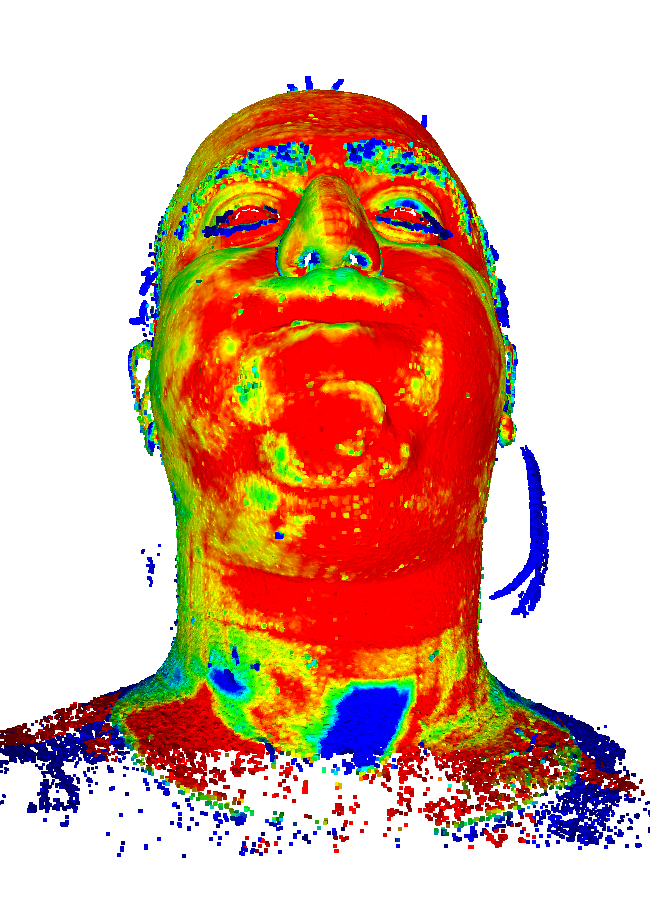}
    \end{subfigure}
    \hfill
    \begin{subfigure}[b]{0.09\textwidth}
        \centering
        \includegraphics[width=\textwidth]{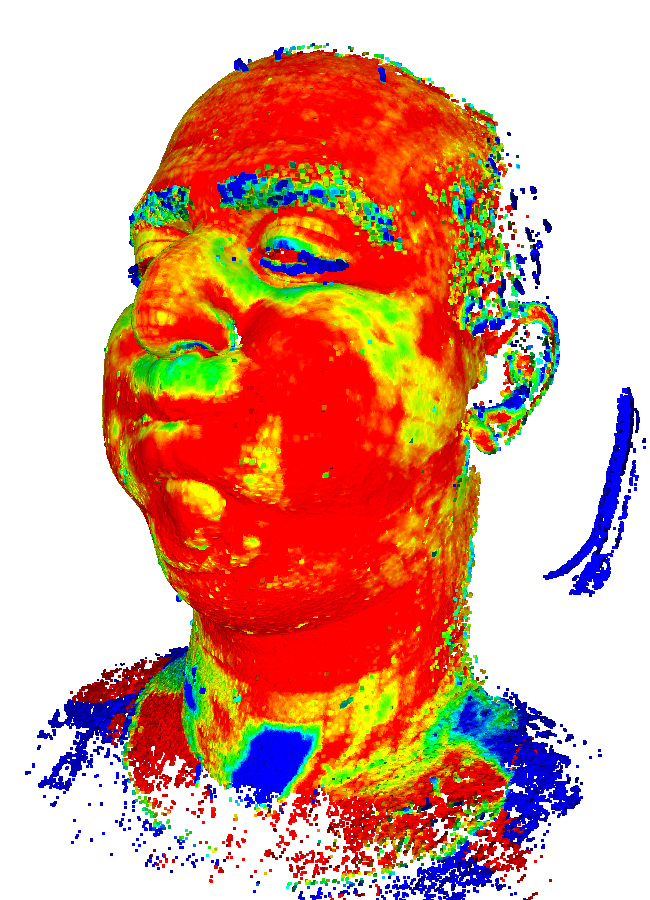}
    \end{subfigure}
    \hfill
    \begin{subfigure}[b]{0.09\textwidth}
        \centering
        \includegraphics[width=\textwidth]{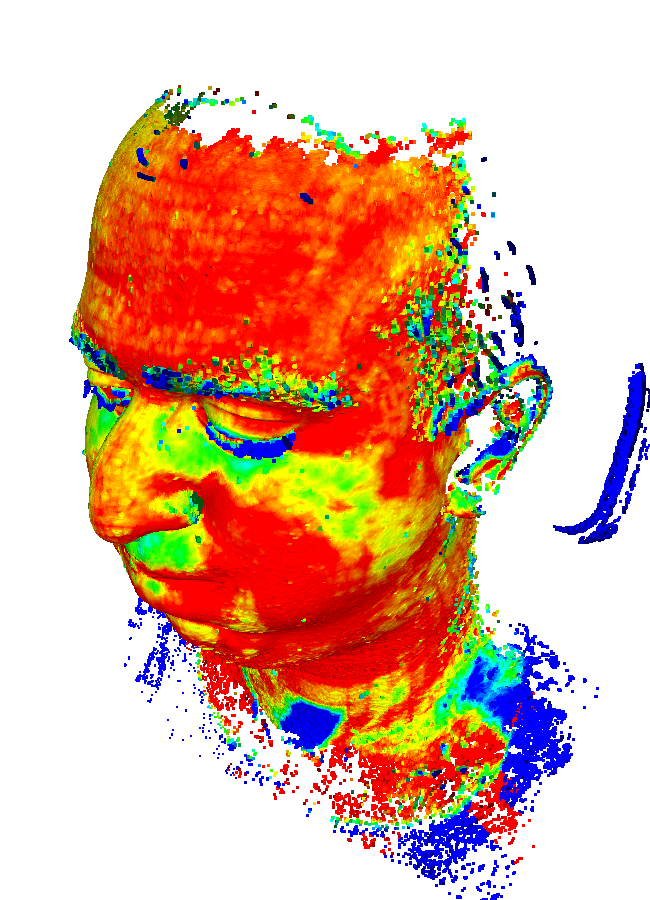}
    \end{subfigure}
    \hfill
    \begin{subfigure}[b]{0.09\textwidth}
        \centering
        \includegraphics[width=\textwidth]{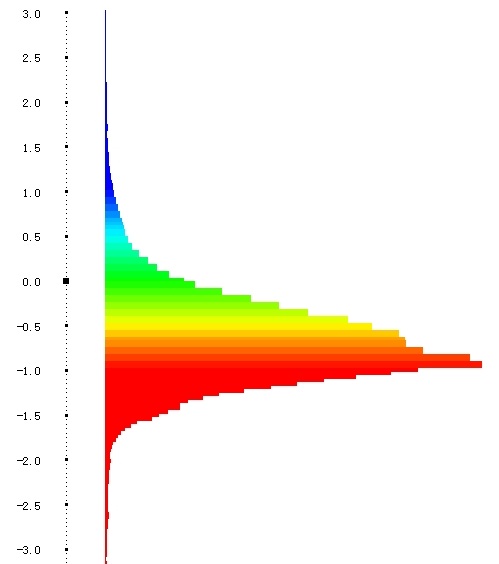}
    \end{subfigure}

    \rotatebox[origin=l]{90}{\makebox[0.09\textwidth][c]{\footnotesize Ours}}
    \hfill
    \begin{subfigure}[b]{0.09\textwidth}
        \centering
        \includegraphics[width=\textwidth]{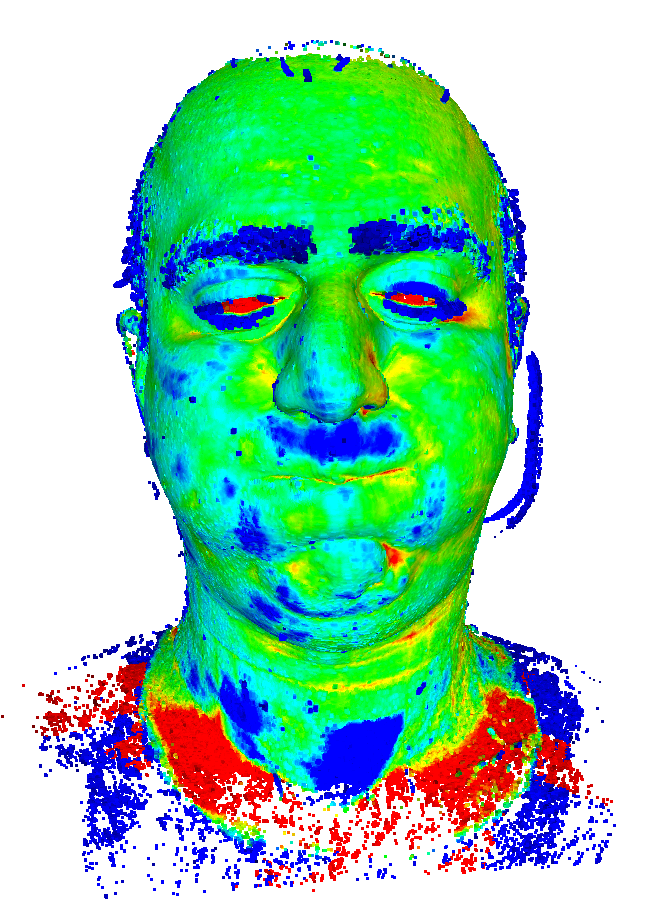}
    \end{subfigure}
    \hfill
    \begin{subfigure}[b]{0.09\textwidth}
        \centering
        \includegraphics[width=\textwidth]{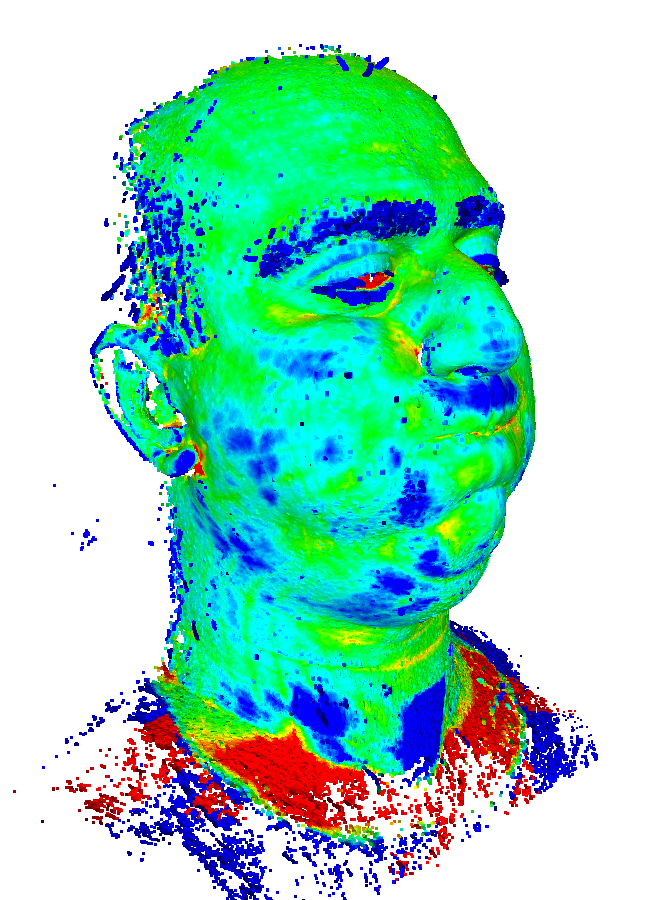}
    \end{subfigure}
    \hfill
    \begin{subfigure}[b]{0.09\textwidth}
        \centering
        \includegraphics[width=\textwidth]{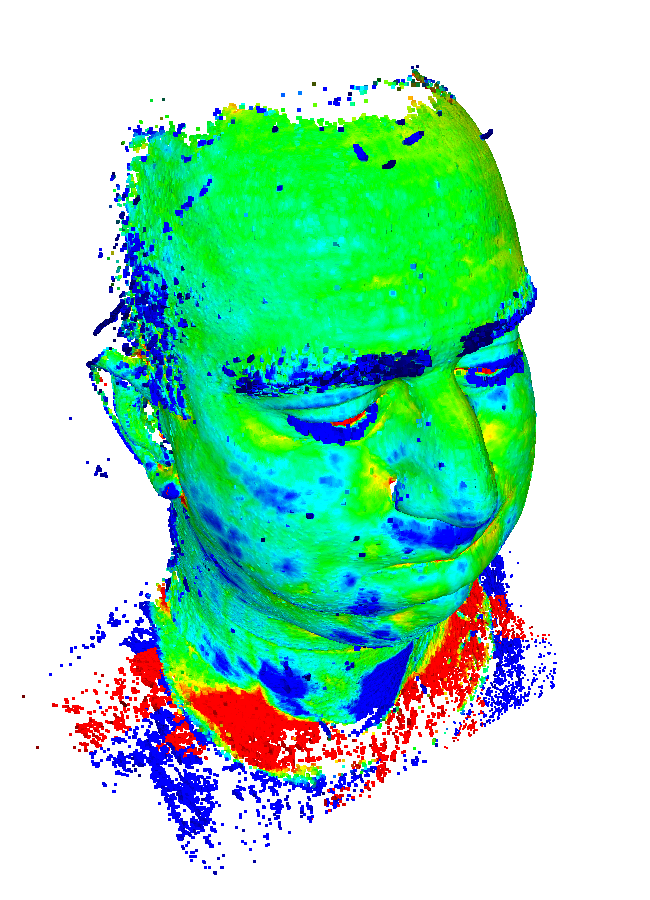}
    \end{subfigure}
    \hfill
    \begin{subfigure}[b]{0.09\textwidth}
        \centering
        \includegraphics[width=\textwidth]{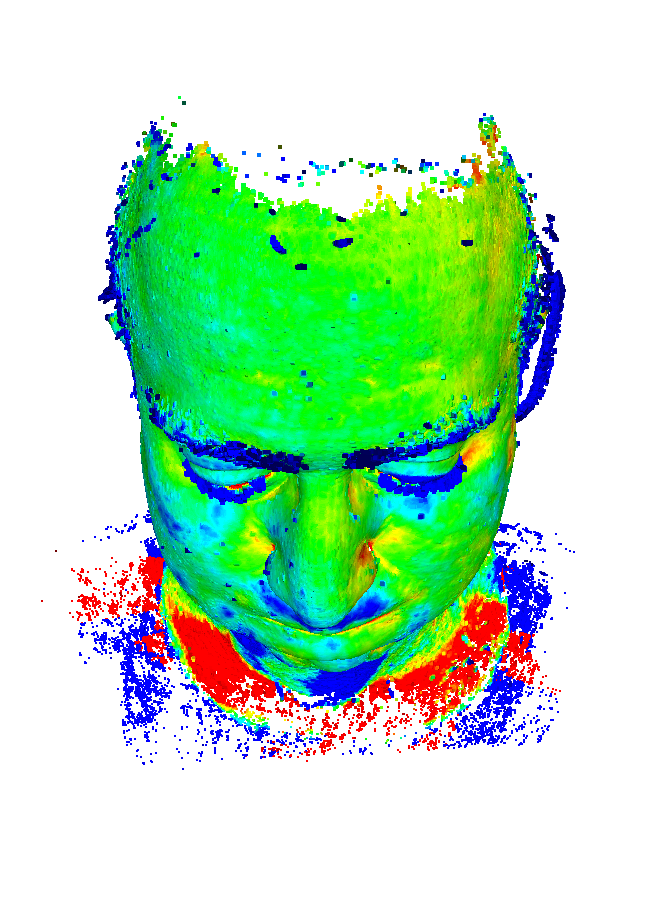}
    \end{subfigure}
    \hfill
    \begin{subfigure}[b]{0.09\textwidth}
        \centering
        \includegraphics[width=\textwidth]{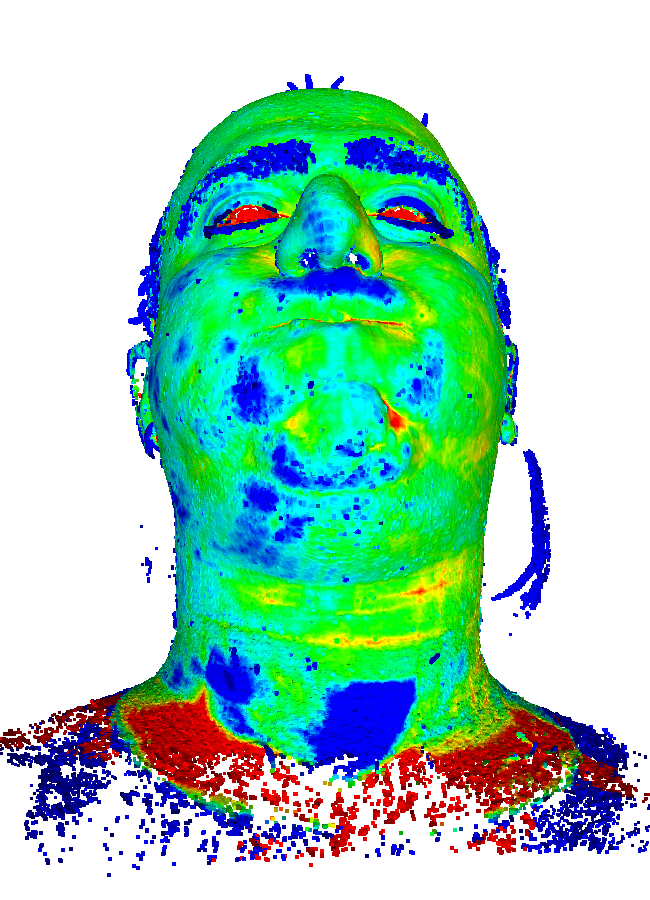}
    \end{subfigure}
    \hfill
    \begin{subfigure}[b]{0.09\textwidth}
        \centering
        \includegraphics[width=\textwidth]{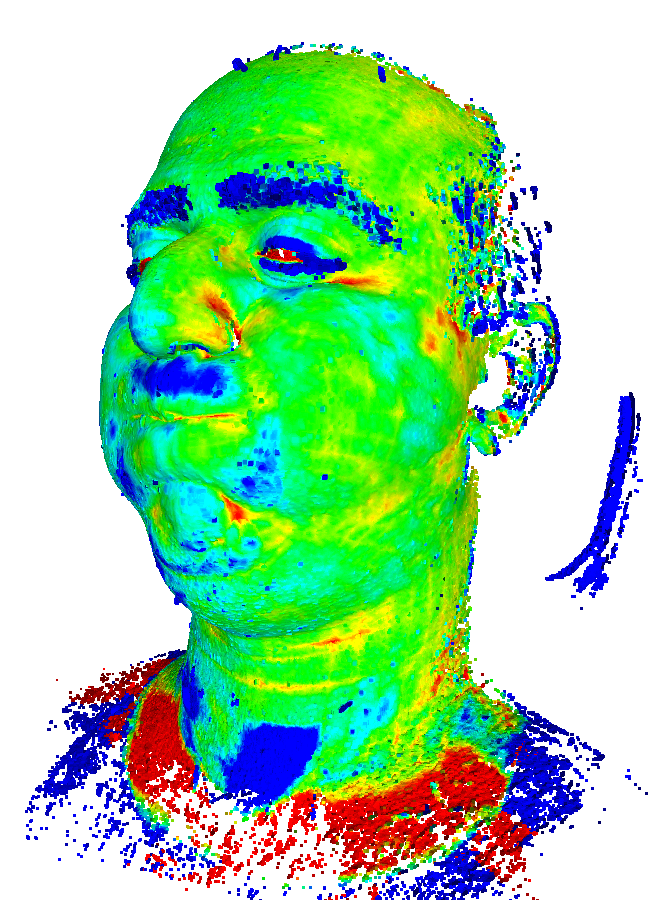}
    \end{subfigure}
    \hfill
    \begin{subfigure}[b]{0.09\textwidth}
        \centering
        \includegraphics[width=\textwidth]{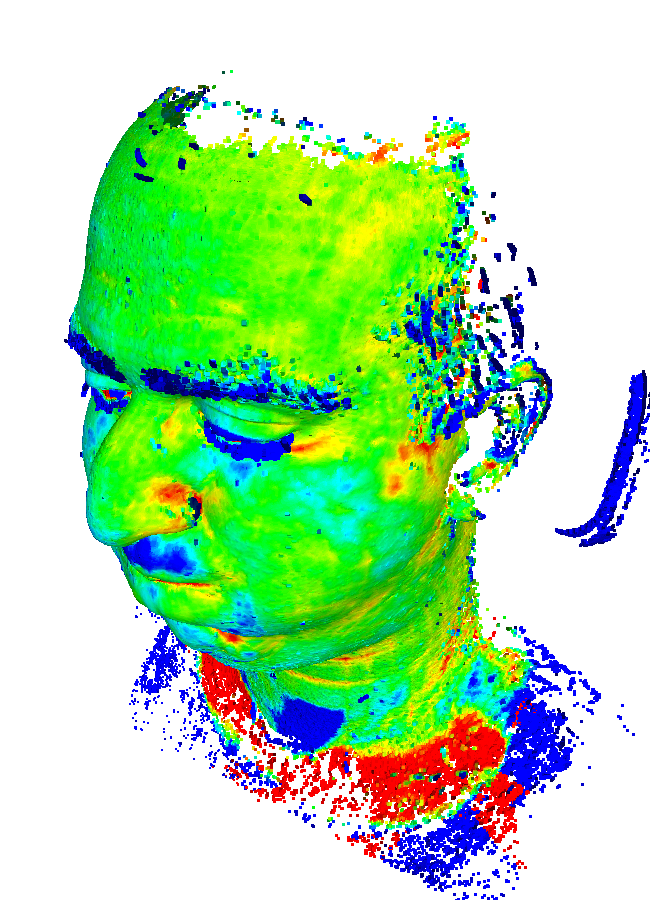}
    \end{subfigure}
    \hfill
    \begin{subfigure}[b]{0.09\textwidth}
        \centering
        \includegraphics[width=\textwidth]{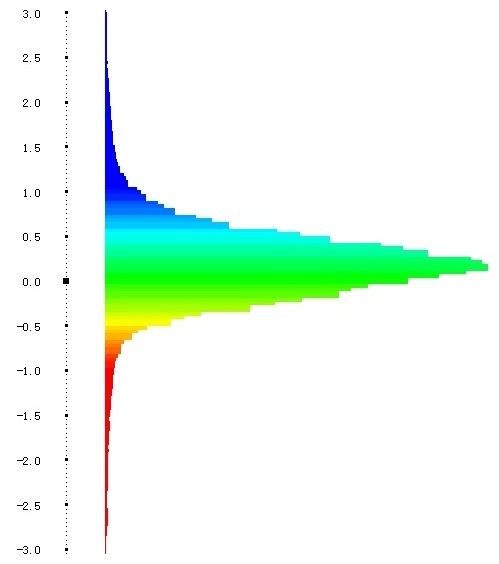}
    \end{subfigure}
   \caption{Comparative visualization of MVS reconstruction: COLMAP, Pixel-Perfect SfM, VGGSfM, and our method. All models are reconstructed using the COLMAP MVS pipeline with varying intrinsic. We compute point-to-ground-truth mesh distances, color-encoding them (red for negative, blue for positive distances), and visualize RGB models alongside histograms. Our method with refined intrinsics, exhibits more green points, indicating closer alignment with the ground-truth model.}
    \label{fig:MVS-comparison}
\end{figure*}

\begin{figure*}[ht]
    \centering

    \rotatebox[origin=l]{90}{\makebox[0.09\textwidth][c]{\footnotesize Linear}}
    \hfill
    \begin{subfigure}[b]{0.09\textwidth}
        \centering
        \includegraphics[width=\textwidth]{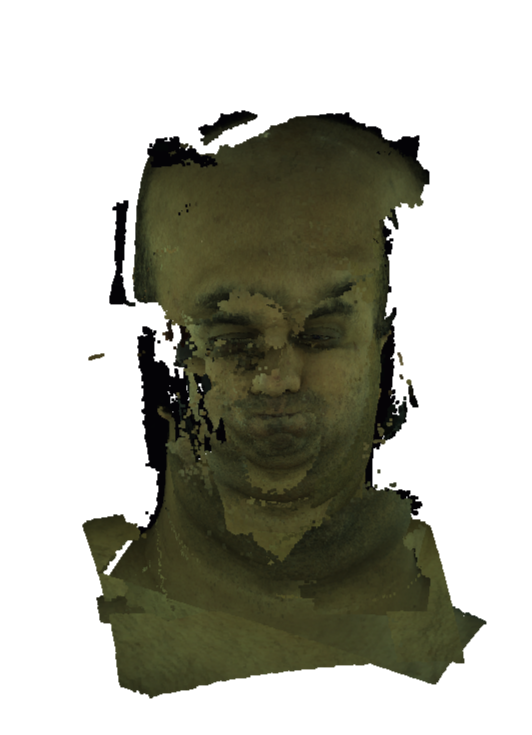}
    \end{subfigure}
    \hfill
    \begin{subfigure}[b]{0.09\textwidth}
        \centering
        \includegraphics[width=\textwidth]{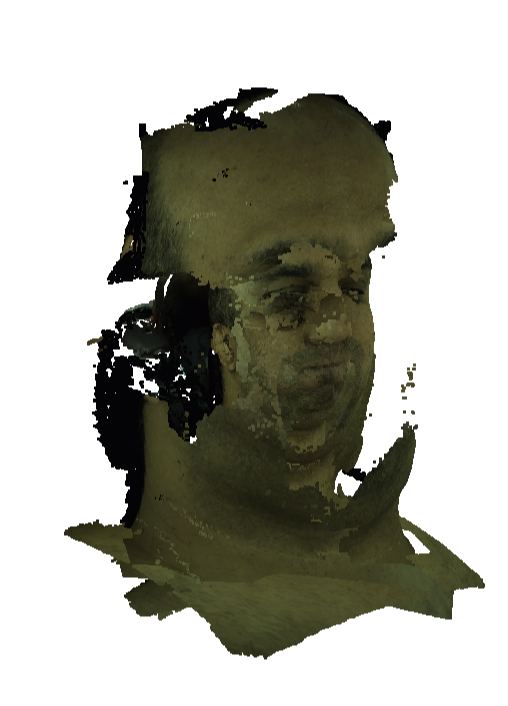}
    \end{subfigure}
    \hfill
    \begin{subfigure}[b]{0.09\textwidth}
        \centering
        \includegraphics[width=\textwidth]{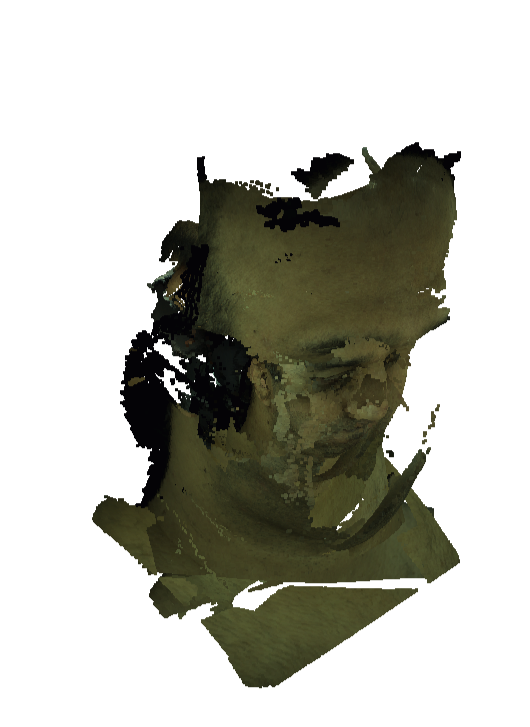}
    \end{subfigure}
    \hfill
    \begin{subfigure}[b]{0.09\textwidth}
        \centering
        \includegraphics[width=\textwidth]{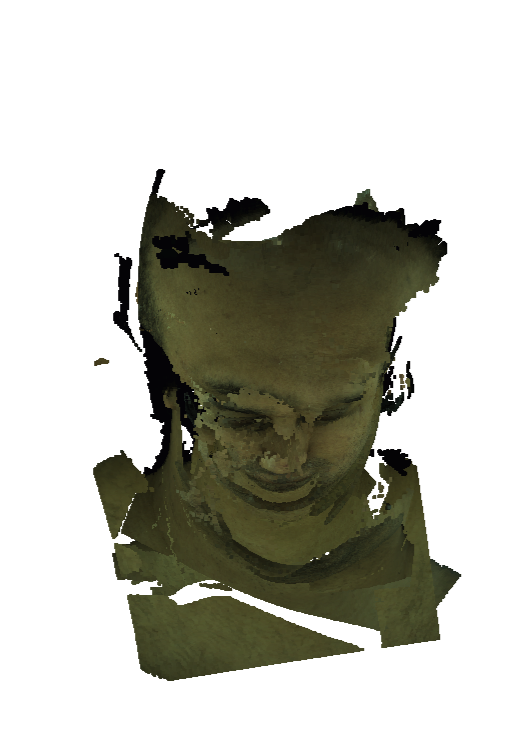}
    \end{subfigure}
    \hfill
    \begin{subfigure}[b]{0.09\textwidth}
        \centering
        \includegraphics[width=\textwidth]{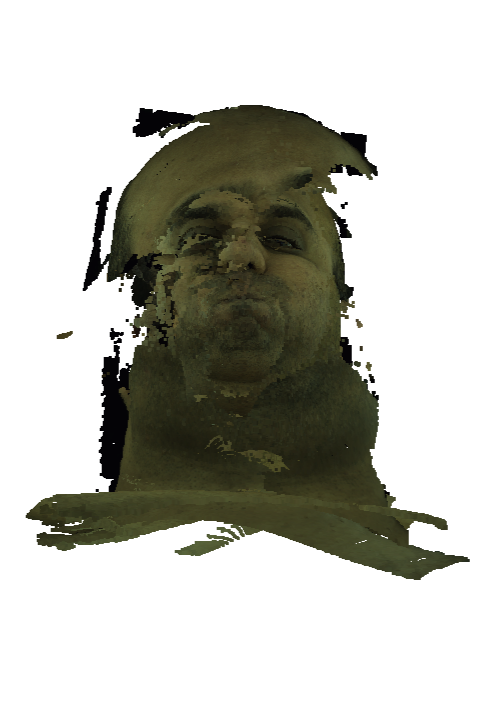}
    \end{subfigure}
    \hfill
    \begin{subfigure}[b]{0.09\textwidth}
        \centering
        \includegraphics[width=\textwidth]{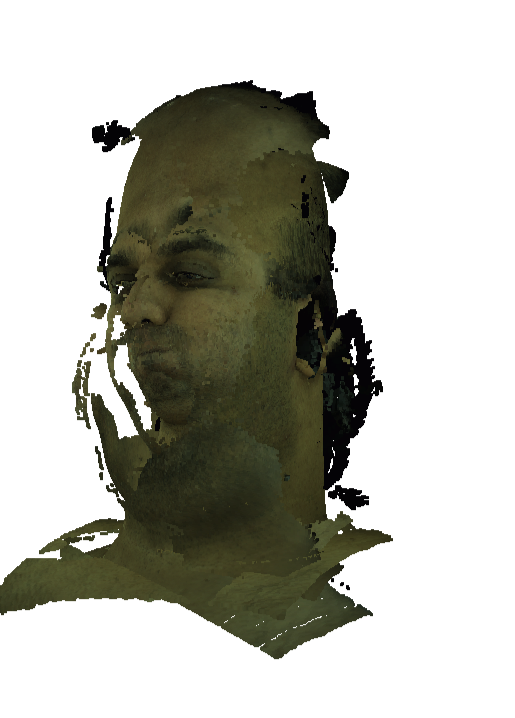}
    \end{subfigure}
    \hfill
    \begin{subfigure}[b]{0.09\textwidth}
        \centering
        \includegraphics[width=\textwidth]{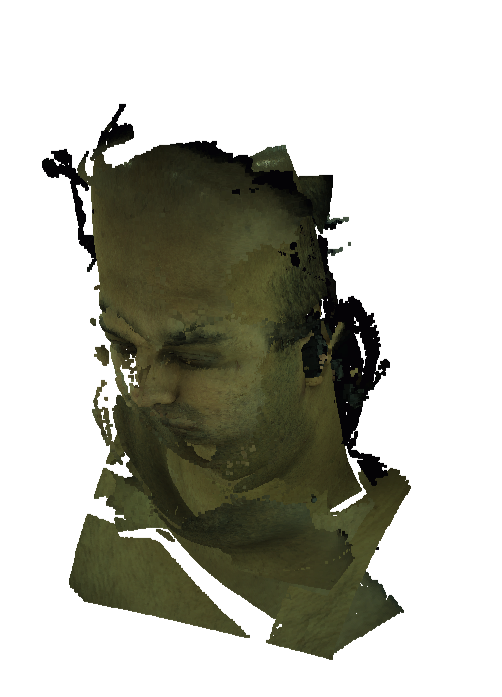}
    \end{subfigure}

    \rotatebox[origin=l]{90}{\makebox[0.09\textwidth][c]{\footnotesize Linear + Ours}}
    \hfill
    \begin{subfigure}[b]{0.09\textwidth}
        \centering
        \includegraphics[width=\textwidth]{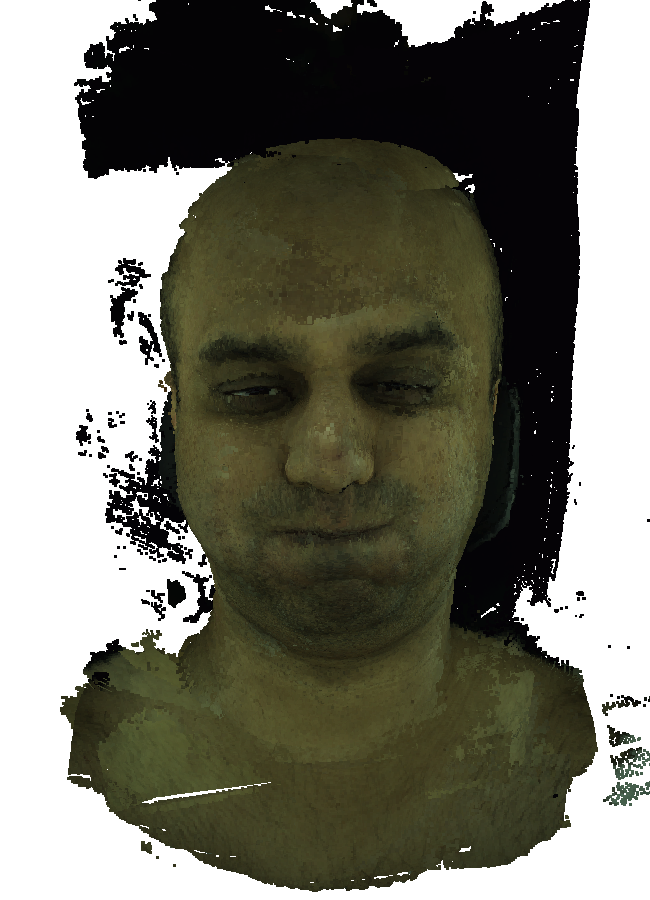}
    \end{subfigure}
    \hfill
    \begin{subfigure}[b]{0.09\textwidth}
        \centering
        \includegraphics[width=\textwidth]{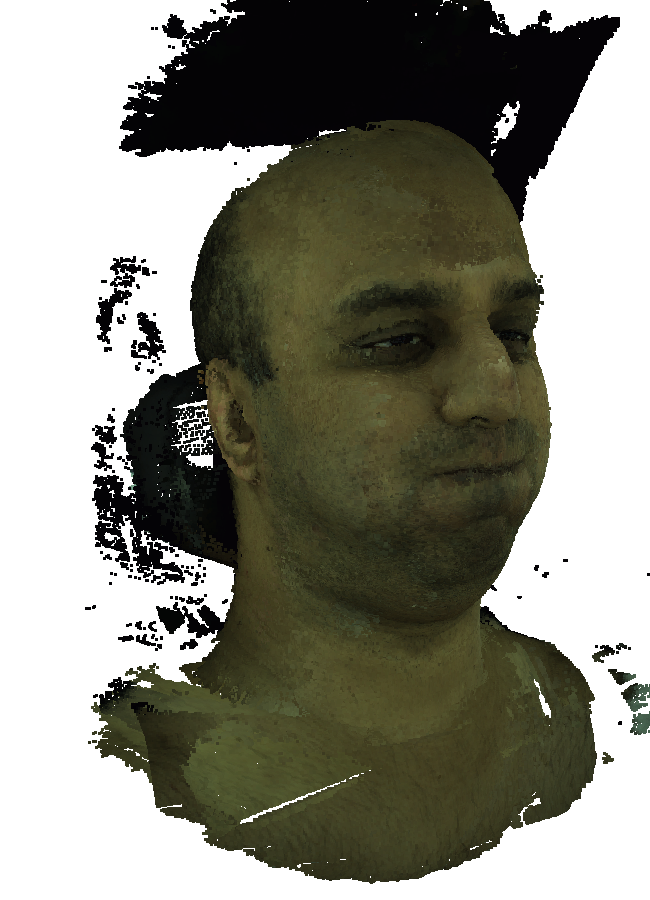}
    \end{subfigure}
    \hfill
    \begin{subfigure}[b]{0.09\textwidth}
        \centering
        \includegraphics[width=\textwidth]{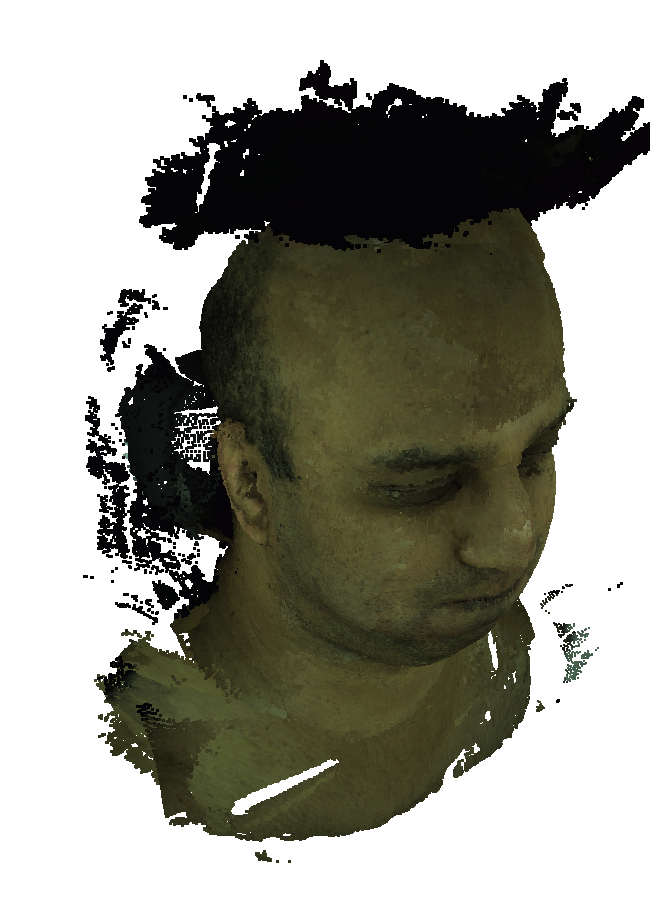}
    \end{subfigure}
    \hfill
    \begin{subfigure}[b]{0.09\textwidth}
        \centering
        \includegraphics[width=\textwidth]{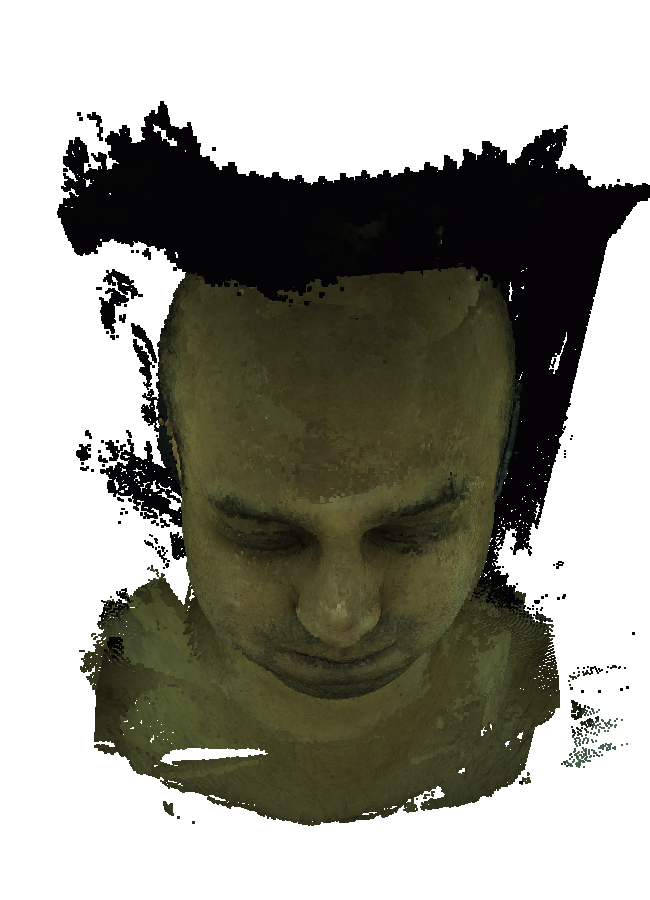}
    \end{subfigure}
    \hfill
    \begin{subfigure}[b]{0.09\textwidth}
        \centering
        \includegraphics[width=\textwidth]{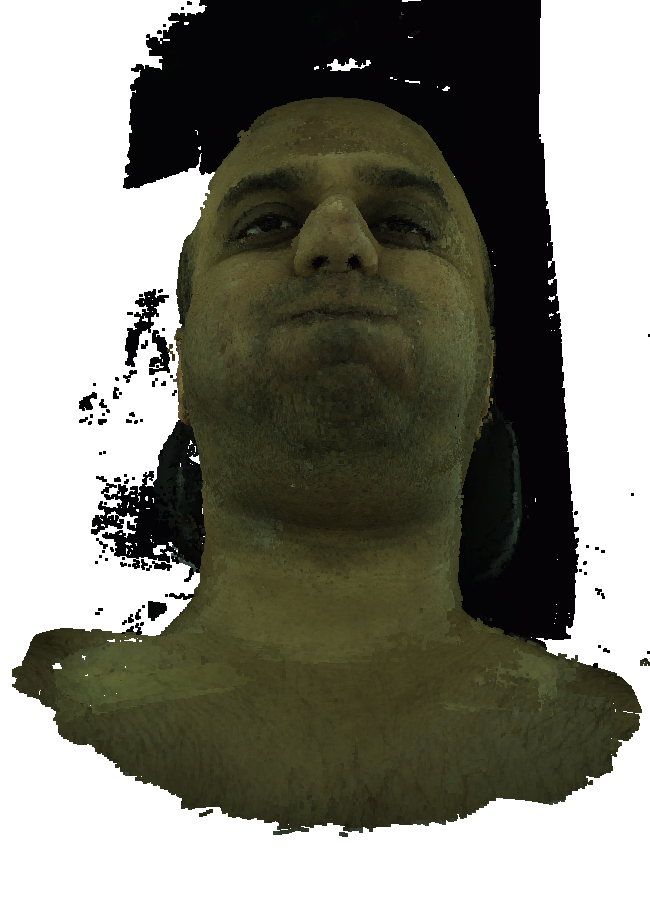}
    \end{subfigure}
    \hfill
    \begin{subfigure}[b]{0.09\textwidth}
        \centering
        \includegraphics[width=\textwidth]{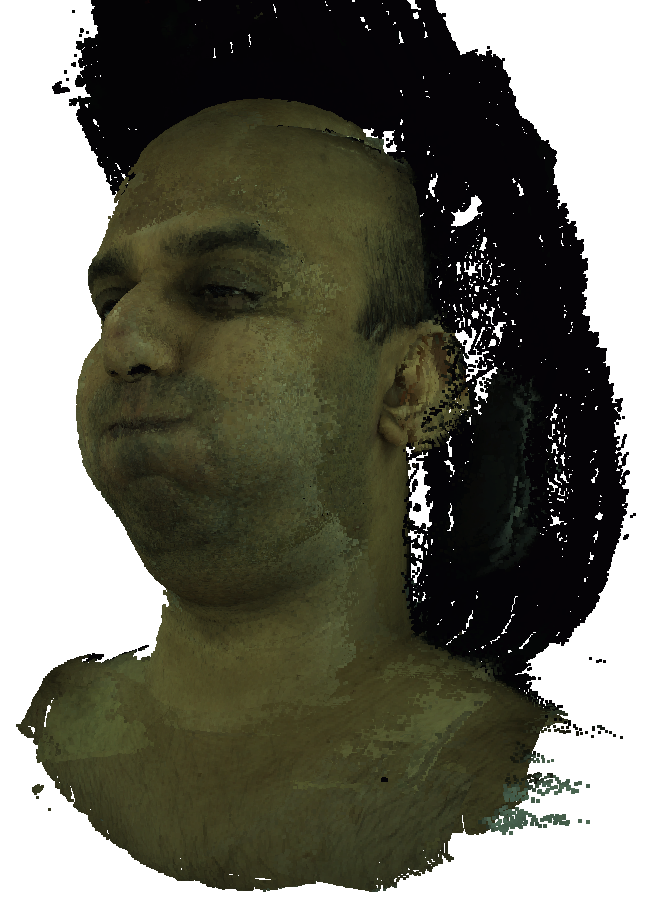}
    \end{subfigure}
    \hfill
    \begin{subfigure}[b]{0.09\textwidth}
        \centering
        \includegraphics[width=\textwidth]{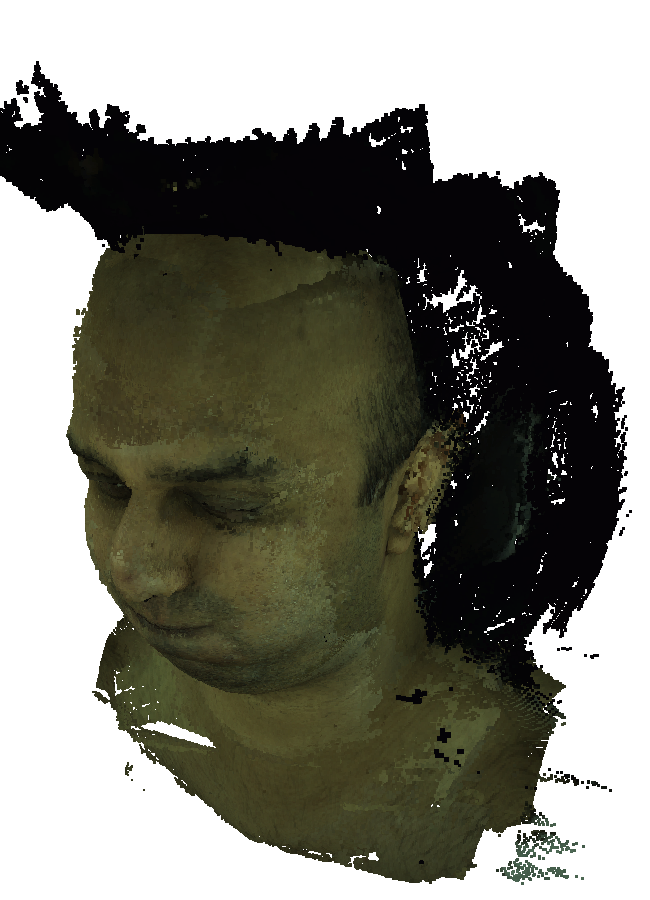}
    \end{subfigure}

    \rotatebox[origin=l]{90}{\makebox[0.09\textwidth][c]{\footnotesize DPT}}
    \hfill
    \begin{subfigure}[b]{0.09\textwidth}
        \centering
        \includegraphics[width=\textwidth]{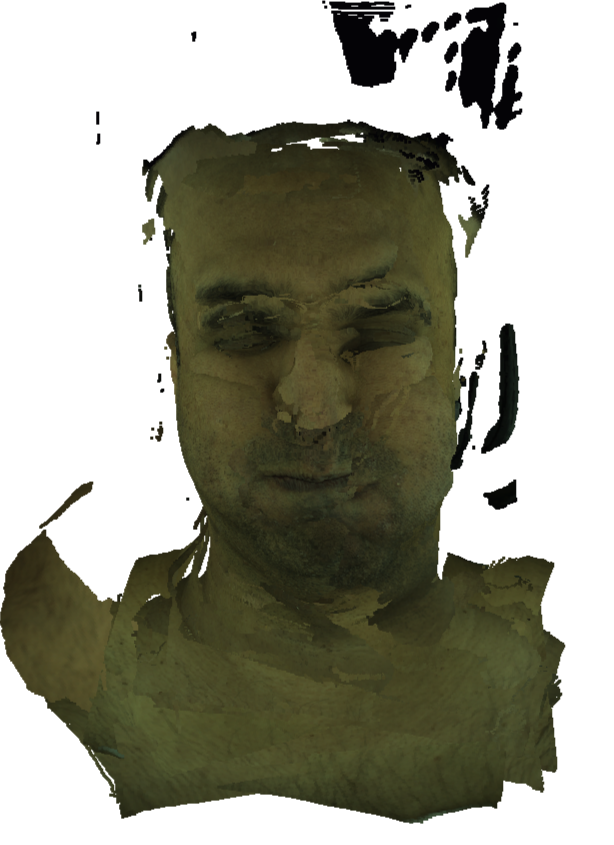}
    \end{subfigure}
    \hfill
    \begin{subfigure}[b]{0.09\textwidth}
        \centering
        \includegraphics[width=\textwidth]{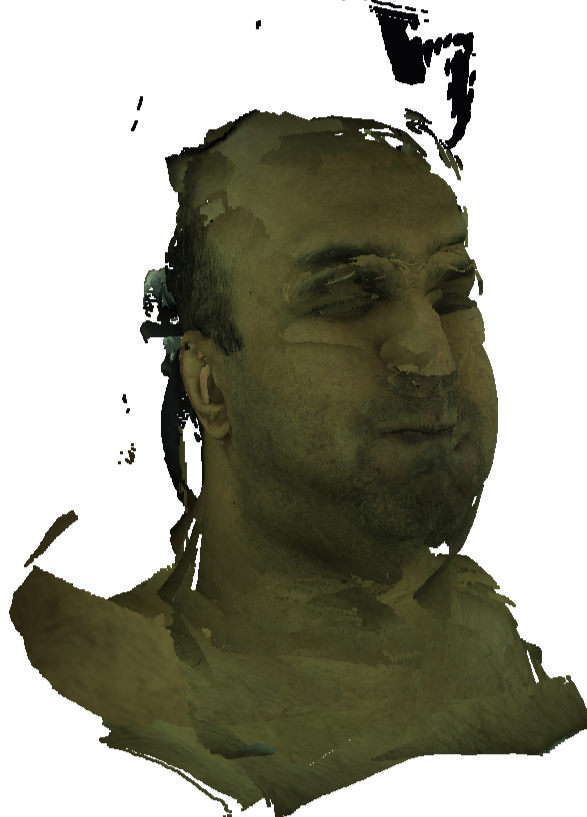}
    \end{subfigure}
    \hfill
    \begin{subfigure}[b]{0.09\textwidth}
        \centering
        \includegraphics[width=\textwidth]{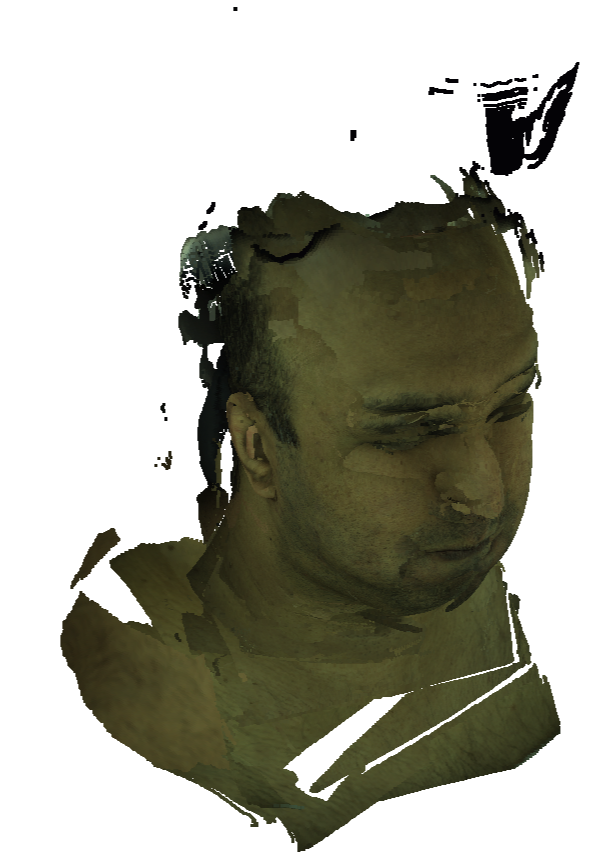}
    \end{subfigure}
    \hfill
    \begin{subfigure}[b]{0.09\textwidth}
        \centering
        \includegraphics[width=\textwidth]{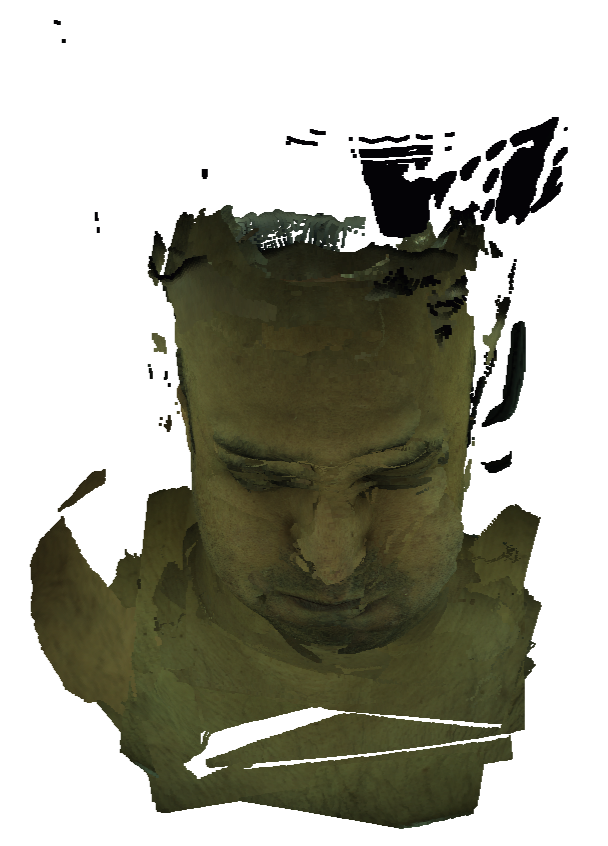}
    \end{subfigure}
    \hfill
    \begin{subfigure}[b]{0.09\textwidth}
        \centering
        \includegraphics[width=\textwidth]{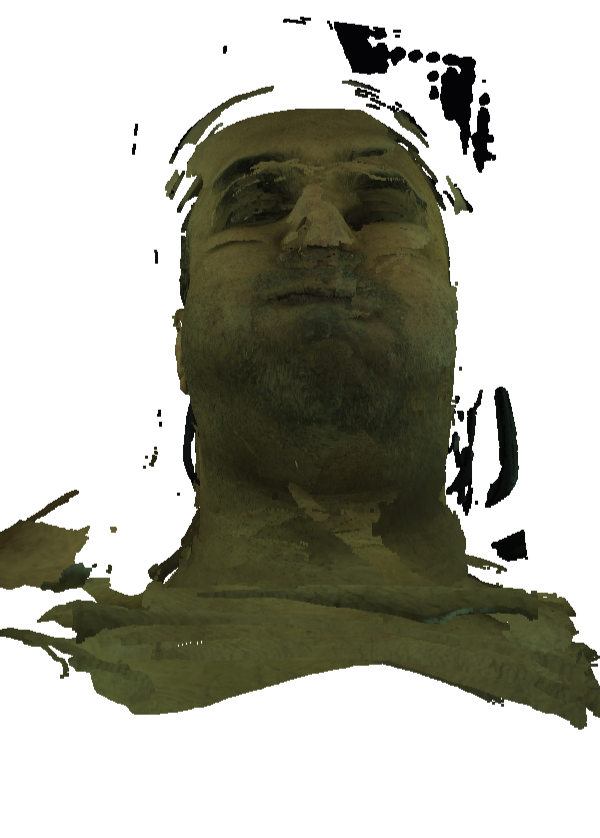}
    \end{subfigure}
    \hfill
    \begin{subfigure}[b]{0.09\textwidth}
        \centering
        \includegraphics[width=\textwidth]{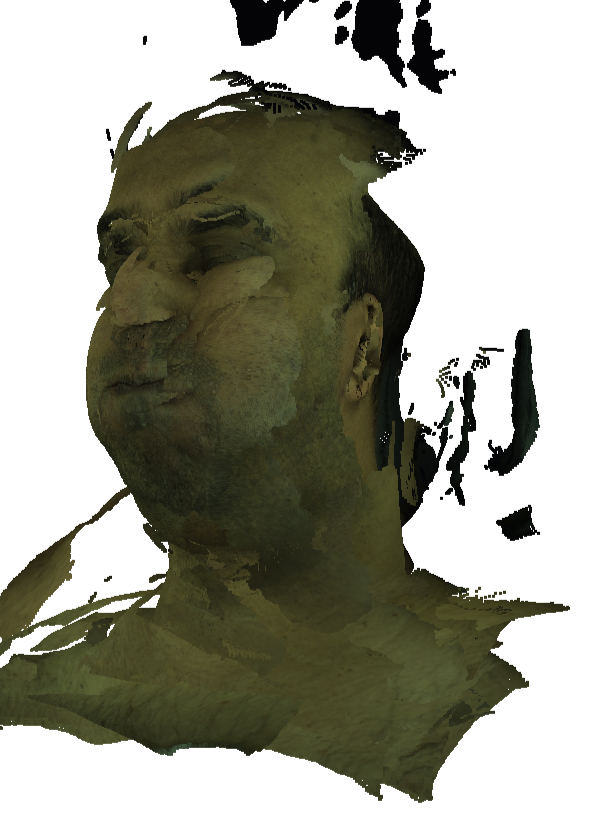}
    \end{subfigure}
    \hfill
    \begin{subfigure}[b]{0.09\textwidth}
        \centering
        \includegraphics[width=\textwidth]{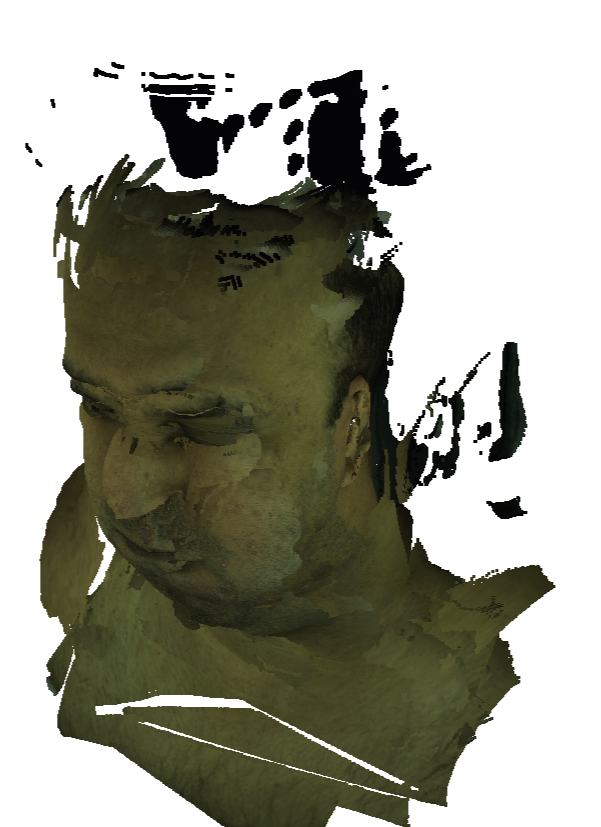}
    \end{subfigure}

    \rotatebox[origin=l]{90}{\makebox[0.09\textwidth][c]{\footnotesize DPT + Ours}}
    \hfill
    \begin{subfigure}[b]{0.09\textwidth}
        \centering
        \includegraphics[width=\textwidth]{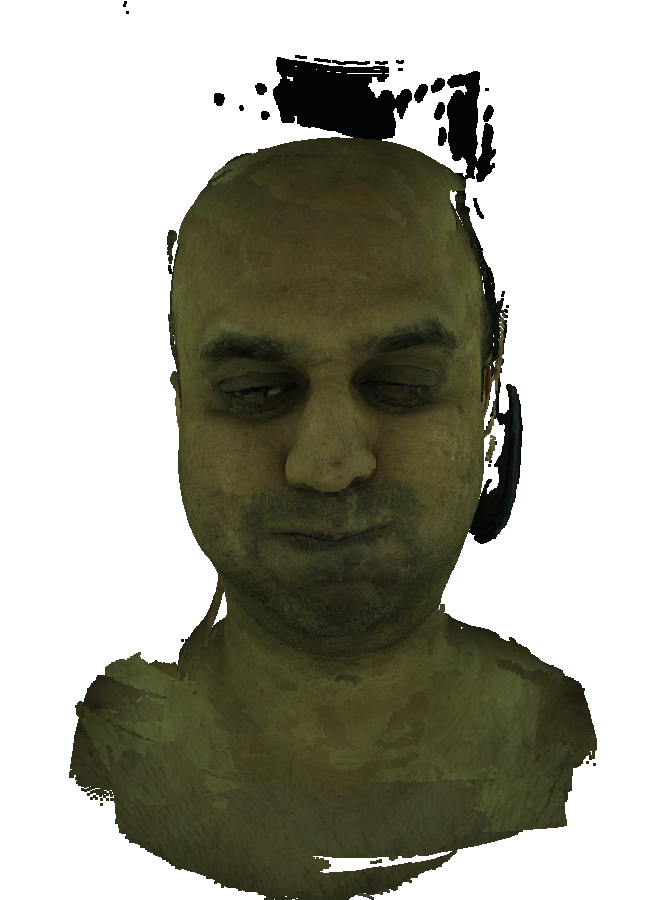}
    \end{subfigure}
    \hfill
    \begin{subfigure}[b]{0.09\textwidth}
        \centering
        \includegraphics[width=\textwidth]{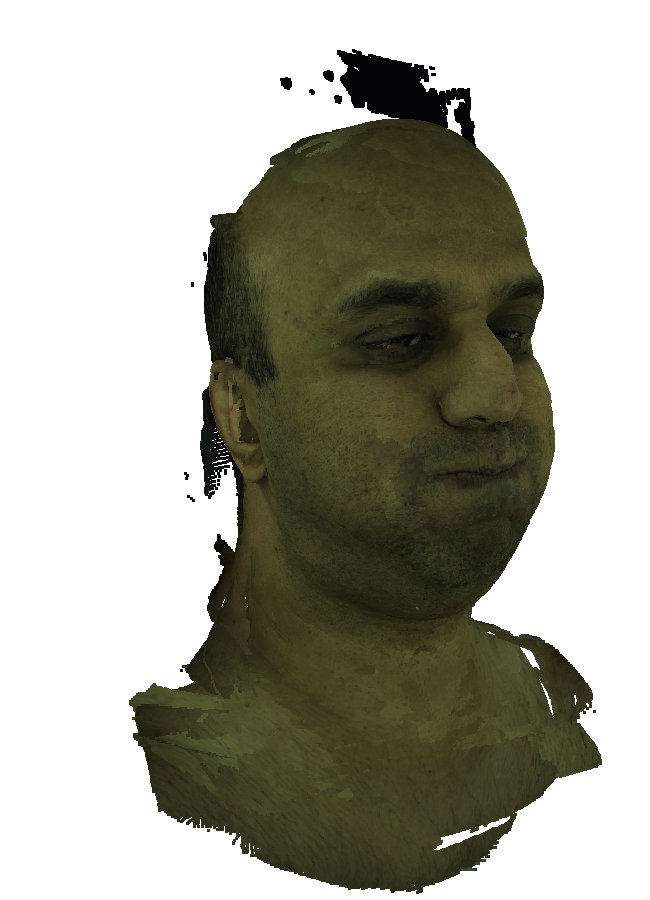}
    \end{subfigure}
    \hfill
    \begin{subfigure}[b]{0.09\textwidth}
        \centering
        \includegraphics[width=\textwidth]{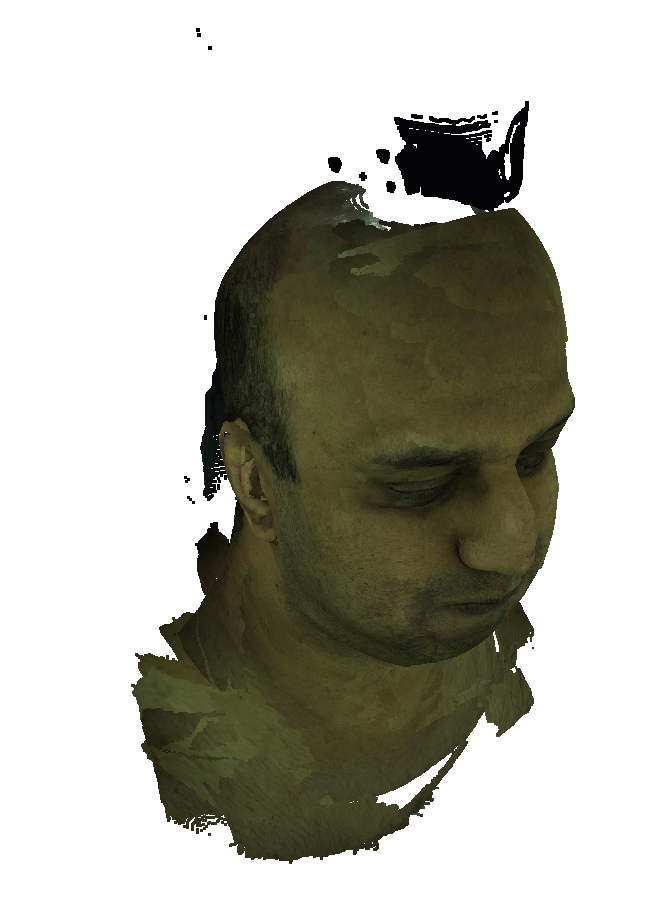}
    \end{subfigure}
    \hfill
    \begin{subfigure}[b]{0.09\textwidth}
        \centering
        \includegraphics[width=\textwidth]{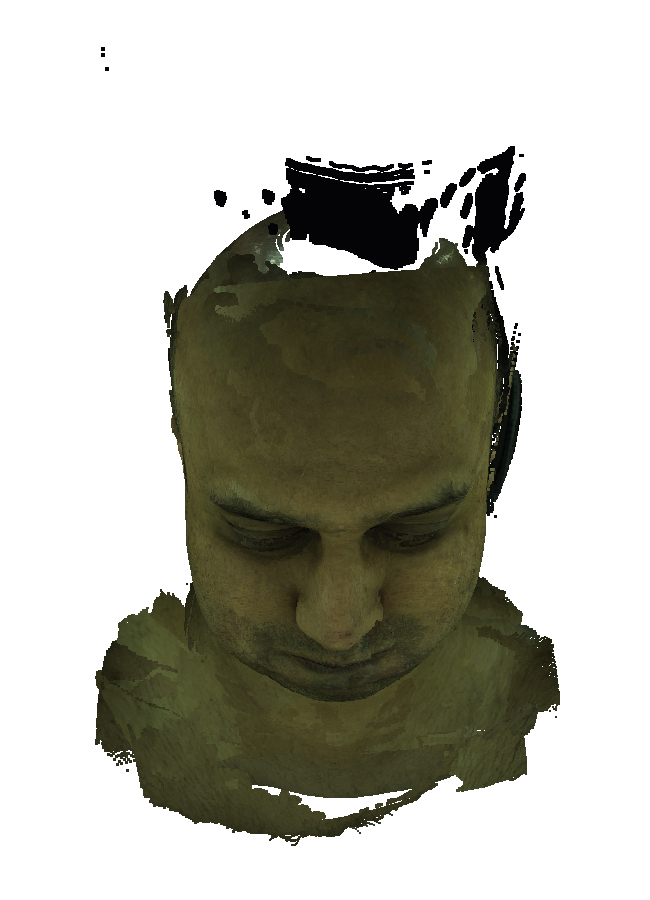}
    \end{subfigure}
    \hfill
    \begin{subfigure}[b]{0.09\textwidth}
        \centering
        \includegraphics[width=\textwidth]{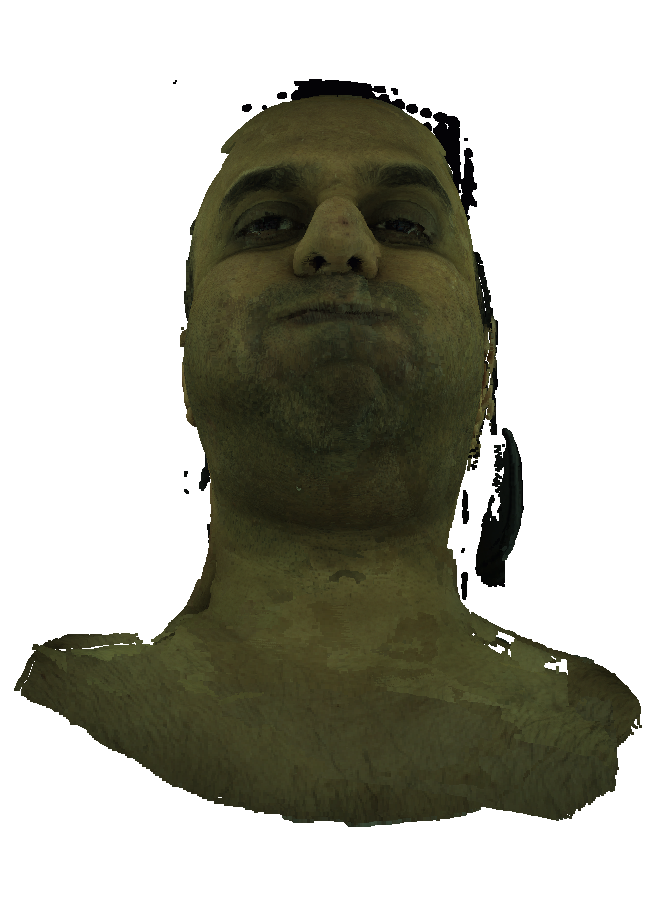}
    \end{subfigure}
    \hfill
    \begin{subfigure}[b]{0.09\textwidth}
        \centering
        \includegraphics[width=\textwidth]{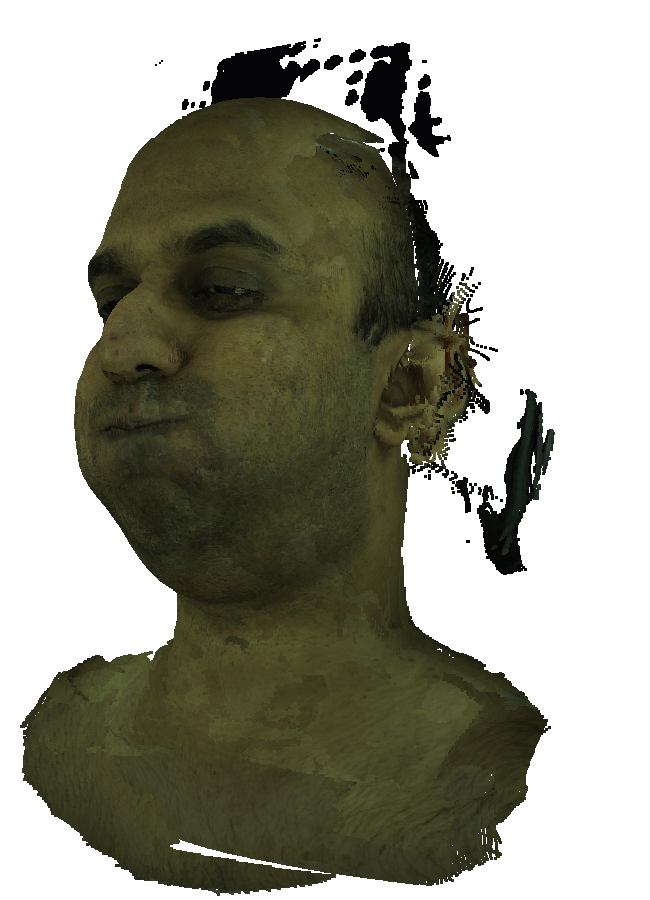}
    \end{subfigure}
    \hfill
    \begin{subfigure}[b]{0.09\textwidth}
        \centering
        \includegraphics[width=\textwidth]{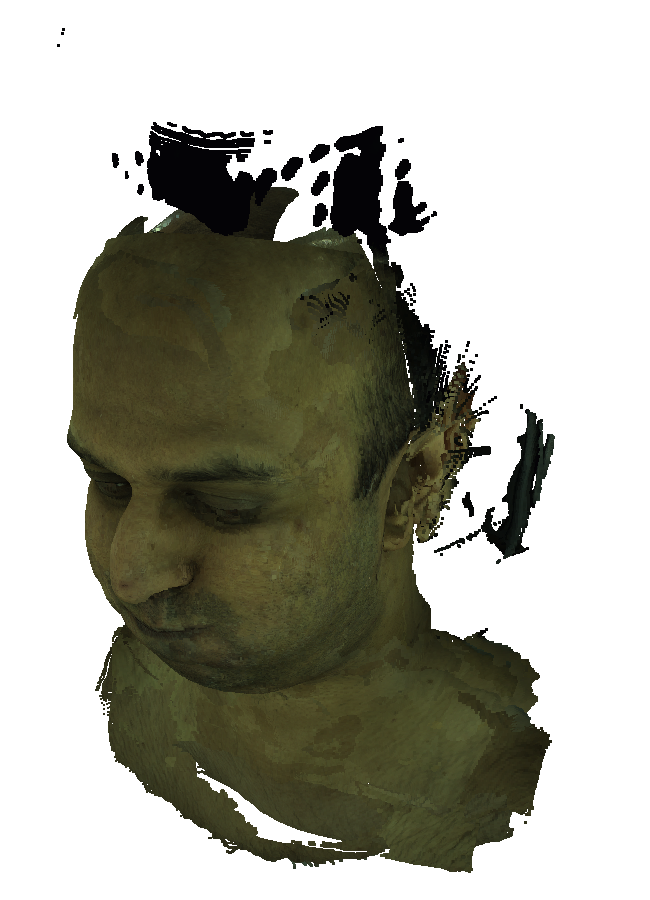}
    \end{subfigure}
    
   \caption{Comparative DUSt3R reconstruction result. All models utilize ground-truth extrinsics and are aligned and scaled to the same pose for better visualization. Our refined intrinsics greatly improve the reconstruction quality for both heads.}
   
    \label{fig:DUSt3R-comparison}
\end{figure*}

\begin{figure}[ht]
    \vspace{1.5mm}
    \centering
    \begin{subfigure}[b]{0.155\textwidth}
        \centering
        \includegraphics[width=0.55857385\textwidth]{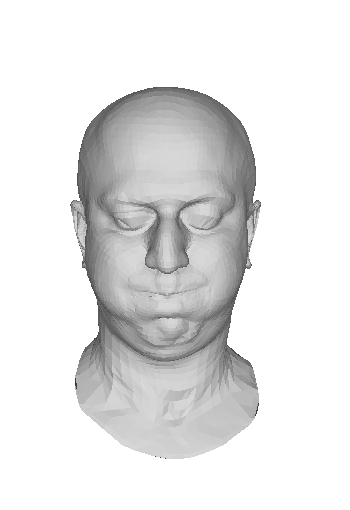}
        \caption{\centering Ground-truth \quad\quad\quad\quad\quad}
    \end{subfigure}
    \hfill
    \begin{subfigure}[b]{0.155\textwidth}
        \centering
        \includegraphics[width=0.55857385\textwidth]{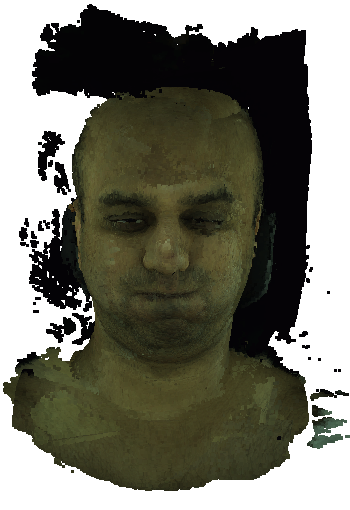}
        \caption{\centering Linear head with refined intrinsics}
    \end{subfigure}
    \hfill
    \begin{subfigure}[b]{0.155\textwidth}
        \centering
        \includegraphics[width=0.55857385\textwidth]{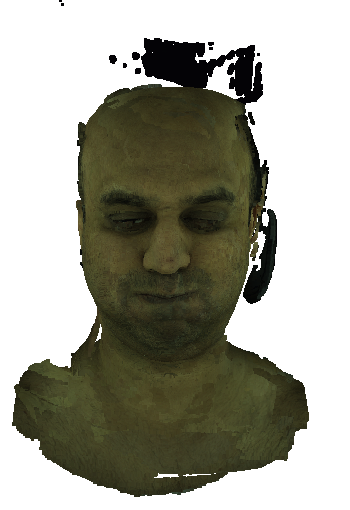}
        \caption{\centering DPT head with refined intrinsics}
    \end{subfigure}
    \\
    \begin{subfigure}[b]{0.235\textwidth}
        \centering
        \includegraphics[width=\textwidth]{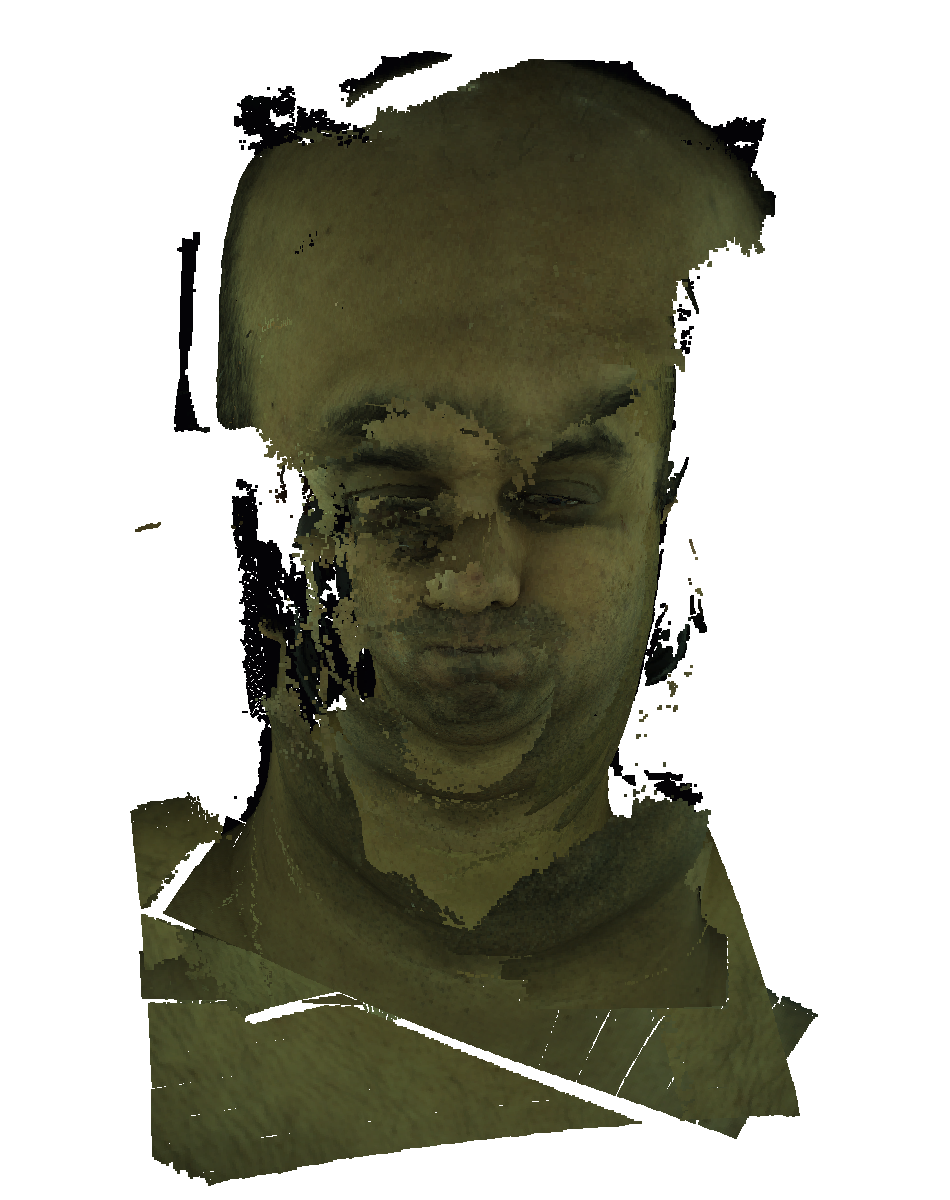}
        \caption{\centering Linear head}
    \end{subfigure}
    \hfill
    \begin{subfigure}[b]{0.235\textwidth}
        \centering
        \includegraphics[width=\textwidth]{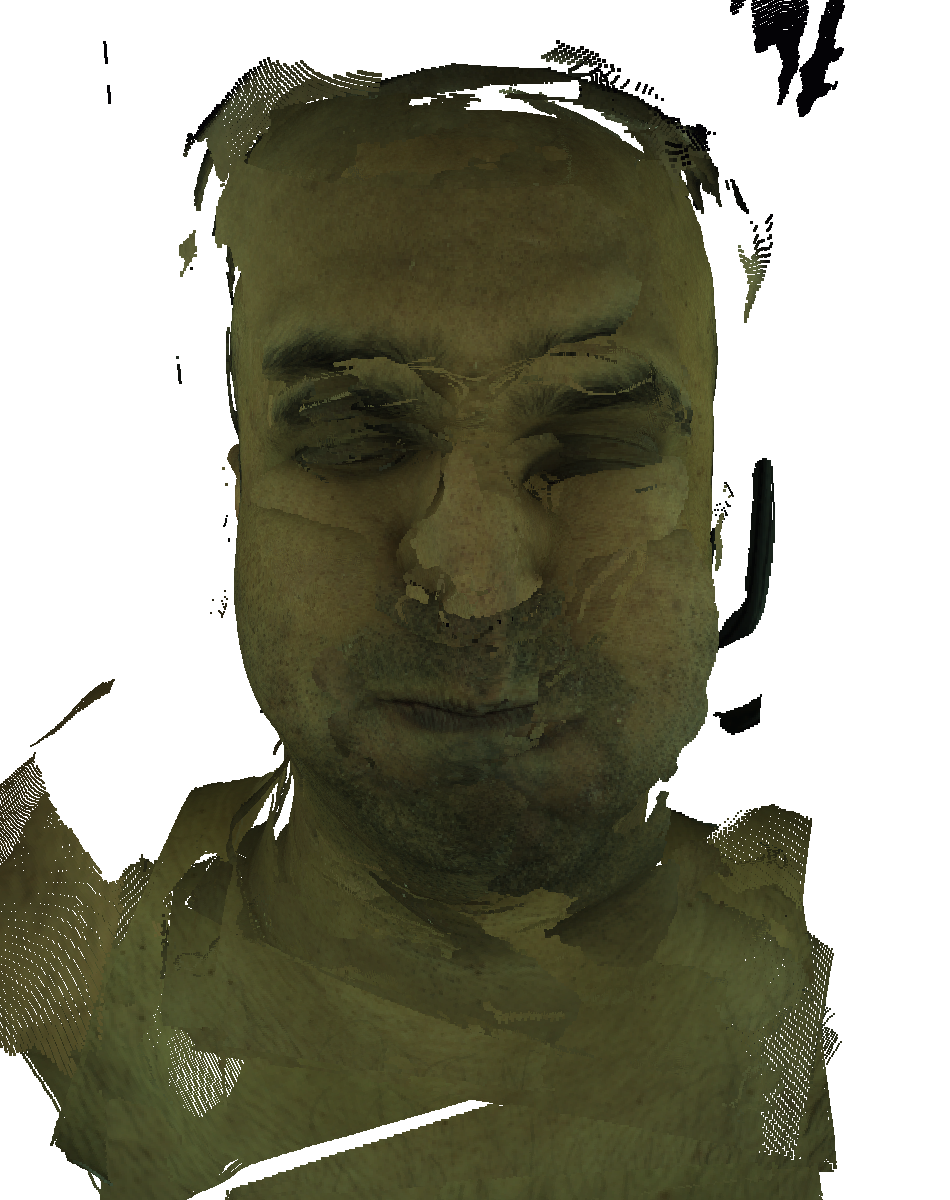}
        \caption{\centering DPT head}
    \end{subfigure}
    
   \caption{Comparison of DUSt3R-reconstructed models vs. ground-truth. First is ground-truth; the next four are DUSt3R reconstructions with extrinsics fixed to ground-truth values: Linear head (our refined intrinsics), DPT head (our refined intrinsics), Linear head (DUSt3R-estimated intrinsics), DPT head (DUSt3R-estimated intrinsics). Models are visualized in the same coordinate system. DUSt3R's learning-based method to compute point maps results in large distortions.}
   
    \label{fig:dust3r_distortion}
\end{figure}

\subsection{Quantitative Results}
\label{sec:experiment-quantitative}
We compared our approach with several state-of-the-art methods, including COLMAP, Pixel-Perfect SfM, DUSt3R, and VGGSfM, on eight frames of the E057 expression of the Multiface dataset. For COLMAP, Pixel-Perfect SfM, and VGGSfM, we append a BA stage with fixed extrinsics to evaluate their performances when ground-truth extrinsics are known. For DUSt3R, it already provides the option to fix extrinsics. For our method, we use the sparse output of Pixel-Perfect SfM with SuperPoint and SuperGlue as our input.

Table \ref{quantitative_result} report absolute and relative errors for focal lengths and principal points. Our multi-frame approach achieved the lowest errors for focal lengths, with $\mathrm{focal}_\mathrm{abs,mean}$ of $5.405$ and $\mathrm{focal}_\mathrm{rel,mean}$ of $0.712$\textperthousand. If multi-frame optimization is disabled, our method still surpasses others, with $\mathrm{focal}_\mathrm{abs,mean}$ of $6.598$ and $\mathrm{focal}_\mathrm{rel,mean}$ of $0.870$\textperthousand. For principal points, our multi-frame approach achieved $1.994$ for $\mathrm{pp}_\mathrm{abs,mean}$ and $1.335$\textperthousand\, for $\mathrm{pp}_\mathrm{rel,mean}$, outperforming others except for COLMAP. However, the differences are ignorable, and COLMAP has significantly higher focal length errors. Because camera intrinsics are defined by both focal lengths and principal points, the overall intrinsics of COLMAP are still very inaccurate and can lead to inconsistency in downstream reconstruction tasks.

Among all the methods, DUSt3R achieves the worst performance, with errors significantly higher than all other approaches. Unlike traditional geometry-based SfM pipelines, DUSt3R employs a learning-based method to estimate dense point maps, which are then used to recover camera parameters. While this approach can produce visually plausible 3D reconstructions, it lacks strict geometric constraints, leading to significant inaccuracies in intrinsic estimation.

VGGSfM, on the other hand, demonstrates exceptional robustness in camera registration, successfully registering all 38 cameras. It also provides more accurate extrinsics estimation than other methods when ground-truth extrinsics are unknown. However, VGGSfM assumes the principal points are centered at the image center and only supports the simple pinhole camera model with $f_x = f_y$, limiting its ability to recover accurate intrinsic parameters. As a result, its intrinsic reconstruction remains comparable to other SfM methods and is still surpassed by our method.

These quantitative results highlight the effectiveness of our approach. In practice, the errors of our approach are negligible, meaning that in most tasks our results can be used as if they were obtained by a dedicated calibration process.

\subsection{Qualitative Results}
\label{sec:experiment-qualitative}

Accurate camera intrinsics are crucial for downstream tasks such as dense 3D reconstruction. Since intrinsics are difficult to visualize directly, we assess their impact by comparing reprojection errors and evaluating 3D reconstructions generated with different intrinsic estimates.

Fig. \ref{fig:reprojection-vis} visualizes reprojection errors for intrinsics estimated by COLMAP, Pixel-Perfect SfM, VGGSfM, and ours. We sample a few points on ground-truth mesh and project them onto images using both ground-truth and estimated intrinsics. Our method achieves the smallest reprojection error.

Fig. \ref{fig:MVS-comparison} compares multi-view stereo (MVS) reconstructions using COLMAP’s MVS pipeline with different intrinsics as input. We compute point-wise distances to the ground-truth mesh and visualize them in RGB colors, where green denotes near-zero deviation. Our method exhibits a tighter concentration around green, demonstrating superior intrinsic accuracy compared to other methods.

Additionally, we evaluate the effect of intrinsics on DUSt3R reconstructions, using its linear and DPT head models. We provide DUSt3R with ground-truth extrinsics but test two sets of intrinsics: (1) intrinsics estimated by DUSt3R and (2) intrinsics refined by our method. As shown in Fig. \ref{fig:DUSt3R-comparison}, the models using DUSt3R's intrinsics exhibit large noise and depth inconsistencies, whereas models reconstructed using our refined intrinsics achieve more stable and accurate geometry. Moreover, because DUSt3R's intrinsics significantly deviate from ground-truth, the reconstructed models differ in scale from the ground-truth mesh, as shown in Fig. \ref{fig:dust3r_distortion}.

\subsection{Ablation Study}
\label{sec:experiment-ablation}
We conducted an ablation study to evaluate the impact of each component, as summarized in Table~\ref{ablation_result}. Starting with the baseline using only the reprojection term (Row 1), the absolute focal length and principal point errors are $7.307$ and $2.391$, respectively. Adding the extrinsics regularization term (Row 2) slightly increases the focal length error, highlighting the importance of progressive coefficient adjustment. With this adjustment (Row 3), errors decrease to $6.616$ and $2.241$. Including the dense feature reprojection term (Row 6) further reduces errors to $6.598$ and $2.227$. Finally, adding the intrinsics variance term for multi-frame optimization (Row 7) achieves the best results, with absolute focal length error of $5.405$ and absolute principal point error of $1.994$, showcasing the combined effectiveness of these components.

\section{CONCLUSIONS}

\label{sec:conclusion}
We proposed a dense-feature-driven multi-frame camera calibration method for large-scale camera arrays. By proposing extrinsics regularization, dense feature reprojection, and intrinsics variance terms with multi-frame optimization, our approach achieves calibration-level precision without dedicated captures. Compatible with existing SfM pipelines, it offers an efficient solution for large-scale setups. For future work, we aim to further optimize computational efficiency for handling a large number of input frames and extend our method to accommodate more complex camera models, including cameras with lens distortions.

\bibliographystyle{IEEEtran}
\bibliography{refs}

\end{document}